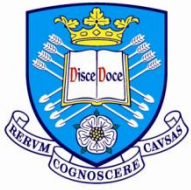

Aerospace Engineering.
Computer Science.

**Aerospace Engineering
Final Year Project**

# Design of Artificial Intelligence Agents for Games using Deep Reinforcement Learning

**Andrei-Claudiu Roibu**

May 2019

Supervisor: **Prof. Dr. Eleni Vasilaki**

Dissertation submitted to the University of Sheffield in partial fulfilment of the requirements for the degree of

**Master of Engineering**

# Abstract

In order perform a large variety of tasks and to achieve human-level performance in complex real-world environments, Artificial Intelligence (AI) Agents must be able to learn from their past experiences and gain both knowledge and an accurate representation of their environment from raw sensory inputs. Traditionally, AI agents have suffered from difficulties in using only sensory inputs to obtain a good representation of their environment and then mapping this representation to an efficient control policy. Deep reinforcement learning algorithms have provided a solution to this issue. In this study, the performance of different conventional and novel deep reinforcement learning algorithms was analysed. The proposed method utilises two types of algorithms, one trained with a variant of Q-learning (DQN) and another trained with SARSA learning (DSN) to assess the feasibility of using direct feedback alignment, a novel biologically plausible method for back-propagating the error. These novel agents, alongside two similar agents trained with the conventional backpropagation algorithm, were tested by using the OpenAI Gym toolkit on several classic control theory problems and Atari 2600 video games. The results of this investigation open the way into new, biologically-inspired deep reinforcement learning algorithms, and their implementation on neuromorphic hardware.

# Acknowledgements

I would like to express my thanks and gratitude to my supervisor, Prof. Dr. Eleni Vasilaki, for her guidance throughout this project and her entrusting me with such a difficult task, especially considering that my academic background is in aerospace and aeromechanical engineering, and not computer science.

I would also like to acknowledge the help that I have received from the researchers and PhD students in the Machine Learning and Algorithms research groups, among which I would like to name Avgoustinos Vouros, Luca Manneschi, Nada Yem Abdel Rahman, Zehai Tu and Juan Jose Giraldo Gutierrez. I would also like to thank Dr. Twin Karmakharm and Dr. Peter Heywood from Research Software Engineering (RSE) Sheffield for their support in granting me access to the NVIDIA® DGX-1™ supercomputer.

I would also like to express my gratitude to my former colleagues from Rolls-Royce Plc., which have instilled into me a passion for machine learning and inspired me to pursue a career in this field. They include David Debney (now at the Aerospace Technology Institute), Dr. Dimitris Triantafyllidis, Adam Morton, Maitham Deeb (now at Babylon Health), Dr. Muhannad Alomari, Mihai Mitache (now at Jaguar Land Rover), Dr. Bryce Conduit and Dr. Chloe Jo Palmer.

I must acknowledge the help, encouragement and stimulus that I received from my loved ones and good friends, among which I would like to mention Oana Pelea, Răzvan Apetrei, Ruxandra Mîndru, Adrian Stan, Andreea Maria Oncescu, Dr. Alexandra Ștefănescu (who also helped proof read this report), Andrei Rădulescu, Laura Heraty, Deacan Robinson and Will Gilmour.

Finally, I would like to thank my family for their support over the past few years, particularly my father, Crinel Roibu, and my mother, Raluca Roibu.

# Contents

# List of Tables

# List of Figures

# Nomenclature

## Symbols

| | |
|---|---|
| $A$ | Agent Action |
| $a$ | Arbitrary Agent Action |
| $B$ | DFA Random Fixed Weights |
| $b$ | Bias |
| $C$ | Cloning Step Number |
| $D$ | Experience-Replay Memory |
| $d$ | Derivative of a Variable |
| $E$ | Environment |
| $e$ | Experience Vector |
| $f, g, s$ | Functions |
| $G$ | Expected Return |
| $g$ | Non-Linear Activation Function |
| $h$ | Height |
| $I$ | Image |
| $J$ | Cost Function |
| $i, j$ | Variables |
| $k$ | Kernel Filter |
| $L$ | Loss Function |
| $M$ | Total Number of Episodes |
| $N$ | Experience-Replay Capacity |
| $m, n$ | Metrics |
| $P$ | Generated Distribution |
| $p$ | Dynamics of MDP |
| $Q$ | Action-Value Function |
| $q$ | Random Action Value |
| $R$ | Reward |
| $r$ | Random Reward |
| $S$ | State |
| $s$ | Random State |
| $T$ | Total Time |
| $t$ | Time |
| $V$ | State-Value Function |
| $W$ | Weight |
| $w$ | Width |
| $x, y$ | Variables |
| $Z$ | Weighted Sum |

## Greek Symbols

| | |
|---|---|
| $\alpha$ | Learning Rate |
| $\beta$ | RMSprop Decay |
| $\gamma$ | Discount Factor |
| $\Delta$ | Gradient / Derivative |
| $\varepsilon$ | Fraction of Time |
| $\epsilon$ | Small Scalar (RMSprop) |
| $\theta$ | Network |
| $\Lambda$ | Activation |
| $\xi$ | Softmax Temperature |
| $\pi$ | Policy |
| $\tau$ | Age Metric |
| $\phi$ | Feature Map |
| $\Psi$ | Weighted Average |

## Subscripts and Superscripts

| | |
|---|---|
| c | Cloned Network |
| $i, j$ | Variables |
| $l$ | Layer Number |
| $out$ | Output Layer |
| $slice$ | Section of the Input Data |
| $T$ | Transpose of a Matrix |
| $t$ | Time |
| $'$ | Future State |
| $*$ | Optimal |
| $\hat{}$ | Empirical Distribution |

## Abbreviations

| | |
|---|---|
| AI | Artificial Intelligence |
| A3C | Asynchronous Actor-Critic Agents |
| BP | Backpropagation |
| CNN | Convolutional Neural Network |
| CONV | Convolutional Layer |
| CPU | Central Processing Unit |
| DFA | Direct Feedback Alignment |
| DL | Deep Learning |
| DNN | Deep Neural Networks |
| DQN | Deep Q-Network |
| DRL | Deep Reinforcement Learning |
| DSN | Deep SARSA Network |
| DSQN | Deep SARSA Q-Network |
| GPU | Graphics Processing Unit |
| HPC | High Performance Computer |
| LSTM | Long-Short Term Memory |
| M | Million |
| MDP | Markov Decision Process |
| ML | Machine Learning |
| MOOC | Massive Open Online Course |
| NN | Neural Network |
| RAM | Random Access Memory |
| ReLU | Rectified Linear Unit |
| RGB | Red-Green-Blue |
| RL | Reinforcement Learning |
| RNN | Recurrent Neural Network |
| RSE | Research Software Engineering |
| SSH | Secure Shell |
| STDP | Spike-Timing-Dependent Plasticity |
| TD | Temporal Difference |
| TDL | Temporal Difference Learning |

# 1 Introduction, Aims and Objectives

## 1.1 Project Background

Artificial Intelligence Agents are increasingly being used in the real world, interacting with humans across a large variety of tasks. According to the Law of Accelerating Returns (Kurzweil, 1999), this trend will increase exponentially over the coming years. To achieve human-level performance in complex environments, AI Agents must be able to learn from their past experiences and gain both knowledge and an accurate representation of their environment from raw sensory inputs.

This can be achieved through reinforcement learning, a class of machine learning techniques which enables agents to interact with their environment while striving to achieve a goal. The application of reinforcement learning algorithms ranges from Uber and Tesla's driverless cars to Google's AlphaGo algorithm which defeated the World Champion in the game of Go. Agents trained with reinforcement learning algorithms often reach new, non-intuitive solutions, which have both amazed and frightened experts, with public figures such as Elon Musk and Steven Hawking warning about the dangers of AI. Though, traditionally, these agents have suffered from difficulties in efficiently representing their environment by only using sensory inputs. Deep reinforcement learning techniques, however, provide a solution to this issue by using deep neural networks to extract high-level features from sensory data, which are then fed to the reinforcement learning algorithm.

This project examined how different deep reinforcement learning algorithms can be used to create AI agents capable of learning only by using raw sensory inputs. The performance of both conventional and novel agents was evaluated, including the first-time application of a biologically-inspired learning algorithm in the field of DRL. The results of this investigation provided more insight into how DRL agents work and opened the door to new, biologically-inspired deep reinforcement learning algorithms and their implementation on neuromorphic hardware. The created agents are freely accessible on GitHub: https://github.com/aroibu1/uos_drl .

## 1.2 Aims and Objectives

The aim of this project was to investigate the performance of different AI agents created with deep reinforcement learning algorithms by testing their ability to learn how to solve different games of varied complexity. The project objectives are presented in Table 1.1, alongside their status and semester-based deadlines.

*Table 1.1: Project Objectives*

| Objective | Semester | Achieved? |
|---|---|---|
| Gain an understanding of artificial neural networks and deep learning techniques. | 1 | Yes |
| Investigate reinforcement learning algorithms and how to implement them together with deep neural networks. | 1 | Yes |
| Examine previous studies and the current state of the art by means of a literature review. | 1 | Yes |
| Develop an AI Agent by using a Deep Reinforcement Learning algorithm. | 2 | Yes |
| Train the AI Agent to play a selected game by using only sensory inputs. | 2 | Yes |
| Collect data on the performance of the AI Agent and compare it to previous studies. | 2 | Yes |
| Discuss the obtained results and present the appropriate conclusions of the performed work. | 2 | Yes |
| **Optional Objectives** | **Semester** | **Achieved?** |
| Modify the AI Agent and compare the performance of different Deep Learning and Reinforcement Learning algorithms. | 2 | Yes |
| Present the work done and obtained results at a national conference. | 2 | No |
| Publish a journal paper detailing the work done and the obtained results. | 2 | No |

## 1.3  Overview of the Report

This report presents all the work done during this project, together with relevant technical information regarding deep and convolutional neural networks, as well as classical and deep reinforcement learning. The interested reader can find a series of more intuitive explanations, as well as some general and more advanced resources in Appendix A. The more important concepts relating to this project are discussed in the body of the report.

The report opens with a detailed literature review of the fields of deep and reinforcement learning, in Section 2. This contains a brief history of each field and a discussion of previous work and the state of the art. This helped identify any knowledge gaps that could be filled through this study.

Following this, some discussion about the requirements, analysis and relevant project management considerations is carried out in Section 3. In this section, the tools and methods used during the testing of the various agents are discussed, as well as some hardware considerations.

Section 4 then presents the numerical methods and techniques used in developing the DRL algorithms. This section focuses on the relevant mathematical apparatus employed in this project.

Building on this, Section 5 presents the methodology employed in developing the various algorithms, as well as how these algorithms are constructed together with implementation particularities.

In Section 6, a discussion is carried out regarding the various implementation details, ranging from the test environments to the employed programming language, code implementation and procedures used in tuning and evaluating the agents.

Section 7 then presents a discussion of the obtained results, and an analysis of how the different architectures impact the performance of the various agents, both during and after training.

Finally, the report closes with Section 8, which discusses the conclusions and the proposed further work, and Section 9 which is a self-review of the author.

## 2    Literature Review and Related Work

This section provides a detailed insight into the performed background reading in the fields of Deep and Reinforcement Learning. The brief history of each field is presented, followed by a discussion of previous relevant work. The knowledge thus captured will be used in developing the relevant methods, architectures and algorithms.

## 2.1   Deep Learning

Deep Learning represents a family of algorithms which are part of the machine learning field of AI methods. In general, agents trained using machine learning gain an understanding of their environment by extracting features and patterns from raw data. Nevertheless, the performance of these agents is dictated by how these features were formalized in advance by a programmer, as was the case of the IBM Deep Blue chess-playing agent (Campbell, Hoane and Hsu, 2002).

In deep learning, however, complex features are constructed automatically out of progressively simples, less abstract representations of the data, across a series of steps. The ability of DL agents to gather experience from their environment increases with the number of levels of this nested hierarchy of features (Goodfellow, Bengio and Courville, 2016). Thus, agents can learn advanced concepts and gain experience from large datasets without the need for formal rules to be defined in advance. In one of his papers, Krizhevsky has found that DL agents can actually derive better features than those formalized by a human programmer when sufficient data is fed to the system (Krizhevsky, Sutskever and Hinton, 2012). This has allowed agents trained in this fashion to solve tasks which are intuitive to humans and hard to formalize, such as recognising objects.

### 2.1.1   Brief History of Deep Learning

In their book, Goodfellow, Bengio and Courville suggest that the idea of intelligent machines was first proposed in the ancient Greek myths of Pygmalion, Daedalus and Hephaestus (Goodfellow, Bengio and Courville, 2016). Modern DL, though, emerged in the 1940s as a series of biologically inspired methods and algorithms designed in order to better understand the functions of the brain. Initial models used a series of inputs and manually tuned weights to produce positive and negative outputs for an input signal (McCulloch and Pitts, 1943). Later models were capable of learning the weights automatically from input training data (Rosenblatt, 1958). Despite this initial biologically inspired work, modern DL is based on theories originating in algebra, probabilities and numerical optimization (Goodfellow, Bengio and Courville, 2016).

Two important steps forward in the field of DL were made in 1986. The first was the idea of distributed feature representation, which proposed that simple computation units (or neurons) representing features should be placed in a network and used to construct multiple subsequent complex features (Hinton, McClelland and Rumelhart, 1986). The second was the successful implementation of the backpropagation algorithm, which tunes the network weights by back-propagating the error computed between the target and the obtained outputs (Rumelhart, Hinton and Williams, 1986).

The breakthrough that launched the modern DL wave occurred in 2006, when a deep belief network was forward trained one layer at a time and tuned by using backpropagation (Hinton, Osindero and Teh, 2006). Coupled with the larger availability of datasets due to the digitization of society, the increased computing power of CPUs and GPUs and the increased complexity of models and network infrastructures, this has led to the development of the modern deep learning field of AI.

## 2.1.2 Machine Vision and Convolutional Neural Networks

Vision is probably the most important human sense in terms of gaining an accurate representation of one's environment, with up to 80% of human learning and cognition being tied to it (Dinh et al., 2002; Ripley and Politzer, 2010; San Roque et al., 2015). Therefore, in order to allow AI agents to gain an accurate representation of their environment, it is necessary to give them the ability to process, interpret and understand images.

Convolutional Neural Networks (CNNs) represent a type of deep neural network inspired by work done in how images are processed in the early visual cortex (Hubel and Wiesel, 1963) and are currently some of the most used algorithms in the field of computer vision. These networks make use of several elements (Table 2.1) which allow them to learn all the necessary filters required in processing an input image without the need of human-defined features. These elements are further discussed in Section 4.1.

*Table 2.1: CNN elements*

| Element | Description |
|---|---|
| Convolutional Layer | - Transforms an input volume into an output volume of different size;<br>- Exploits local spatial correlations present in images; |
| Padding | - Adds a layer of zeros around the border of a volume, which helps keep more of the information from the volume border when convolutions are applied; |
| Pooling Layer | - Reduces the height and width of an input;<br>- Reduces the required computational power and makes feature detectors more invariant to input positions;<br>- Can be either max-pooling or average-pooling:<br>  o Max-pooling stores the maximum value present in a pooling window;<br>  o Average-pooling stores the average value of the elements in a pooling window; |

The first CNN was introduced in 1989 by Weibel (Weibel et al., 1989) and it shared weights along a single temporal dimension (LeCun and Bengio, 2003). In the same year, LeCun proved that the backpropagation algorithm can be used in training a NN which was fed directly with images instead of feature vectors (LeCun et al., 1989).

These two findings contributed to the development of the LeNet-5 CNN, which recognized hand-written digits from $32x32$ pixel greyscale images (LeCun et al., 1998). This CNN was composed out of 3 convolutional, 2 pooling and 2 fully connected layers and had approximately $60k$ parameters. As the network becomes deeper, the $32x32$ input images are reduced to $5x5$ features maps spread over 16 channels. The first fully connected layer maps all the 400 nodes to 120 neurons, while the second connects 84 neurons to each of the features from 0 to 9. The LeNet-5 algorithm has several particularities when compared to modern CNNs. Firstly, it does not apply any padding. Then, it uses average pooling layers, as this was the preferred method at the time. It also uses a sigmoid activation function although modern networks typically use other functions such as ReLU (Nair and Hinton, 2010). Moreover, in order to reduce the computational cost and to force the extraction of specific features, the second convolution is only applied to select features. Finally, the weights of the Euclidian Radial Basis function, which computes the L2 distance between the targets and the outputs, are also set manually rather than learned.

Despite the success of LeNet-5 in digit classification, CNNs only started to be widely used in the computer vision field after the publication of the AlexNet architecture (Krizhevsky, Sutskever and Hinton, 2012). Compared to LeNet-5, this was a much deeper network, containing 5 convolutional and 3 fully connected layers and consisting of approximately $60M$ parameters, being capable of classifying $227x227$ coloured images. The main innovations introduced by AlexNet was the use of ReLU activation functions, a softmax function in the final layer, and using GPUs for training. The ReLU activation function reduces the probability of gradients vanishing or exploding when training deep networks and allows the network to obtain a sparse representation. This leads to faster learning speeds, reduced network sizes and a reduction in the chance of overfitting. The softmax function allowed the network to classify images into any of 1000 different classes. Starting with AlexNet, all editions of the ImageNet Large Scale Visual Recognition Challenge (ILSVRC) were won by CNNs (Figure 2.1). The Top-5 error indicates the percentage of images for which the correct, desired class, was not within the top-5 predicted classes.

Based on the AlexNet findings, Simonyan and Zisserman introduced VGG16, a simpler yet much deeper network, which proposed chaining multiple small convolutional layers with ReLU activations (Simonyan and Zisserman, 2015). The increased depth induced many nonlinear transformations which allowed the network to learn complex parameters, yet it's very large size ($138M$ parameters) made it very computationally expensive to train.

The increased depth could also lead to exploding or vanishing gradients during training. This led to the creation of several other new architectures. To solve this problem, Lin, Chen and Yan proposed the idea of inception modules, where several convolution layers and a max-pool layer are trained simultaneously and then concatenated (Lin, Chen and Yan, 2014). Thus, instead of selecting the filter sizes in advance, the user can allow the network to determine the optimum filter size and associated parameters. This idea was then developed into the GoogLeNet architecture, which was smaller than AlexNet despite being much deeper (Szegedy et al., 2015).

Meanwhile, He observed that with an increase in depth, the network accuracy saturates and then degrades due to inherent training and optimization difficulties, rather than due to overfitting (He et al., 2015). This prompted him to propose the concept of Residual Networks (ResNets) in which connections using an identity or convolutional block are established between the output of a single or multiple layers and the input to those layers (He et al., 2015). This function preserves much of the original information, which prevents exponential degradation to 0 by backpropagating errors directly to earlier layers. Using this finding, Szegedy combined the residual network with the inception idea to create the Inception-ResNet algorithm (Szegedy et al., 2016). Hu took this further, by proposing the "Squeeze-and-Excitation" blocks, which model interdependencies between channels by making use of fully-connected layers, the inception idea and residual blocks (Hu et al., 2018).

Rather than experimenting with different blocks and architectures, Zoph proposed a new concept named Neural Architecture Search, which uses an RNN with reinforcement learning to generate a CNN architecture and tune its hyperparameters so that to maximise its accuracy for a given dataset and defined reward (Zoph and Le, 2017). This idea was used to generate the NASNet model (Zoph et al., 2018). Liu used a similar approach, however instead of reinforcement learning, he used a sequential model-based optimization strategy (Liu et al., 2018). His approach constructs and tests simple CNN models, and then stacks the bests ones, repeating until a maximum number of blokes is reached. This approach obtained similar accuracies to NASNets but trained $5x$ faster.

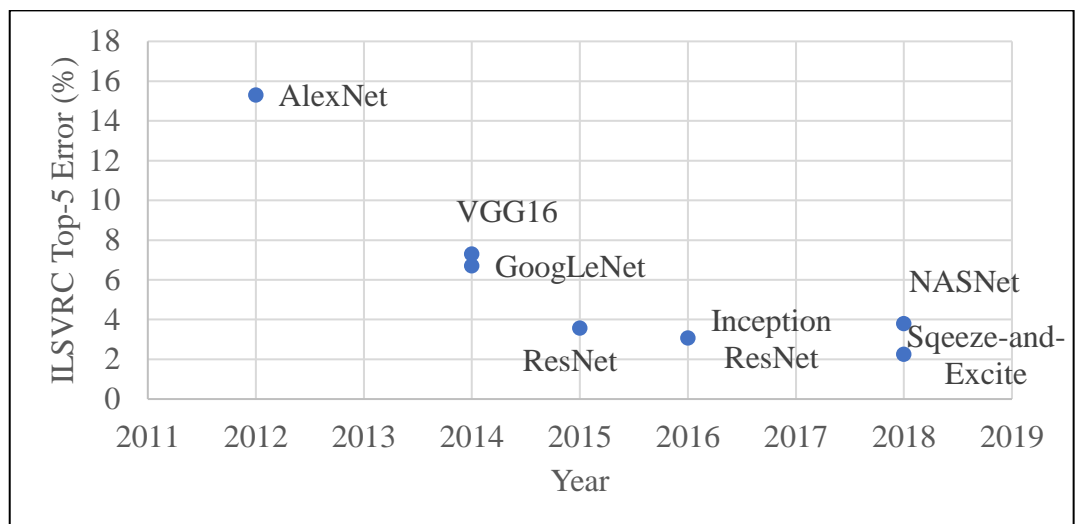

*Figure 2.1: Evolution of ImageNet top-5 error rates*

### 2.1.3 Training Algorithms

The backpropagation (BP) algorithm was first developed in the 1960s (Schmidhuber, 2015), successfully implemented in training neural networks in the 1980s (Rumelhart, Hinton and Williams, 1986; LeCun et al., 1989) and introduced to deep learning in 2006 (Hinton, Osindero and Teh, 2006). It is currently one of the most common training algorithms, working by backward propagating the loss function through successive mutual connections. Despite its proven performance, it is not the only training algorithm available for NNs, with alternatives such as Contrastive Hebbian Learning (Xie and Seung, 2003) and Target-Propagation (Lee et al., 2015) being proposed among others.

Another interesting alternative is the Direct Feedback Alignment algorithm, proposed by Nøkland (Nøkland, 2016) which is derived from the direct feedback alignment method (Lillicrap et al., 2014). This biologically plausible algorithm disconnects the forward and backward paths, propagating the loss directly from the output layer to each individual hidden layer. In contrast to BP, DFA trains each hidden layer independently, by using a set of fixed weights, and provides a different update gradient to that of steepest descent. This method could provide a step towards biologically plausible deep learning and its implementation on neuromorphic hardware.

Due to its nature, ML can be used in conjunction with neuromorphic hardware to solve problems, test neuroscience theories and create biologically-plausible systems capable of learning and adapting (Neftci et al., 2017; Schuman et al., 2017). Although BP is used in neuromorphic systems, it either requires tailoring for each hardware type or that the learning takes place off the neuromorphic chip. This happens as BP requires information not local to the computational block in order to perform highly precise calculations involving the error signal and symmetrical weights (Neftci et al., 2017). Moreover, BP requires multiple alternating forward and backward passes that need to be synchronized in order to allow the backward pass to access information about the activity of the forward pass, which is not compatible with biological systems (Detorakis, Bartley and Neftci, 2018). The restrictive nature of BP in terms of neuron and network type and topology has led to the creation of other biologically inspired methods, such as STDP and Hebian Learning (Schuman et al., 2017). Nevertheless, these have not yet been demonstrated to be widely applicable. DFA neither requires symmetrical weights nor reciprocal connections, and can provide local training signals, specific to each computational block. Moreover, it provides a short path for the local error signal, being a potential contender for neuromorphic hardware applications (Nøkland, 2016). Although both Nøkland, Han and Yoo found that DFA performs worse than BP for a simple CNN, the later have suggested a method for overcoming this limitation by combining BP and DFA in the Binary Direct Feedback Alignment algorithm, which has shown good performance for a CNN (Han and Yoo, 2019).

## 2.2 Reinforcement Learning

Like DL, Reinforcement Learning represents a class of algorithms part of the ML field of AI methods. RL represents a computational approach to learning a series of actions based on an agent's perception of its current state and the consequences of its direct interactions with its environment while pursuing a goal. These three aspects (sensation, action and goal) and their interaction with the agent are captured by a formal framework known as a Markov Decision Process (Sutton and Barto, 2018).

For an agent to learn a policy of actions which will guide it through different states, it needs to deal with the uncertainty associated with an unknown environment and grasp the causes and effects of the actions it performs in order to reach an explicitly defined goal. In other words, an agent needs to sense and explore its environment in order to gain knowledge, and then exploit that knowledge to find which actions it needs to take to reach its goal. Due to the uncertainties associated with the environment, actions can have both short- and long-term consequences, impacting both future environment states and any actions which the agent will take when in those states. To fully predict the effects of its actions, an agent needs to know all the present and future states of its environment, which is often impossible. This is known as the explore-exploit dilemma.

To solve this dilemma, an agents needs to take random actions in order to explore its environment and progressively favour those which seem the best, while also constantly monitoring its environment (Sutton and Barto, 2018). The best actions can be determined by their value, or the total amount of reward that the agent can expect from an action when in each state. However, the previously predicted value of an action could change as the different states change. Thus, the prediction accuracy is influenced by the size of the considered timeframe. Therefore, the function which predicts the value of each action needs to be recalculated in each state based on current observations and the gathered agent experience. Value functions are key to solving RL problems, as they dictate both the exploration strategy and the generation of action policies.

Opposite to other ML techniques, such as DL, which essentially extrapolate a pattern or correlation between input and output datasets, RL agents learn by trying to maximise an often noisy reward signal, which they learn from interactions with their environment (Sutton and Barto, 2018). While DL training data is correlated, independent and organised, the reward signal in RL can be dissociated from agent actions, with correlations occurring only over many time steps (Mnih et al., 2013).

## 2.2.1 Brief History of Reinforcement Learning

In their book, Sutton and Barto consider that modern reinforcement learning resulted from the intertwining of three research threads: optimal control, trial-and-error learning and temporal-difference learning (Sutton and Barto, 2018).

The optimal control thread refers to research aiming at designing the control mechanism for a dynamical system. This problem was originally investigated by Bellman, who proposed the concepts of system state and value function, also known as a Bellman equation (Bellman, 1956) and defined the MDP for the optimal control problem (Bellman, 1957). Later, Bellman proposed dynamic programming as a way to solve the Bellman equation (Bellman, 2003). This method, though, suffers from high computational costs when the number of states increases (Sutton and Barto, 2018).

While the optimal control thread looks at different controller designs, the trial-and-error thread is concerned with environment exploration by random action evaluation. The concepts behind this idea were first demonstrated using electro-mechanical machines (Ross, 1933). In AI, this learning method was first proposed by Turing in this work concerning pleasure-pain systems (Turing, 1948) and implemented by Farley and Clark, which designed a NN capable of learning by trial-and-error (Farley and Clark, 1954). This first success was followed by other algorithms, such as STeLLA (Andreae, 1963) and BOXES (Michie and Chambers, 1968). Another milestone was set by Tsetlin, who proposed using trial-and-error learning to solve the explore-exploit dilemma for the multi-armed bandit problem (Tsetlin, 1973).

Finally, the temporal-difference learning thread proposes a series of methods for calculating and updating a quantity by using the differences between successive future predictions of that quantity. TD ideas were first implemented by Samuel to solve a game of checkers (Samuel, 1959) and later developed into the TD(0) learning rule, part of a dynamic system controller (Witten, 1977). Klopf combined TD ideas with trial-and-error methods to create the concept of generalized reinforcement (Klopf, 1972), which was further developed by Barto, Sutton and Anderson into the actor-critic learning method (Barto, Sutton and Anderson, 1983). Building on the work done by Samuel and Witten, Sutton also proposed the TD($\lambda$) algorithm (Sutton, 1988). The TD and optimal control ideas were combined by Watkins in creating the Q-learning algorithm (Watkins, 1989). Agents trained with this algorithm estimate the value of an action in a given state by assessing the long-term reward of that action, making this algorithm desirable when the optimal publicity is sought. Rummery and Niranjan proposed a modified version of Q-learning, known as SARSA (Rummery and Niranjan, 1994), where the agent will assess the value of an action based on the policy that it has used to arrive in its current state. This learning approach yields good performance while training, although it might not reach the optimal policy.

## 2.2.2 Deep Reinforcement Learning

Over the years, numerous applications were proposed in order to demonstrate the learning abilities of different RL algorithms. One such implementation was TD-Gammon (Tesauro, 1995), which made used of a $TD(\lambda)$ algorithm and a NN trained with backpropagation which calculated the state value function. The agent was one of the first to learn by self-playing against itself (Sutton and Barto, 2018). The algorithm, yet, suffered from the fact that the NN prediction was aided by the randomness of the dice roll (Pollack and Blair, 1996). It was also prone to weight oscillations or even divergence due to the use of a NN function approximator for the RL algorithm (Baird, 1995; Tsitsiklis and Van Roy, 1997; Gordon, 2001). The TD-Gammon algorithm was later used to estimate the state value function for the IBM Watson (Tesauro et al., 2012). Although the IBM Watson employed specialized hand-designed feature vectors to represent the states and made use of an bespoke action value function, it was able to learn and answer natural language questions and showed better-than-human decision-making skills (Tesauro et al., 2013).

Based on these findings, Mnih from DeepMind proposed the DQN algorithm, which makes use of a deep CNN, the Q-learning algorithm and an experience replay mechanism (Mnih et al., 2013). Compared to the IBM Watson, DQN automates the feature design by using a CNN fed with sufficient data (Krizhevsky, Sutskever and Hinton, 2012), which estimates the action value function. A semi-gradient form of Q-learning is also used instead of TD-Gammon, as the later requires full knowledge of all possible states, which would be time and computationally expensive (Sutton and Barto, 2018). The instabilities observed in TD-Gammon are also mediated by the use of the experience replay mechanism, which works by storing every experience that the agent has in a dataset and them randomly samples and feeds these experiences to the Q-learning algorithm (Lin, 1992). This helps the CNN break data correlations, prevents parameter oscillations or divergence and increases learning efficiency by allowing experiences to be used in the calculation of multiple weights (Mnih et al., 2013, 2015). The algorithm is trained using the Atari 2600 learning and testing environment, which provides the agent with $210x160$ RGB image inputs at 60 Hz (Bellemare et al., 2013). DQN was tested on 49 Atari games, taking only the video, the reward and a list of actions as inputs and achieving human comparable performance in 29 of them (Mnih et al., 2015).

Using their original findings, the DeepMind team further improved the DQN algorithm performance by using two CNNs: one learned a value function for determining the policy while another calculated the policy value (van Hasselt, Guez and Silver, 2015). They also further refined greedy policy associated errors (Bellemare et al., 2015). The experience reply mechanism was then modified to prioritize specific experiences (Schaul et al., 2016) and to normalise learning between the two CNNs (Wang et al., 2016). Learning times were reduced and performance improved at the cost of increased running time by aggregating the outputs of different policies ran across different learning episodes (Osband et al., 2016) and rescaling the rewards by using adaptive normalization (van Hasselt et al., 2016). Using a universal goal-directed value function approximator (Schaul et al., 2015) together with the policy of a previously trained agent allowed the team to train a single network to learn from and play multiple games (Rusu et al., 2016). Training was also sped up by using multiple parallel agents which generated experience of the same environment for the experience replay mechanism (Nair et al., 2015). Utilizing Nair's findings, Mnih replaced experience replay with asynchronous gradient descent, which decorrelated the data and allowed him to train the NN using SARSA (Mnih et al., 2016), while Vezhnevets proposed using a RNN in conjunction with the CNN to learn and update the long-term action plan (Vezhnevets et al., 2016). Using the collective lessons learned through their work, the DeepMind team designed the AlphaGo (Silver et al., 2016) and AlphaGo Zero (Silver et al., 2017) algorithms, the latter being trained solely through self-play without any real world guidance. More recent improvements have allowed DRL agents to solve increasingly complex games by using short term exploration bonuses (Burda et al., 2018) and to predict the reward functions from human feedback by using a deep NN (Ibarz et al., 2018).

The DQN algorithm has also been used for solving complex physics applications in continuous domains (Lillicrap et al., 2016). This opened the way for DRL to be used in complex real-world problems. OpenAI's Dactyl algorithm is capable of training a robotic arm to solve a physical system by using only simulations (Andrychowicz et al., 2018). In contrast to DQN, it uses a separately trained CNN to predict a physical object's pose, an LSTM RNN to represent the policy and a Proximal Policy Optimization RL algorithm to optimize the policy and calculate the value function (Schulman et al., 2017; Andrychowicz et al., 2018).

Despite most research efforts being concentrated on optimizing Q-learning based DRL algorithms, some authors have analysed the possibility of using the SARSA algorithm. Zhao kept Mnih's DQN architecture but used SARSA instead of Q-learning, fining that his DSN converges smoother, is more stable for complex systems and even achieves better scores on some games versus DQN (Zhao *et al.*, 2017). Similarly, Xu introduced DSQN (Xu et al., 2018) which modifies van Hasselt's double-Q network (van Hasselt, Guez and Silver, 2015) to use Q-learning in the initial learning stages, to aid optimization, and gradually replaces it with SARSA, as it reduces overestimates and gives a better exploration of the parameter space (Xu et al., 2018).

# 3  Requirements, Analysis and Project Management

Starting from the project aims and objectives presented in Section 1.2, this section presents the project requirements, the tools and methods used for testing and evaluating the results and information concerning the management of the project.

## 3.1  Project Requirements

The primary objective of this project was the reimplementation of a state-of-the-art deep reinforcement learning agent, as close as possible to the version described in literature. Once this was achieved, the investigation focused on creating a series of progressively complex conventional and novel DRL agents. The novel agents were created using the DFA biologically-plausible algorithm. This provided insight into the effectiveness of using DFA in DRL problems and the performance of DFA-based DRL agents when compared with conventional algorithms.

Starting from the objectives presented in Table 1.1, a series of requirements was created for each stated objective. These are presented in Table 3.1.

*Table 3.1: Project Requirements*

| Objective | Requirements |
|---|---|
| Gain an understanding of artificial neural networks and deep learning techniques. | 1. Study general machine learning.<br>2. Study deep learning techniques, particularly deep convolutional neural networks.<br>3. Program both simple and complex CNNs in Python, by using only fundamental packages, such as Numpy, and more complex programming frameworks such Tensorflow and Theano. |
| Investigate reinforcement learning algorithms and how to implement them together with deep neural networks. | 1. Complete courses in general reinforcement learning.<br>2. Create simple games in Python, such as Tic-Tac-Toe or Gridworld, and use RL algorithms, such as TD(0), SARSA and Q-learning, to train agents and understand their functionality. |
| Examine previous studies and the current state of the art by means of a literature review. | 1. Perform a comprehensive literature review of early and recent studies in RL and DL, to understand the underlying principles.<br>2. Examine recent applications of DRL. |
| Develop an AI Agent by using a Deep Reinforcement Learning algorithm. | 1. Complete a course in deep reinforcement learning.<br>2. Using Python, program progressively complex DRL algorithms using programming frameworks to gain an intuitive understanding of how the algorithms work.<br>3. Using different Python packages, program the two conventional DRL algorithms (DQN and |
| | DSN) and the two novel methods using DFA for calculating the gradients. |
| | 4. Tune the agent's hyperparameters by using simpler games which are less computationally demanding. |
| Train the AI Agent to play a selected game by using only sensory inputs. | 1. Select a RL learning environment and adapt the previously created agents to use it. |
| | 2. Use a series of simple games to test the network and fine tune the its hyperparameters. |
| | 3. Test the agents using a more complex game. |

## 3.2 Analysis and Evaluation

After being created, the AI agents used the OpenAI Gym's Atari environment for training and evaluation, as it contains the Arcade Learning Environment (Bellemare et al., 2013) used in the original papers by Mnih and Zhao (Mnih *et al.*, 2013; Zhao *et al.*, 2017). The emulator passed the agent a video input, representing the environment states, and the game scores, or the reward, which was used in training the CNN predicting the action-value functions.

During training, agent progress was measured using methods similar to those described by Mnih and Zhao in their respective papers (Mnih *et al.*, 2013; Zhao *et al.*, 2017). Information about the agent-obtained rewards was collected, both during and after training, for evaluation in line with the objectives outlined in Table 1.1. The main evaluation criteria are presented in Table 3.2.

*Table 3.2: Evaluation Criteria*

| Evaluation Criteria | Methodology |
|---|---|
| Training performance | - Data concerning the agent's total acquired reward is collected for each training episode; <br> - The running average is calculated for the obtained data and plotted over the number of training episodes; |
| Agent performance during training | - The training process is paused after an equal number of training episodes, and the agent parameters are frozen; <br> - The agent plays 100 evaluation episodes, storing the total acquired reward for each episode; <br> - The mean and standard deviation are calculated and plotted over the number of training episodes; |
| Agent performance after training | - At the end of the training process, the agent's parameters are frozen; <br> - The agent plays 1000 evaluation episodes, storing the total acquired reward for each episode. The number of episodes ensures that the obtained results are statistically significant; <br> - The mean and standard deviation are calculated and plotted; <br> - Tests for normality are performed on the obtained results to ensure that the rewards are not stochastic, including calculating the mean, median, skewness, kurtosis factor and plotting the |
| | data distribution using a Q-Q and bar plot (further details in Appendix B); |
| Agent comparison | - The various agents used for the current investigation are assessed by plotting their mean distribution and standard deviation;<br>- If multiple agents have close performance, the Wilcoxon signed rank test is used to determine if the two datasets are significantly different (further details in Appendix B);<br>- The obtained results are then compared with similar results from literature, for both AI and human players;<br>- The score of a human player (the author) was introduced for the Atari agents, calculated as the mean over 30 games after 2 hours of training, in line with Mnih's approach (Mnih *et al.*, 2015). |

## 3.3 Risks and Mitigations

As with any investigation of complex novel algorithms, this study could have been impacted by a set of challenges. A series of risks, their descriptions and proposed mitigations are presented in Table 3.3.

*Table 3.3: Potential Risks and Proposed Mitigations*

| Risk Name | Risk Description | Proposed Mitigation |
|---|---|---|
| Time Constraints | Understanding the field of deep and deep reinforcement learning represents a time-consuming process, particularly considering that the author has limited knowledge and experience of the field. Also, developing, training, tuning, testing and validating progressively complex deep reinforcement learning algorithms represents a time-consuming process. | Careful planning has been put in place to ensure enough time is left for the author to acquire the necessary knowledge and create the relevant algorithms while ensuring that valid experimental data is obtained. Also planning considers other academic commitments that the author has. |
| Processing Power Constraints | DRL models are very computationally expensive in terms of required processing power. Moreover, if access to an HPC cluster can be obtained, high space allocation demands at these cluster can cause large queue waiting times. | Access has been secured to use The University of Sheffield's HPC cluster. A strategy has also been put in place, where simple codes and the prototypes of the complex codes are run on a local machine to ensure that they are bug-free. Also, enough time is allotted to batch jobs to ensure that their execution does not overrun, causing the HPC to terminate them by triggering a KILL signal. Finally, batch jobs have been split into smaller |
| | | jobs, constantly saving their results and parameters, ensuring that even if a KILL signal is triggered, not all the data, progress and results are lost. |
| Storage Constrains | DRL models are expensive in terms of the required RAM memory. | Enough RAM memory is requested on the HPC to prevent it from triggering a KILL signal, terminating the jobs for overrunning the allocated memory. |
| Results or Training Data Inconsistencies | Due to the stochastic nature of DRL algorithms, variations can appear in the training performance or the obtained results, especially if, during training, the algorithms remain stuck in a local minimum or maximum. | Several techniques have been employed to prevent this from occurring, including advanced initialization methods for the parameters and training optimisation algorithms. Care was given during the hyperparameter training to identify potential values which might lead to instability during training. Also, a statistically significant number of data points where collected during evaluation (1000) so that any outlier data points do not impact the overall measured performance of the algorithms. |
| Unexpected Performance or Results | Due to the novel nature of some of the algorithms, they might not perform as well as expected or they could be difficult to implement in code using the available programming frameworks. | The employed methodology ensures that progressively complex algorithms are developed and tested, which would provide an early warning if there are any issues with the novel agents. Then, multiple versions of the novel algorithms have been tested across different games, which gives a better indication of the agent's overall performance. Finally, provisions have been put in place to continue the project in case results are not obtained by the allotted deadline. |

## 3.4 Hardware Considerations

DNNs and CNNs are typically computationally expensive processes. This is because their parameters, which are represented by matrices of randomly initialized numbers, are trained across multiple iterations through matrix multiplications and additions. This process allows the networks to lean the required highly abstract features used for solving tasks.

To increase the time efficiency of data generation, three approaches were used for this project. Firstly, simple algorithms and prototypes of the more complex algorithms were run on the local machine used for writing the codes. Then, the DNN codes used as prototypes and for solving classical control problems were trained and evaluated using server-grade CPUs on an HPC. This was due to their large availability, which allowed multiple simulations to be run in parallel, which proved particularly useful during the agents' hyperparameter tuning, when 408 simulations were run in series and parallel. Finally, the more complex Atari 2600 agents, which use deep CNNs, were trained and evaluated using GPUs. This is due to the ability of GPUs to perform very large amounts of matrix multiplications in parallel, which allowed training times to be reduced by an order of magnitude compared to CPUs. To gain an intuition of why this approach was preferred, the CartPole game utilized with a DNN has an input state space of 4 inputs and takes approximately 1 hour to train for 500 episodes on a CPU. The input state space of an Atari game such as Breakout is a box of dimensions $210x160x3$. This is a modest size screen, but at the same time it is equivalent to 33,600 inputs, which is 8400 times more than CartPole.

All simulations were conducted on ShARC, the largest HPC cluster at The University of Sheffield. Access was also granted by RSE Sheffield to use research-group specific ShARC nodes, including the Nvidia® DGX-1™ deep learning super computer, which further increase the projects time efficiency. Hardware details are presented in Appendix C.

## 3.5   Project Management

For the first part of this investigation the focus has been on developing a better understanding of deep and reinforcement learning techniques, as well as of novel DRL algorithms and the mathematical methods employed by them. As suggested by the Gantt Chart in Appendix D, all these tasks have successfully been completed ahead of schedule.

The focus for the second part of the academic year consisted in the implementation of the derived methods in order to develop progressively more complex DRL algorithms, in accordance with Table 3.1. For training and evaluation, access to better hardware was secured, as the methods proposed for this investigation are computationally demanding. The quick development of some of the simple algorithms has allowed the agent training and evaluation to start earlier than intended. However, the development of some of the more complex novel agents has been impaired by issues with the code implementation, which means that not all intended AI agents have been coded by the writing of this report. Provisions have been made to ensure that work on these agents will continue into the summer.  The literature review has also continued beyond the original plan, as several new papers relevant to the project have been published in the early months of 2019.

# 4  Methods

This section presents the numerical methods and techniques used in developing the DRL algorithms required for this investigation. Knowledge about the mathematical methods and employed techniques is crucial for both creating the algorithms and understanding the obtained results.

## 4.1  Deep Learning

### 4.1.1  The Neuron

Neural networks represent a computing system composed of many highly interconnected simple functions, represented by processing elements called neurons (Caudill, 1987; Ng et al., 2019) (Figure 4.1).

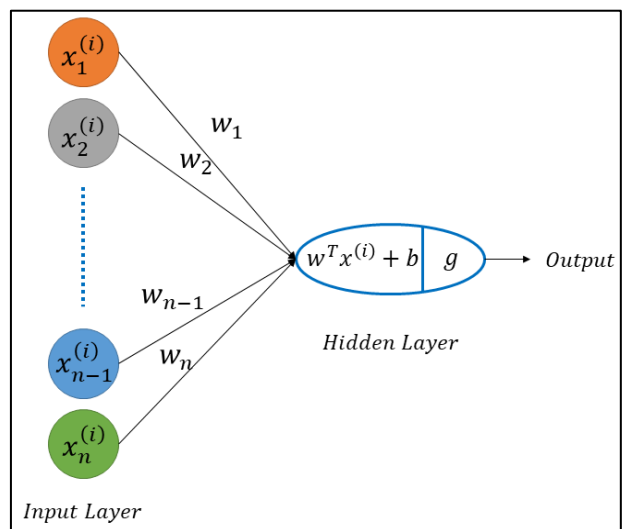

*Figure 4.1: Schematic of a single neuron*

A neuron is a simple computational unit that receives several inputs, computes an affine transformation for these inputs (4.1.1) and then applies an element-wise non-linear activation function (Goodfellow, Bengio and Courville, 2016). The neuron outputs either one or a set of non-linear activations, proportional to its degree of activation. These allow the network to derive complex features, linking the inputs and outputs. There are numerous types of activation functions, including sigmoid (4.1.2), tanh (4.1.3) and ReLU (4.1.4) (Figure 4.2). The ReLU function is currently the most used activation for deep neural networks, as it is less computationally expensive than the other functions while solving the vanishing gradient issue (Goodfellow, Bengio and Courville, 2016).

$$z = \sum_{i=1}^{n} w_i x_i + b \qquad (4.1.1)$$

$$g(z) = \frac{1}{1 - e^{-z}} \qquad (4.1.2)$$

$$g(z) = \tanh(z) = \frac{e^z - e^{-z}}{e^z + e^{-z}} \qquad (4.1.3)$$

$$g(z) = \max(0, z) \tag{4.1.4}$$

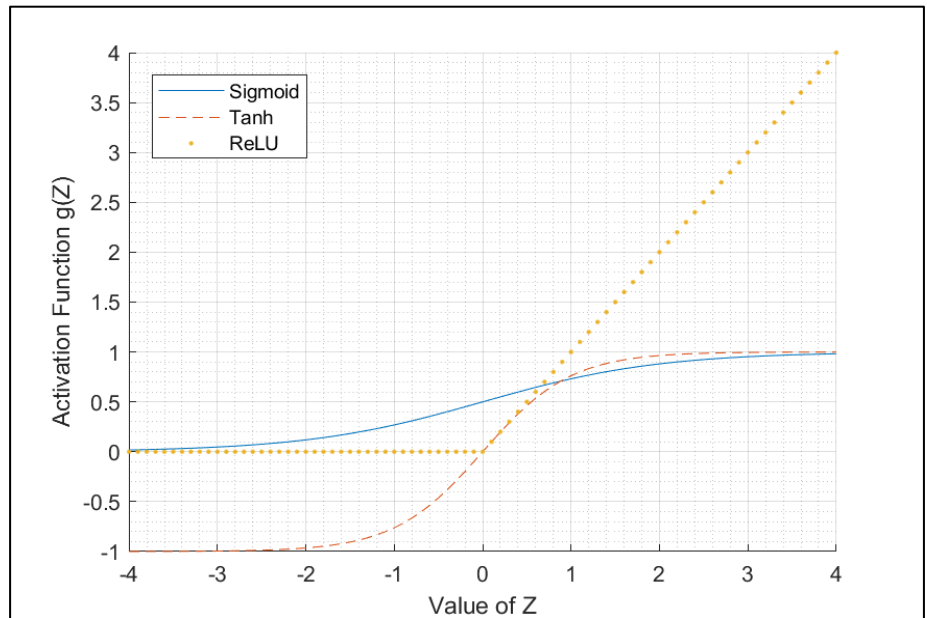

*Figure 4.2: Comparison of different activation functions*

## 4.1.2 Deep Feed-Forward Neural Networks

By stacking together multiple neurons across multiple hidden layers, in such a way that neurons within a layer are interconnected with neurons from both the previous and the next layers, a deep neural network can be obtained (Figure 4.3). During forward propagation (4.1.5)-(4.1.6) are true for a layer $l$ of a DNN. An intuitive explanation of DNNs is presented in Appendix A.

$$Z^{[l]} = W^{[l]} \Lambda^{[l-1]} + b^{[l]} \tag{4.1.5}$$

$$\Lambda^{[l]} = g^{[l]}(Z^{[l]}) \tag{4.1.6}$$

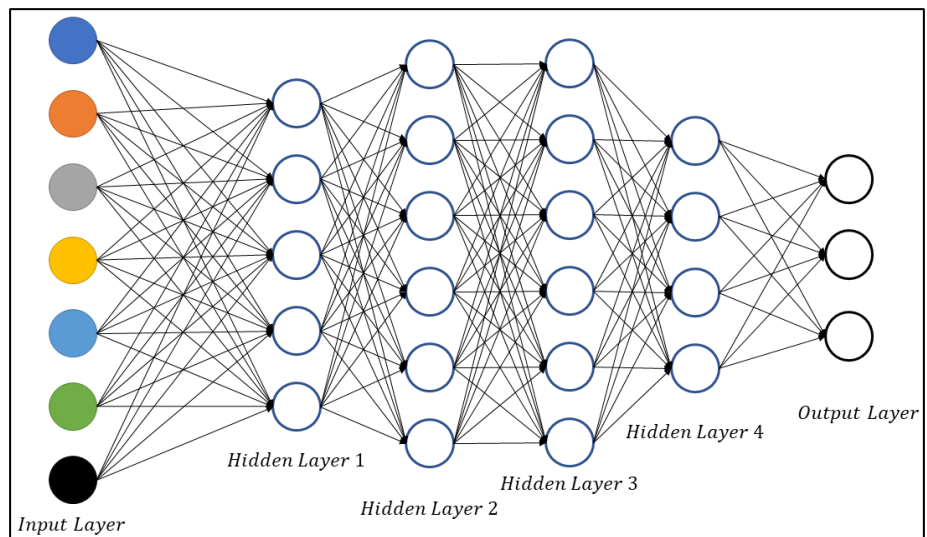

*Figure 4.3: Example of a Deep Feed-forward Neural Network.*

### 4.1.3 Weight and Bias Initialization

Before training the network, both weights and biases need to be initialised. Weights are usually initialized randomly from a normal uniform distribution, while biases are typically initialised to zero. The random initialisation of the weights creates asymmetry at initialisation, which prevents gradients from exploding or dissolving too quickly. This issue can be further mediated by resampling those weights which have values more than two standard deviations away from the mean. This is known as sampling from a truncated normal distribution. Furthermore, the issue of vanishing or exploding gradients can be addressed by means of heuristic techniques, such as the Xavier initialisation (Glorot and Bengio, 2010). This approach aims at keeping the scale of the gradients roughly the same in all layers by making the variances equal across the network (4.1.7).

$$W^{[l]} = W^{[l]}_{random\ normal} \cdot \sqrt{\frac{2}{size^{[l-1]} + size^{[l]}}} \tag{4.1.7}$$

### 4.1.4 Gradient Computing

To determine how different the network's predicted output is from the intended output, a cost function is usually calculated by using the cross-entropy between the two value sets. The form of this cost function depends on the specific form of the network and the model generated distribution $P_{model}$ (4.1.8) (Goodfellow, Bengio and Courville, 2016). One commonly used loss function is the mean squared error function (4.1.9).

$$J(\theta) = -\mathbb{E}_{x,y \sim \hat{P}_{data}} \log P_{model}(\boldsymbol{y}|\boldsymbol{x}) \tag{4.1.8}$$

$$J(\theta) = \frac{1}{n} \sum_{i=1}^{n} (y_i - \hat{y}_i)^2 \tag{4.1.9}$$

The goal of the network is to learn by minimizing $J(\theta)$ in the direction of steepest descent, thus improving its prediction accuracy. Gradient descent (Figure 4.4) is governed by the learning rate $\alpha$, which governs how big the steps towards the local minimum are. This influences how fast and smooth the network will converge. $J(\theta)$, or $J(\boldsymbol{W}, \boldsymbol{b})$, is a function of the products and transformations of the various $W^{[l]}_{i,j}$ and $b^{[l]}_{i,j}$ belonging to each of the neurons in the DNN. Thus, the derivative of $J(\theta)$ with respect to each parameter provides and indication of the gradient by which these parameters need to change for the network to converge.

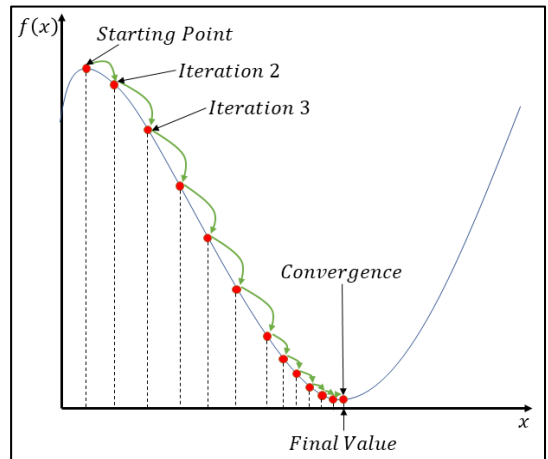

*Figure 4.4: Schematic representation of Gradient Descent*

Dynamic programming algorithms, such as BP or DFA, are used in calculating the gradients of different components, which helps in training the network.

### 4.1.4.1 Backpropagation

The backpropagation algorithm (Rumelhart, Hinton and Williams, 1986) works by backward propagating the error from the output layer through the reciprocal connections of the network, allowing the gradient calculation (Figure 4.5). The gradients for a hidden layer $l$ are calculated using (4.1.10)-(4.1.14).

$$dZ^{[l]} = d\Lambda^{[l]} \, \Delta g^{[l]}(Z^{[l]}) \tag{4.1.10}$$

$$dW^{[l]} = dZ^{[l]} \, \Lambda^{[l-1]} \tag{4.1.11}$$

$$db^{[l]} = dZ^{[l]} \tag{4.1.12}$$

$$d\Lambda^{[l-1]} = W^{[l]T} dZ^{[l]} \tag{4.1.13}$$

$$dZ^{[l]} = (W^{[l+1]T} dZ^{[l+1]}) \, \Delta g^{[l]}(Z^{[l]}) \tag{4.1.14}$$

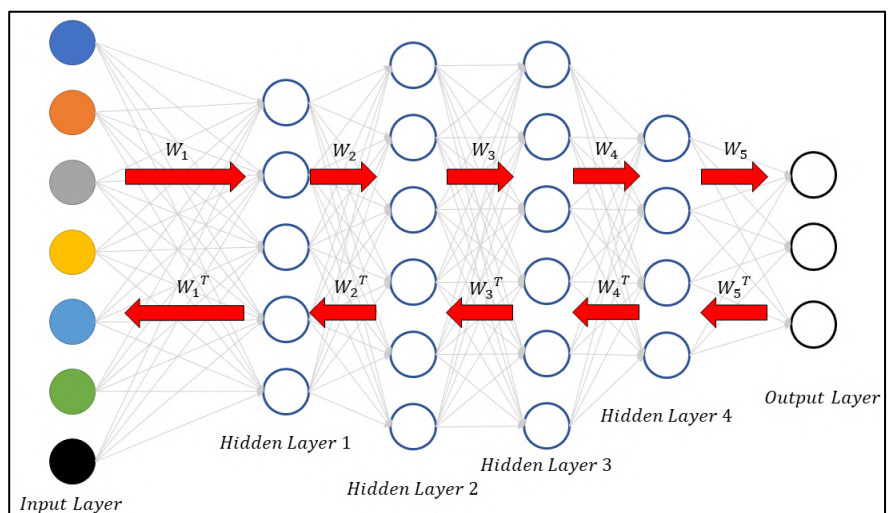

*Figure 4.5: Feed-forward and feed-backward paths in Backpropagation*

### 4.1.4.2  Direct Feedback Alignment

The direct feedback alignment algorithm (Nøkland, 2016) uses different fixed random weights and propagates the error directly from the output layer to each hidden layer (4.1.15). This provides a different update direction to backpropagation (Figure 4.6).

$$dZ^{[l]} = \left(B \cdot dZ^{[out]}\right) \Delta g^{[l]}\left(Z^{[l]}\right) \tag{4.1.15}$$

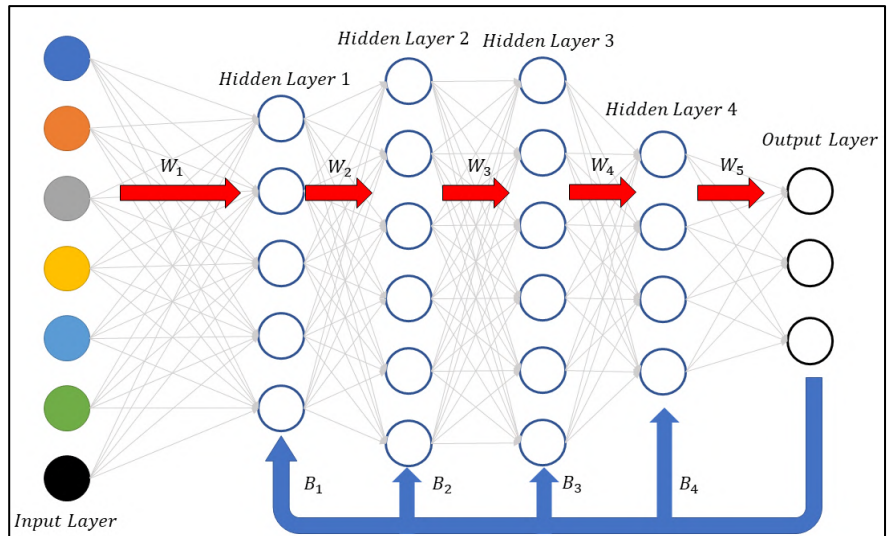

*Figure 4.6: Feed-forward and feed-backward paths in Direct Feedback Alignment*

### 4.1.5  Optimisation Algorithms

The calculated gradients are used in updating the network parameters in the direction of steepest descent (4.1.16)-(4.1.17), thus minimising the DNN error. (Goodfellow, Bengio and Courville, 2016).

$$W^{[l]} = W^{[l]} - \alpha \cdot \frac{\partial J(W,b)}{\partial W^{[l]}} = W^{[l]} - \alpha \cdot dW^{[l]} \tag{4.1.16}$$

$$b^{[l]} = b^{[l]} - \alpha \cdot \frac{\partial J(W,b)}{\partial b^{[l]}} = b^{[l]} - \alpha \cdot db^{[l]} \tag{4.1.17}$$

Optimisation algorithms modify (4.1.16)-(4.1.17) in order to improve training speed and performance. RMSprop (Hinton, 2014), which is used in this investigation, uses data minibatches and calculates a weighted average of derivative squares for each time step, changing the learning rate. By using an exponentially weighted moving average (4.1.18)-(4.1.19), RMSprop ensures that older parameter values have a smaller influence than newer values. This allows it to adjust the learning rate according to the magnitude of the gradients.

$$\Psi_{dW}^{[l]} = \beta \, \Psi_{dW}^{[l]} + (1-\beta) \, dW^{[l]^2} \tag{4.1.18}$$

$$\Psi_{db}^{[l]} = \beta \, \Psi_{db}^{[l]} + (1-\beta) \, db^{[l]^2} \tag{4.1.19}$$

$$W^{[l]} = W^{[l]} - \alpha \cdot \frac{dW^{[l]}}{\sqrt{\Psi_{dW}^{[l]} + \epsilon}} \tag{4.1.20}$$

$$b^{[l]} = b^{[l]} - \alpha \cdot \frac{db^{[l]}}{\sqrt{\Psi_{db}^{[l]} + \epsilon}} \tag{4.1.21}$$

### 4.1.6 Batch Sizes

This represent an indication of the number of training examples fed into the DNN during one training step. The size of the batch has an impact on the network error, and thus its training performance. Thus, the batch size can be treated as a hyperparameter and requires tuning, however for the purpose of this investigation it is kept fixed at 32 for all simulations.

### 4.1.7 Deep Convolutional Neural Networks

Digital images are typically represented as 3D matrices, each pixel containing 3 numbers between 0 and 255 representing the intensity of the red, green and blue light channels. Convolutions are mathematical operations used by DNNs to extract features from data with a grid-like structure, such as images. Their output $s$ denotes how the shape of the input function $f$ is modified by the kernel function $g$ (4.1.22).

$$s(t) = (f * g)(t) = \int_{-\infty}^{\infty} f(\tau)g(t - \tau)d\tau \tag{4.1.22}$$

If the input and kernel functions are tensors, and $t$ is an integer, (4.1.22) becomes (4.1.23). If the input tensor is a 2D image $I$ which is convolved with a kernel filter $k$, and considering the properties of convolutions, (4.1.23) becomes (4.1.24) (Figure 4.7) (Goodfellow, Bengio and Courville, 2016).

$$s(t) = (f * g)(t) = \sum_{\tau = -\infty}^{\infty} f(\tau)g(t - \tau) \tag{4.1.23}$$

$$s(t) = (k * I)(i, j) = \sum_m \sum_n I(i - m, j - n)k(m, n) \tag{4.1.24}$$

Like the neurons in a DNN, each filter (kernel) has a series of weights $W$ and biases $b$. An image is convolved with a filter by element-wise multiplication. All convolved elements are then summed up and a bias is added, after which they are stacked to obtain a new 3D volume (Figure 4.8), to which an activation function is applied. This helps the network detect different features, such as edges, curves or shapes. In a CNN, several parallel convolutions can be used to produce a set of linear activations. Also, multiple convolutional layers in series can help the network generate progressively more complex features (Figure 4.9). The dimensions of a convolved element are given by (4.1.25). The filter size represents the dimension of the convolution filter, the stride shows the amount by which the filter shifts, and padding represents the depth of the padding layer (Figure 4.10). Further explanations of the composing elements of a CNN can be found in Table 2.1.

$$dimension_{new} = \frac{dimension_{old} - filter\ size + 2 * padding}{stride} \quad (4.1.25)$$

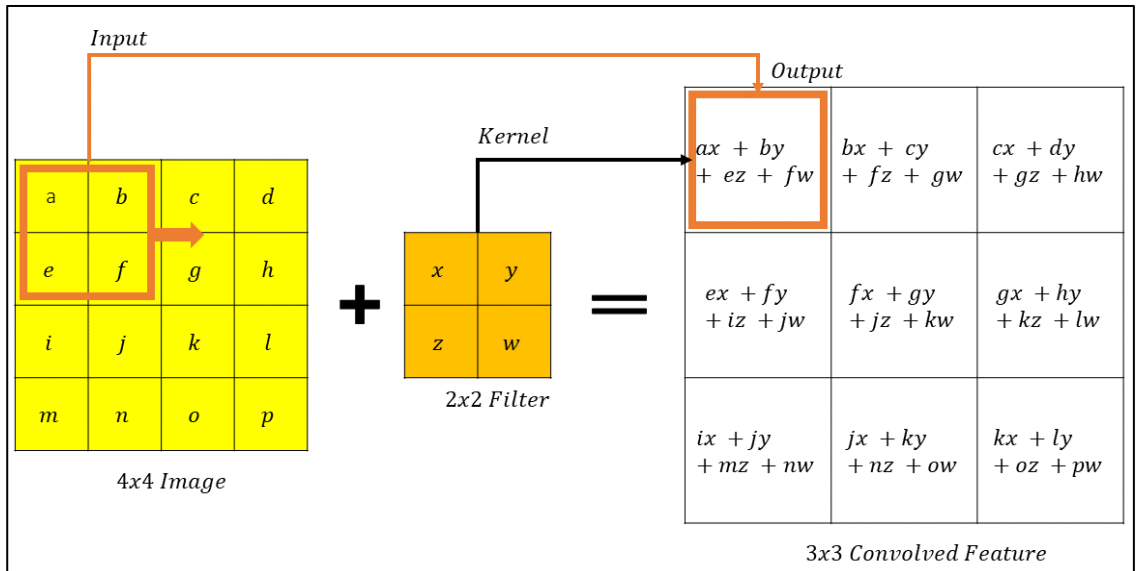

Figure 4.7: Convolution operation for a 4x4 image

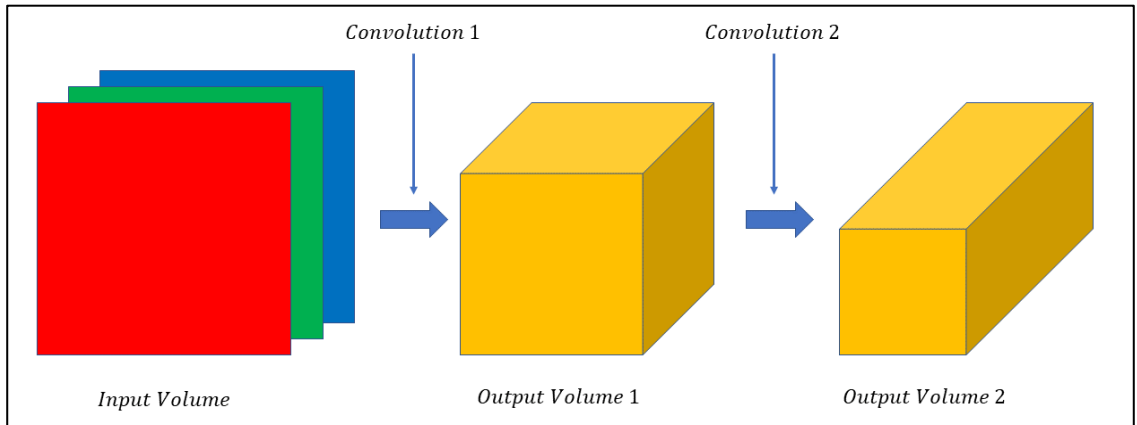

Figure 4.8: New output volumes obtained using the convolution operation

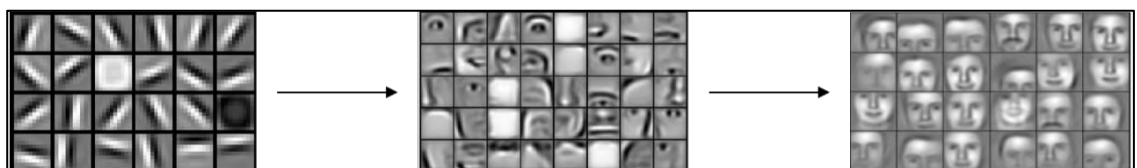

Figure 4.9: Progressively complex features generated by a CNN (Ng, 2019a)

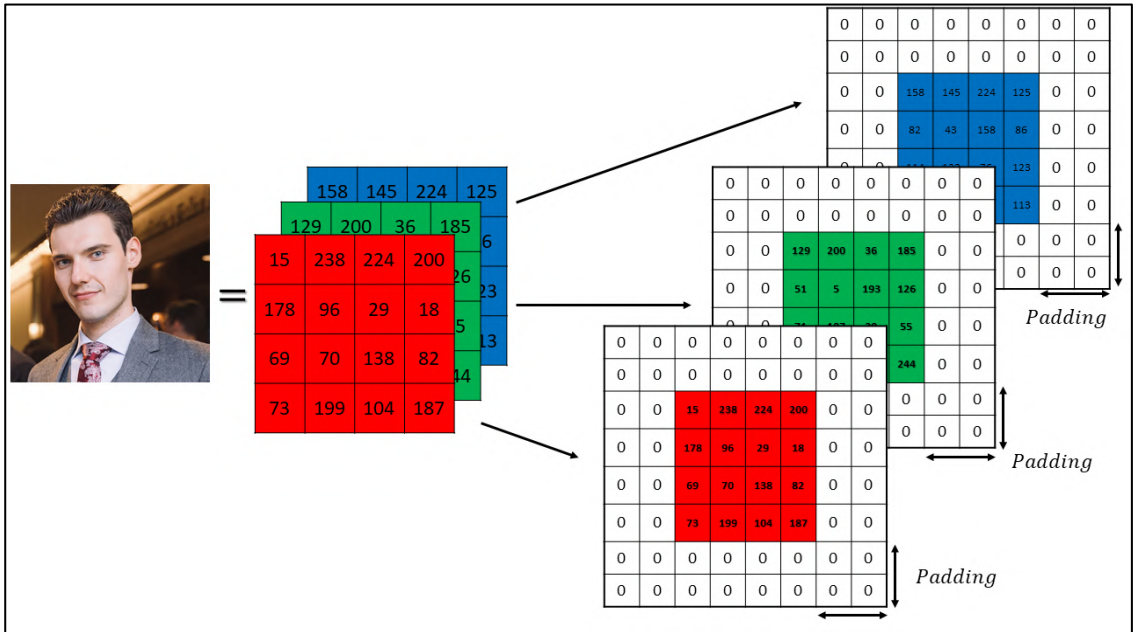

*Figure 4.10: Padding added to all the layers of an input image to preserve its feature*

During a backward pass through the network, the gradient of the activation $\Lambda$, the filter weights $W$ and bias $b$ are given by (4.1.25)-(4.1.27). The weight and bias need to be summed over all the slices.

$$d\Lambda \mathrel{+}= \sum_{i=0}^{h} \sum_{j=0}^{w} W \times dZ_{ij} \qquad (4.1.25)$$

$$dW_{slice} \mathrel{+}= \sum_{i=0}^{h} \sum_{j=0}^{w} \Lambda_{slice} \times dZ_{ij} \qquad (4.1.26)$$

$$db_{slice} \mathrel{+}= \sum_{i=0}^{h} \sum_{j=0}^{w} dZ_{ij} \qquad (4.1.27)$$

## 4.1.8 Pooling

The pooling layer works by sliding a square window over the output of a CONV layer and storing either the average or maximum value. Although it has no parameters to train, during a backward pass it allows the gradients to backpropagate by using a mask function which captures the location of the maximum value of the error or distributes its value equally, depending on the pooling layer type.

## 4.1.9 Fully Connected Layer

The final layers of a CNN are usually like those found in conventional DNNs and are known as fully connected layers. These layers are a flatten version of the final convolution layer, with an equal number of elements.

## 4.2 Reinforcement Learning

### 4.2.1 The Markov Decision Process

When interacting with the environment, RL agents record their states, actions and goal. States are the specific configuration of the environment, sensed by the agent through its inputs and influenced by its actions. The goal of an agent is to maximise the amount of reward, which is user defined, that it receives from the environment as feedback to its actions. Negative rewards, for example, can be used to incentivise an agent to perform optimally by penalising it. The total measure of all possible future rewards of an action is defined as the value. All these concepts are formalised in the MDP framework (Figure 4.1), a tuple consisting of states, actions, rewards, state-transition probabilities and reward probabilities (Sutton and Barto, 2018) with specific dynamics (4.2.1).

$$p(s', r \mid s, a) = \Pr\{S_t = s', R_r = r \mid S_{t-1} = s, A_{t-1} = a\} \qquad (4.2.1)$$

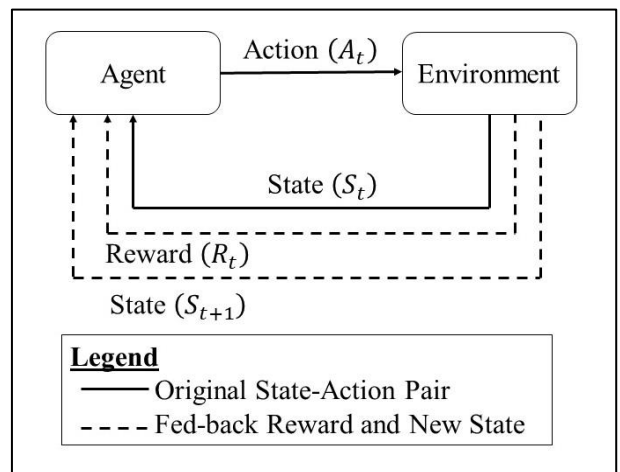

*Figure 4.11: The interaction between an Agent and its Environment in an MDP*

An agent's goal is formalised as the sum of all expected returns, discounted by a factor $\gamma$ (typically equal to 0.9), which accounts for the time-related prediction difficulties.

$$G_t = \sum_{i=0}^{\infty} \gamma^i R_{t+i+1} \qquad (4.2.2)$$

The decision component of an MDP is given by the policy $\pi$, depicting the probability with which actions are selected when in a state.

### 4.2.2 The Explore-Exploit Dilemma

RL agents need to collect enough data to accurately estimate the value $q$ of their actions at each timestep without acting sub-optimally for too long by over-exploring their environment.

$$q_t(a) = \mathbb{E}[R_t \mid A_t = a] \qquad (4.2.3)$$

Epsilon-greedy represents a solution to the explore-exploit dilemma. It has agents explore their environment by selecting random actions for an $\varepsilon$ portion of time.

Otherwise, agents act greedily, exploiting the highest value action without considering any long-term effects (4.2.4).

$$A_t = argmax_a Q_t(a) \tag{4.2.4}$$

Another solution to the explore-exploit dilemma is given by the softmax algorithm (4.2.5). Compared to epsilon-greedy, which chooses equally among all actions, which could lead to instability, softmax ranks and weighs the actions based on their estimated value. Sutton and Barto find that by varying the temperature $\xi$, the policy could approach determinism. Nevertheless, for this investigation, it is assumed that $\xi = 1$, as varying it with time would require prior insight into the real action-values (Sutton and Barto, 2018).

$$\pi(a|s) = \frac{e^{\frac{Q_\pi(a)}{\xi}}}{\sum_{i=1}^n e^{\frac{Q_\pi(i)}{\xi}}} \tag{4.2.5}$$

Most RL problems are non-stationery, meaning that environment statistics are not constant over time. The value of actions in these cases can be estimated by averaging the rewards after multiple selections of an action (4.2.6)-(4.2.9). It was found that value functions do not converge for a constant $\alpha = 1/n$ learning rate, yet this is generally not an issue (Sutton and Barto, 2018).

$$Q_{n+1} = \frac{1}{n} \sum_{i=1}^n R_i \tag{4.2.6}$$

$$Q_{n+1} = \frac{1}{n} \left( R_n + \sum_{i=1}^{n-1} R_i \right) = \frac{1}{n} \left[ R_n + (n-1) \frac{1}{n-1} \sum_{i=1}^{n-1} R_i \right] \tag{4.2.7}$$

$$Q_{n+1} = \frac{1}{n} [R_n + (n-1)Q_n] \tag{4.2.8}$$

$$Q_{n+1} = Q_n + \alpha (R_n - Q_n) \tag{4.2.9}$$

### 4.2.3  Solving the Markov Decision Process

Given a state and a policy, an MDP can be solved by means of state and action-value functions. The state-value function estimates the expected return when an agent is in each state $s$ (4.2.10), while the action-value function estimates the expected return when the agent also takes an action $a$ (4.2.11).

$$V_\pi(s) = \mathbb{E}[G_t \mid S_t = s] \tag{4.2.10}$$

$$Q_\pi(s, a) = \mathbb{E}[G_t \mid S_t = s, A_t = a] \tag{4.2.11}$$

Considering (4.2.7)-(4.2.9), the fact that $\pi = \pi(a|s)$, the properties of the expectation equation and the law of total expectation, and replacing them in (4.2.10), the Bellman Equation for $V_\pi$ is obtained (4.2.15).

$$V_\pi(s) = \mathbb{E}\left[\sum_{i=0}^{\infty} \gamma^i R_{t+i+1} \mid S_t = s\right] \tag{4.2.12}$$

$$V_\pi(s) = \mathbb{E}[R_{t+1} + \gamma G_{t+1} \mid S_t = s] \tag{4.2.13}$$

$$V_\pi(s) = \sum_a \pi(a|s) \sum_r \sum_{s'} p(s', r \mid s, a) \left\{r + \gamma \mathbb{E}_\pi\left[\sum_{i=0}^{\infty} \gamma^i R_{t+i+2} \mid S_t = s'\right]\right\} \tag{4.2.14}$$

$$V_\pi(s) = \sum_a \pi(a|s) \sum_r \sum_{s'} p(s', r \mid s, a) [r + \gamma V_\pi(s')] \tag{4.2.15}$$

The Bellman Equation for $Q_\pi$ is calculated similarly (4.2.16).

$$Q_\pi(s, a) = \sum_a \pi(a|s) \sum_r \sum_{s'} p(s', r \mid s, a) [r + \gamma Q_\pi(s', a)] \tag{4.2.16}$$

The Bellman equations are recursive structures, with the value functions at the current step depending on the value functions at the next step. They can be solved by determining the policy which generates the most value, known as the optimal policy, which allows the optimal value functions to be calculated.

$$V_*(s) = \max_\pi \{V_\pi(s)\} \tag{4.2.17}$$

$$Q_*(s, a) = \max_\pi \{Q_\pi(s, a)\} \tag{4.2.18}$$

$Q_*(s)$ gives the optimal return for the given state-action pair. By replacing (4.2.11) in (4.2.13), and then in (4.2.18), $Q_*(s)$ can be defined recursively in terms of $V_*(s)$, relating the state and action-value functions (Sutton and Barto, 2018).

$$Q_*(s, a) = \mathbb{E}[R_{t+1} + \gamma V_*(s) \mid S_t = s, A_t = a] \tag{4.2.19}$$

Under optimality conditions, the expected return of the state-value function is the same as the expected return for performing the optimal action. Thus, (4.2.17) is rewritten as (4.2.20). Replacing (4.2.19) in (4.2.20) yields the Bellman Optimality Equation for $V_*(s)$ (4.2.21). The equivalent equation for $Q_*(s)$ is (4.2.22)

$$V_*(s) = \max_a Q_*(s, a) \tag{4.2.20}$$

$$V_*(s) = \sum_r \sum_{s'} p(s', r \mid s, a) [r + \gamma V_*(s')] \tag{4.2.21}$$

$$Q_*(s,a) = \sum_r \sum_{s'} p(s',r \mid s,a) \left[ r + \gamma \max_{a'} Q_*(s,a) \right] \tag{4.2.22}$$

The optimality equation can be solved by selecting the actions which yield the best values for the future state. For $V_*(s)$ a lookahead search is required to determine all future state-action pairs. For $Q_*(s)$ this is done automatically, as it already looks at future actions maximizing the value function for a given state.

## 4.2.4  Temporal Difference Learning

In the MDP framework, an agent needs to solve two problems: the prediction problem, of finding a value function given a policy, and the control problem, of finding the optimal policy which maximises its expected rewards.

Agents trained with the TDL technique are capable of learning directly from raw experiences, without needing a full model of the environment's dynamics, and updating their parameter estimates based on other learned estimates, which is known as bootstrapping. This technique is fully online, as values are updated during a learning episode rather than at its end. Thus, the state value is updated by using the average return for that state, obtained by modifying (4.2.7).

$$V(s_t) \leftarrow V(s_t) + \alpha \left[ G_t - V(s_t) \right] \tag{4.2.23}$$

From (4.2.10) and (4.2.15), $G_t$ is estimated using bootstrapping (4.2.24). Replacing this in (4.2.23), yields the TD(0) algorithm for solving the prediction problem, which updates $V(s)$ as soon as $s'$ is known.

$$G_t = r + \gamma V(S_{t+1}) \tag{4.2.24}$$

$$V(S) = V(S) + \alpha \left[ R + \gamma V(S') - V(S) \right] \tag{4.2.25}$$

### 4.2.4.1  SARSA

By applying similar principles to the control problem, the action-value function can be continuously updated in-place based on itself (4.2.26). At the end of each episode, the policy can be improved by taking either a greedy action (4.2.4) or a softmax-based action (4.2.5) according to the action-value matrix (4.2.27).

$$Q(S_t, A_t) = Q(S_t, A_t) + \alpha \left[ R_{t+1} + \gamma Q(S_{t+1}, A_{t+1}) - Q(S_t, A_t) \right] \tag{4.2.26}$$

$$A_{t+1}(S_{t+1}) = argmax_A Q(S_t, A_t) \tag{4.2.27}$$

As the argmax function is greedy, it does not provide a full exploration of all state-action pairs. This cannot be achieved using a fully deterministic approach. Still, it can be obtained by using epsilon-greedy with a decaying learning rate (4.2.28), ensuring that all $Q(s,a)$ pairs are updated over an episode. When the optimal policy is found, both the policy and action-value function become constant.

$$\alpha(s,a) = \frac{\alpha_0}{count\ (S,A)} \tag{4.2.28}$$

Thus, for this approach known as SARSA, the agent is updating the action-value function by using actions selected with the in-use policy and alternating between policy evaluation and policy improvement over each episode. This is known as an on-policy algorithm.

### 4.2.4.2  Q-Learning

In contrast to SARSA, agents using the Q-learning algorithm perform random actions for a future state and update their current action-value function by using the maximum value of the $Q$ function for that future state. Thus, they are acting as if they are following a purely greedy policy, although they are not (4.2.29). At the end of each timestep, the agent moves to the next step and selects an action based on the existing policy. Nevertheless, this is not necessarily the same action as the one used during the training state. This approach is an off-policy method, as the agent takes actions randomly and does not follow the in-use policy.

$$Q(S_t, A_t) = Q(S_t, A_t) + \ \alpha\ \left[R_{t+1} + \gamma \max_a Q(S_{t+1}, a)\ - Q(S_t, A_t)\right] \tag{4.2.29}$$

Despite their difference, both SARSA and Q-learning require $Q$ to be estimated. This might be impossible in many applications due to the state space being impractically large. A solution to this issue is using a function approximator to estimate $Q$. NNs represent a type of non-linear function approximator, which is where the DRL concept originates.

# 5 Design of Deep Reinforcement Learning Agents

This section will discuss the methodology employed in developing the RL and DRL algorithms used throughout this investigation. Building on the methods presented in the previous section, this section will also present how the RL and DRL agents are constructed, as well as discuss several implementation details. The pseudocode of all the agents is presented. The codes are freely accessible on GitHub: https://github.com/aroibu1/uos_drl

## 5.1 Design Methodology

Considering the advanced nature of the investigated topic, the limited time resources available for the implementation and to ensure that enough quality results are obtained, the Rapid Application Development (RAD) methodology was selected. This methodology allows rapid prototyping and iterative feedback-based improvements of the different algorithms. The process employed in designing the various agents is presented in Figure 5.1

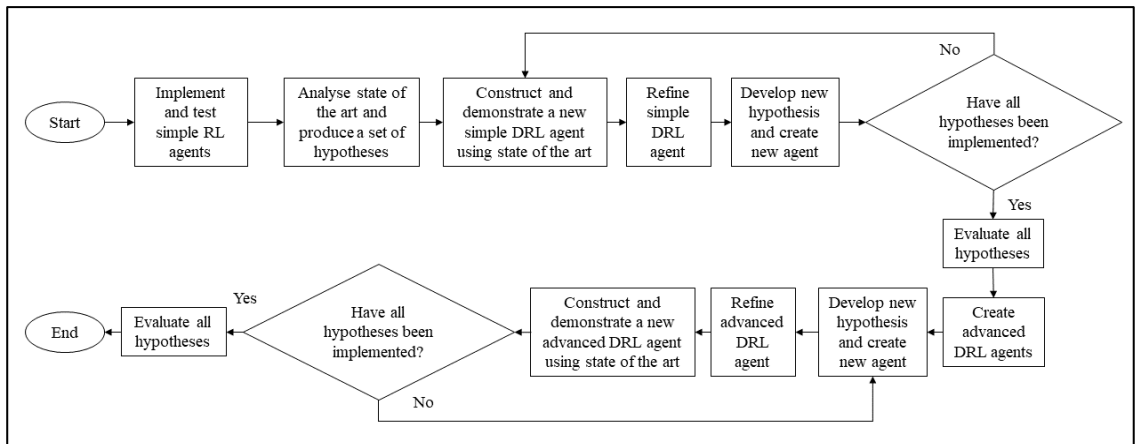

*Figure 5.1: Schematic of RAD-based process utilised during this study*

To understand the differences between the different RL algorithms, the investigation was started with the creation and implementation of two simple RL agents. Once these were complete, the study progressed to more complex, DRL agents. For the purpose of this investigation, the approach used by Mnih in (Mnih *et al.*, 2015) was used in creating the DRL algorithms. Two sets of DRL algorithms were created: a simple set, using a conventional DNN, used to test the hypotheses, followed by a set of complex, CNN-based agents, designed to perform at a level comparable with the state-of-the-art. Experiments were run based on feedback from previous runs, which allowed the agents to be created and then refined through hyperparameter tuning.

## 5.2 Simple SARSA and Q-learning

As described in Figure 5.1, the first step of the investigation consisted in creating two simple RL agents, using SARSA (4.2.26) and Q-learning (4.2.29). This was done in order to capture the specific behaviour of the two algorithms, when tested on simple games.

Both the Q-learning (Algorithm 1) and SARSA (Algorithm 2) agents follow a similar structure, with the implementation differences being in the main for-loop. Neither agent uses a neural network, as the Q-function can be calculated directly, in a tabular manner, due to the small size of the games.

During implementation, all $Q(s, a)$ pairs are initialized to 0 and updated based on the outputs obtained from the games. The policy employed by these agents is epsilon-greedy with $\varepsilon$ decaying with time in order to allow the algorithms to converge. The learning rate $\alpha$ is also set to decay adaptively for every $(s, a)$ pair by dividing it by the number of times that the pair has been updated. The parameters governing these approaches are hyperparameters and require tuning.

---

**Algorithm 1: Simple Q-learning**

---

Initialize action-value function $Q$ to 0 for all $(s, a)$ pairs
Initialize the count dictionary for the adaptive $\alpha$ to 1 for all $(s, a)$ pairs
**For** episode $= 1, M$ **do**
      Initialize sequence $s_1 = \{initial\ state\}$
      Every 100 episodes set $t_\varepsilon = t_\varepsilon + 0.01$
      **For** $t = 1, T$ **do**
            Set $\varepsilon = {0.5}/{t_\varepsilon}$
            With probability $\varepsilon$ select a random action $a_t$
            Otherwise select $a_t = argmax_a Q(s_1, a)$
            Execute $a_t$ and observe the reward $r_t$ and new image $I_{t+1}$
            Set $\alpha = {1}/{count(s_t, a_t)}$
            With probability $\varepsilon$ select a random action $a'$
            Otherwise select $a' = argmax_a Q(s_t, a)$
            Update $Q(s_t, a_t) \mathrel{+}= \alpha \left[ r_{t+1} + \gamma \max_a Q(s_{t+1}, a_{t+1}) - Q(s_t, a_t) \right]$
      **End For**
**End For**

---

---

**Algorithm 2: Simple SARSA**

---

Initialize action-value function $Q$ to 0 for all $(s, a)$ pairs
Initialize the count dictionary for the adaptive $\alpha$ to 1 for all $(s, a)$ pairs
**For** episode $= 1, M$ **do**
      Initialize sequence $s_1 = \{initial\ state\}$
      With probability $\varepsilon$ select a random action $a_1$
      Otherwise select $a_1 = argmax_a Q(s_1, a)$
      Every 100 episodes set $t_\varepsilon = t_\varepsilon + 0.01$
      **For** $t = 1, T$ **do**
            Set $\varepsilon = {0.5}/{t_\varepsilon}$
            Execute $a_t$ and observe the reward $r_t$ and new state $s_{t+1}$
            Set $\alpha = {1}/{count(s_t, a_t)}$
            With probability $\varepsilon$ select a random action $a'$
            Otherwise select $a' = argmax_a Q(s_t, a)$
            Update $Q(s_t, a_t) \mathrel{+}= \alpha [r_{t+1} + \gamma Q(s_{t+1}, a_{t+1}) - Q(s_t, a_t)]$
            Update $a_t = a_{t+1}, s_t = s_{t+1}$

---

End For
End For

## 5.3  Agents for Classical Control Problems

### 5.3.1  Nonlinear Q-function Approximators

As environments and tasks become more complex, tabular methods for $Q$-value iterations become impractical due to the increase in the number of possible state-action pairs. The solution to this problem is to use functions to estimate the value of $Q$. In deep reinforcement learning, a DNN acts as the function approximator for $Q(s, a)$, producing a separate output for each possible action and populating it with the value of that action for the input state. The architectures developed for the purpose of this investigation follow the guidelines given in the paper by Mnih (Mnih *et al.*, 2015).

### 5.3.2  Experience Replay

To improve network performance by getting a better representation of the true data distribution by breaking hidden patterns and correlations, Mnih proposes the concept of experience replay. This technique implies that the last $N$ experiences at each timestep are stored in a vector $e_t = (s_t, a_t, r_t, s_{t+1}, done_t)$, part of dataset $D_t$. *done* represents a Boolean which indicates if an episode is over. $D_t$ acts as a buffer, always containing the more recent $e_t$ values. During training, the agent samples a random minibatch of experiences from $D_t$. The size of the minibatch and $D_t$ represent hyperparameters that are selected by the programmer. If $D_t$ does not contain enough experiences, the agent performs random actions until the buffer is filled.

### 5.3.3  Semi-gradients and Dual Networks

The DNN with weights $W$ is trained by minimising the loss function (5.1) over every iteration, with $y_i$ depending on if the network uses the Q-learning (5.2) or SARSA algorithm (5.3).

$$L_i(W_i) = \mathbb{E}\{[y_i - Q(s, a; W_i)]^2 \mid s, a\} \qquad (5.1)$$

$$y_{i_{Q-learning}} = \mathbb{E}\left[r + \gamma \max_{a'} Q(s', a'; W_{i-1}) \mid s, a\right] \qquad (5.2)$$

$$y_{i_{SARSA}} = \mathbb{E}[r + \gamma Q(s', a'; W_{i-1}) \mid s, a] \qquad (5.3)$$

The gradients are calculated by differentiating the loss with respect to the weights (5.4)-(5.5) and then used to update the weights (5.6)-(5.7).

$$\frac{\partial L_i(W_i)}{\partial W_i}_{Q-learning} = \mathbb{E}\left[r + \gamma \max_{a'} Q(s', a'; W_{i-1}) - Q(s, a; W_i)\right]\frac{\partial Q(s, a; W_i)}{\partial W_i} \quad (5.4)$$

$$\frac{\partial L_i(W_i)}{\partial W_i}_{SARSA} = \mathbb{E}[r + \gamma Q(s', a'; W_{i-1}) - Q(s, a; W_i)]\frac{\partial Q(s, a; W_i)}{\partial W_i} \qquad (5.5)$$

$$W_{t+1_{Q-L.}} = W_t + \alpha \left[ r + \gamma \max_{a'} Q(s', a'; W_t) - Q(s, a; W_t) \right] \frac{\partial Q(s_t, a_t; W_t)}{\partial W_t} \quad (5.6)$$

$$W_{i+1_{SARSA}} = W_t + \alpha [r + \gamma Q(s', a'; W_i) - Q(s, a; W_t)] \frac{\partial Q(s_t, a_t; W_t)}{\partial W_t} \quad (5.7)$$

However, by using the same value function for generating both the targets and the errors during gradient descent, the network uses a semi-gradient rather than a gradient. This can lead to oscillations and divergence, as the same update increases both $Q(s', a')$ and $Q(s, a)$. In order to prevent this, every $C$ steps the network weights are saved into a separate cloned, or target, network $Q_{[c]}$ and used as updates for the Q-learning targets. Thus, (5.6)-(5.7) become (5.8)-(5.9). $C$ represents a hyperparameter.

$$W_{t+1_{Q-L.}} = W_t + \alpha \left[ r_{t+1} + \gamma \max_a Q_{[c]}(s', a'; W_t) - Q(s, a; W_t) \right] \frac{\partial Q(s_t, a_t; W_t)}{\partial W_t} \quad (5.8)$$

$$W_{t+1_{SARSA}} = W_t + \alpha [r_{t+1} + \gamma Q_{[c]}(s', a'; W_t) - Q(s, a; W_t)] \frac{\partial Q(s_t, a_t; W_t)}{\partial W_t} \quad (5.9)$$

### 5.3.4 Proposed Network Architecture

The proposed DNN for solving simple control problems is described in Table 5.1 and Figure 5.1. The paper by Mnih does not indicate how the weights are initialized. For the purpose of this investigation, the "Xavier" Initializer (Glorot and Bengio, 2010) from a truncated distribution was used. The network also uses the RMSprop optimisation algorithm with a 32 minibatch size. The inputs are the agent's observations of the environment $E$ generated with the OpenAI Gym toolkit. During training, actions are selected with either epsilon-greedy or softmax, depending on the test. $\varepsilon$ is set to decay following a power law function (5.10), as it was found that a linear decay, like the one in the original paper, does not allow the agents to learn effectively.

$$\varepsilon = \frac{1}{\sqrt{No. Episodes + 1}} \quad (5.10)$$

*Table 5.1: DNN Architecture*

| Layer | Type | Number of Neurons | Activation Function |
|-------|------|-------------------|---------------------|
| 1st | Input | Based on Observation Space | - |
| 2nd | Hidden Layer 1 | 200 | ReLU |
| 3rd | Hidden Layer 2 | 200 | ReLU |
| 4th | Output | Based on Action Space | Linear |

By using the above details, the proposed simple DQN (Mnih *et al.*, 2015) and simple DSN (Zhao *et al.*, 2017) algorithms are presented below as Algorithm 3 and Algorithm 4. These represent the conventional algorithms. The novel algorithms which employ DFA are largely identical, only differing in having the backward pass decoupled from the forward pass, allowing the DFA algorithm to be implemented.

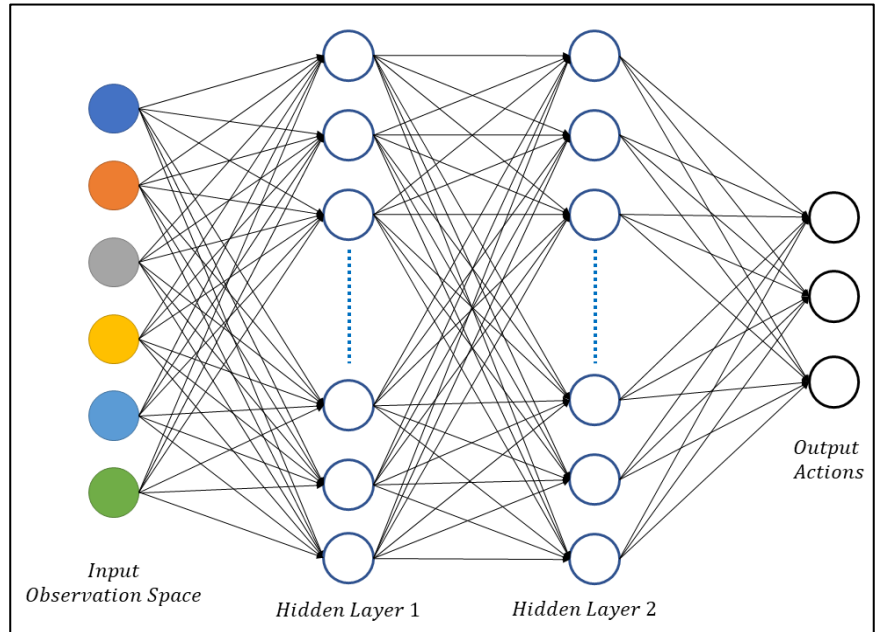

*Figure 5.2: Schematic of proposed DNN architecture*

---

**Algorithm 3: Simple Deep Q-Network (Adapted from: Mnih *et al.*, 2015)**

Initialize experience-replay memory $D$ to capacity $N$
Initialize action-value function $Q$ with "Xavier" randomised weights $W$
Initialize cloned action-value function $Q_{[c]}$ with weights $W_{[c]} = W$
**For** episode $= 1, M$ **do**
        Initialize sequence $s_1$
        **For** $t = 1, T$ **do**
                With probability $\varepsilon$ select a random action $a_t$
                Otherwise select $a_t = argmax_a Q(s_t, a; W)$
                Execute $a_t$ and observe the reward $r_t$ and new state $s_{t+1}$
                Store the experience $(s_t, a_t, r_t, s_{t+1}, done_t)$ in $D$
                Sample a random minibatch $(s_j, a_j, r_j, s_{j+1}, done_j)$ from $D$
                **If** episode terminates at step $j + 1$ **set**
                        $y_j = r_j$
                **Else set** $y_j = r_j + \gamma \max_{a'} Q_{[c]}(s_{j+1}, a'; W_{[c]})$
                Perform a gradient descent step on $\left[y_j - Q(s_j, a_j; W)\right]^2$
                Update network parameters $W$
                Every $C$ steps set $Q_{[c]} = Q$
        **End For**
**End For**

---

**Algorithm 4: Simple Deep SARSA Network (Adapted from: Zhao *et al.*, 2016)**

Initialize experience-replay memory $D$ to capacity $N$
Initialize action-value function $Q$ with "Xavier" randomised weights $W$
Initialize cloned action-value function $Q_{[c]}$ with weights $W_{[c]} = W$
**For** episode $= 1, M$ **do**
        Initialize sequence $s_1$
        With probability $\varepsilon$ select a random action $a_1$

---

Otherwise select $a_1 = argmax_a Q(s_1, a; W)$

**For** $t = 1, T$ **do**

    Execute $a_t$ and observe the reward $r_t$ and new image $I_{t+1}$

    Set $s_{t+1} = s_t$

    Store the experience $(s_t, a_t, r_t, s_{t+1}, done_t)$ in $D$

    Sample a random minibatch $(s_j, a_j, r_j, s_{j+1}, done_j)$ from $D$

    With probability $\varepsilon$ select a random action $a'$

    Otherwise select $a' = argmax_a Q(s_t, a; W)$

    **If** episode terminates at step $j + 1$ **set**

        $y_j = r_j$

    **Else set** $y_j = r_j + \gamma Q_{[c]}(s_{j+1}, a'; W_{[c]})$

    Perform a gradient descent step on $[y_j - Q(s_j, a_j; W)]^2$

    Update network parameters $W$

    Update $a_t = a_{t+1}$

    Every $C$ steps set $Q_{[c]} = Q$

**End For**

**End For**

## 5.4   Agents for Atari 2600

The agents created for playing more complex Atari 2600 are designed in a similar manner to those for classical control problems. Several additions are also made to allow the agents to use raw images as inputs.

### 5.4.1   Pre-Processing

The environment $E$, represented by the Arcade Learning Environment (Bellemare *et al.*, 2013) within the OpenAI Gym toolkit, passes to the agent $210x160x3$ RGB image inputs and a reward $r_t$ given by the score obtained by doing action $a_t$. CNNs can be used to process images, yet they cannot extract information about the system dynamics from static images. To overcome this issue, the Markov property is modified, by defining that state $s_t$ is composed out of a batch of stacked $m$ most recent frames (5.11).

$$s_t = [image_{t-3}, image_{t-2}, image_{t-1}, image_t] \qquad (5.11)$$

To reduce computational expenses, the images are converted to grayscale and cropped, eliminating any unnecessary information. Finally, as Atari games are usually simple, the images can be down sampled to an $84x84$ region. This is done using the nearest-neighbour interpolation method, as it preserves much of the original information (Figures 5.3, 5.4). All these techniques are gathered into the $\phi$ pre-processing function.

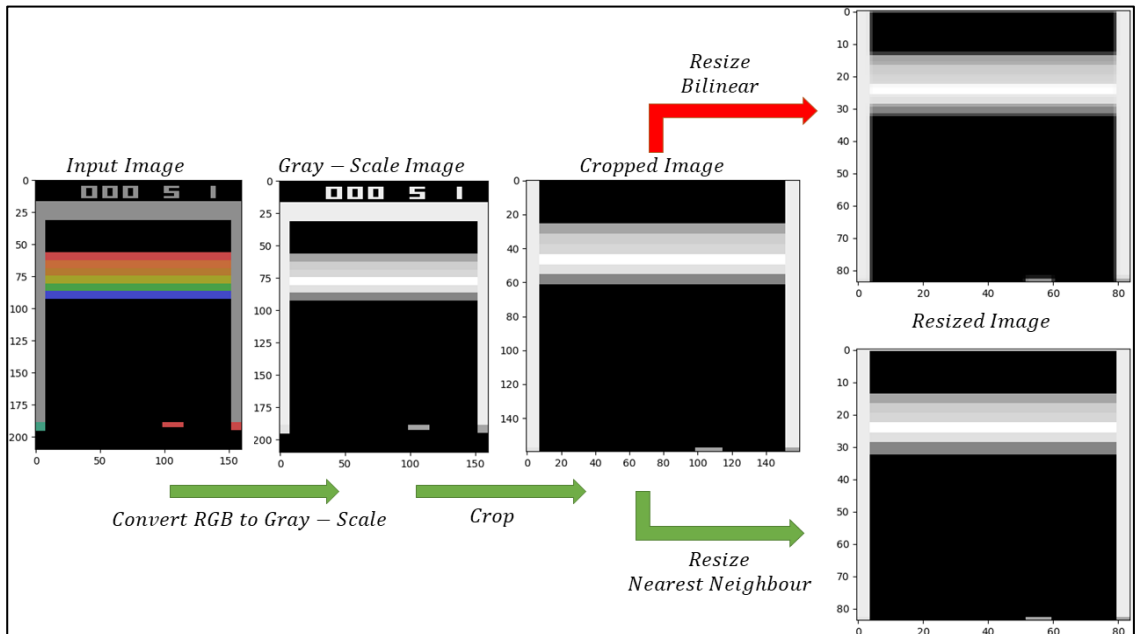

*Figure 5.3: Pre-processing for Breakout*

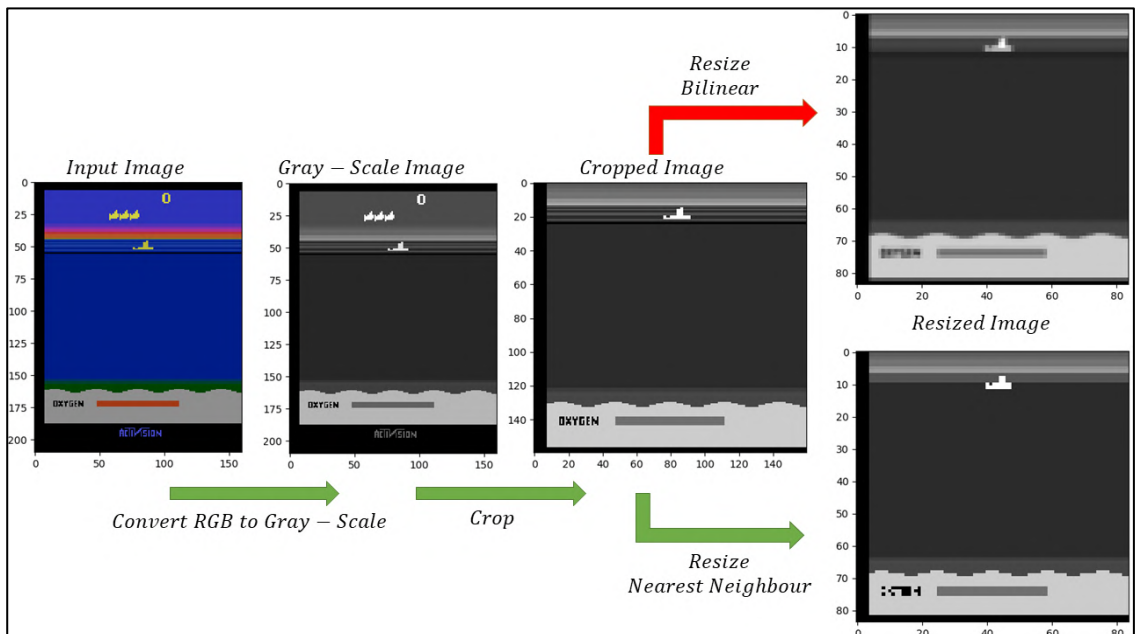

*Figure 5.4: Pre-processing for Seaquest*

## 5.4.2 Experience Replay for Stacked Frames

After processing, each state $s_t$ is represented by an array of size $4x84x84$. If a similar experience mechanism to that described in 5.3.2 was used, the network would suffer from increased memory usage, as at every time step, $s$ and $s'$ would store 8 frames, out of which only 5 are unique due to overlapping. Moreover, $s'_t$ is identical to $s_{t+1}$, which induces redundancy. To solve this issue, the experience replay buffer is designed to store consecutive frames, rather than states, with current states indexed between $t-4:t$ and future states between $t-3:t+1$, with the actions, rewards and done signal all stored at time index $t$ (Figure 5.5).

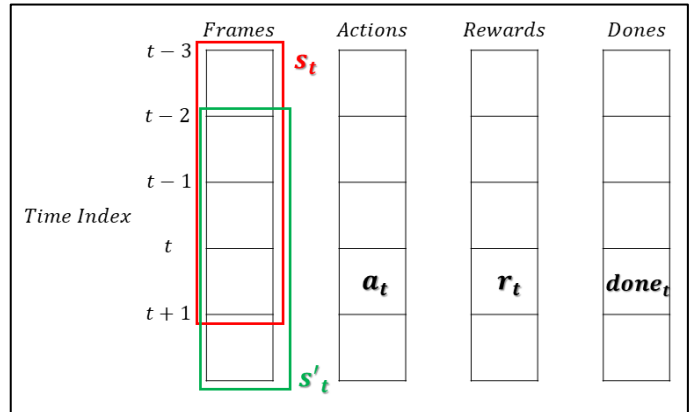

*Figure 5.5: Experience replay buffer*

This approach can cause issues when the boundary between different episodes are within the defined index range. To mediate these issues, at every timestep, a pointer is defined, marking the boundary between the oldest and newest frames. When randomly selecting states, if the index defining the state is ahead of the pointer, or four frames behind it, then it is considered invalid. Similarly, if the index is such that the state contains the Done signal, then the state is considered invalid. (Figure 5.6).

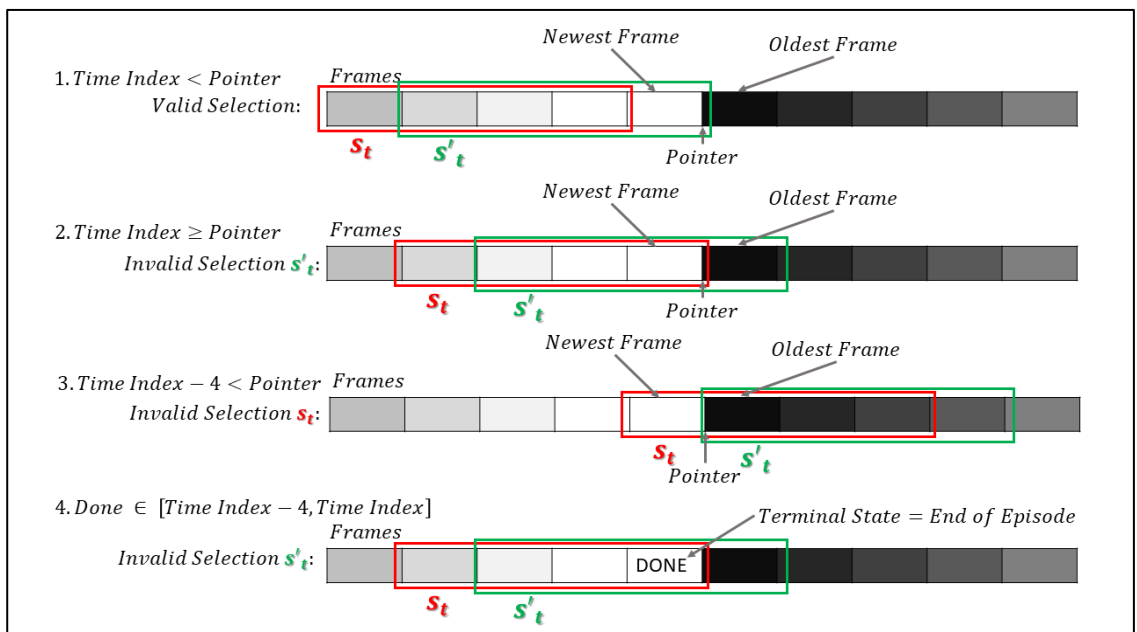

*Figure 5.6: Boundary conditions in state selection*

### 5.4.3  Deep Convolutional Neural Network

The $84x84x4$ map produced by $\phi$ is passed to a CNN (Table 5.2). The proposed CNN also uses the RMSprop optimisation algorithm with a 32 minibatch size. Like the DNN network, the "Xavier" Initializer is used for initialising the weights. For this study, the loss function was calculated using the mean squared error.

The CNN computes the $Q$ action-value function, producing a separate output for each possible action and populating it with the value of that action for the input state (Figure 5.7). The policy is selected either with epsilon-greedy or softmax, with epsilon decaying linearly from 1 to 0.1 over the first million state-action pairs, and after that being kept constant. The discount factor $\gamma$ is also set to 0.99.

*Table 5.2: CNN Architecture*

| Layer | Type | Number of Filters | Size | Stride | Activation Function |
|-------|------|-------------------|------|--------|---------------------|
| 1st | CONV | 32 | 8x8 | 4 | ReLU |
| 2nd | CONV | 64 | 4x4 | 2 | ReLU |
| 3rd | CONV | 64 | 3x3 | 1 | ReLU |
| 4th | Fully Connected | - | 512 | - | Linear |

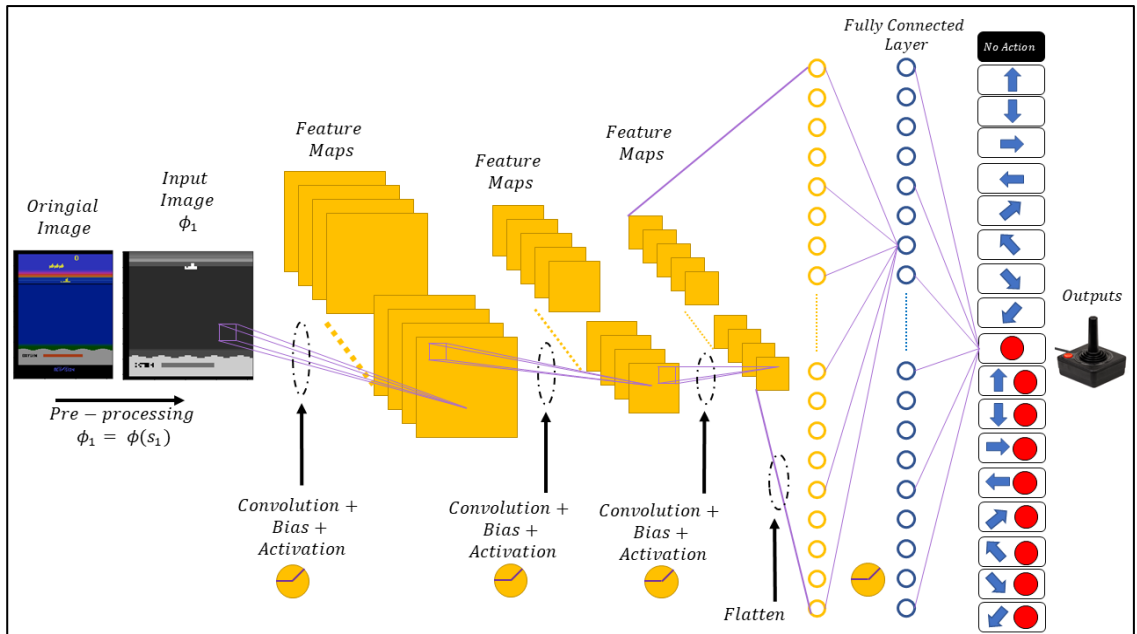

*Figure 5.7: Schematic of proposed CNN architecture*

By using the above methods, the proposed DQN (Mnih *et al.*, 2015) and DSN (Zhao *et al.*, 2017) algorithms are presented below as Algorithm 5 and Algorithm 6.

---

**Algorithm 5: Deep Q-Network (Source: Mnih *et al.*, 2015)**

---

Initialize experience-replay memory $D$ to capacity $N$

Initialize action-value function $Q$ with "Xavier" randomised weights $W$

Initialize cloned action-value function $Q_{[c]}$ with weights $W_{[c]} = W$

**For** episode = 1, $M$ **do**

    Initialize sequence $s_1 = \{I_1\}$

    Pre-process $\phi_1 = \phi(s_1)$

    **For** $t = 1, T$ **do**

        With probability $\varepsilon$ select a random action $a_t$

        Otherwise select $a_t = argmax_a Q(\phi(s_t), a; W)$

        Execute $a_t$ and observe the reward $r_t$ and new image $I_{t+1}$

        Set $s_{t+1} = \{s_t, a_t, I_{t+1}\}$

---

Pre-Process $\phi_{t+1} = \phi(s_{t+1})$
Store the experience $(\phi_t, a_t, r_t, \phi_{t+1})$ in $D$
Sample a random minibatch $(\phi_j, a_j, r_j, \phi_{j+1})$ from $D$
**If** episode terminates at step $j + 1$ **set**

$$y_j = r_j$$

**Else set** $y_j = r_j + \gamma \max_{a'} Q_{[c]}(\phi_{j+1}, a'; W_{[c]})$

Perform a gradient descent step on $\left[y_j - Q(\phi_j, a_j; W)\right]^2$
Update network parameters $W$
Every $C$ steps set $Q_{[c]} = Q$

        **End For**
**End For**

---

**Algorithm 6: Deep SARSA Network (Adapted from: Zhao *et al.*, 2016)**

Initialize experience-replay memory $D$ to capacity $N$
Initialize action-value function $Q$ with "Xavier" randomised weights $W$
Initialize cloned action-value function $Q_{[c]}$ with weights $W_{[c]} = W$
**For** episode $= 1, M$ **do**

    Initialize sequence $s_1 = \{I_1\}$
    Pre-process $\phi_1 = \phi(s_1)$
    With probability $\varepsilon$ select a random action $a_1$
    Otherwise select $a_1 = argmax_a Q(\phi(s_1), a; W)$
    **For** $t = 1, T$ **do**

        Execute $a_t$ and observe the reward $r_t$ and new image $I_{t+1}$
        Set $s_{t+1} = \{s_t, a_t, I_{t+1}\}$
        Pre-Process $\phi_{t+1} = \phi(s_{t+1})$
        Store the experience $(\phi_t, a_t, r_t, \phi_{t+1})$ in $D$
        Sample a random minibatch $(\phi_j, a_j, r_j, \phi_{j+1})$ from $D$
        With probability $\varepsilon$ select a random action $a'$
        Otherwise select $a' = argmax_a Q(\phi(s_t), a; W)$
        **If** episode terminates at step $j + 1$ **set**

$$y_j = r_j$$

        **Else set** $y_j = r_j + \gamma Q_{[c]}(\phi_{j+1}, a'; W_{[c]})$

        Perform a gradient descent step on $\left[y_j - Q(\phi_j, a_j; W)\right]^2$
        Update network parameters $W$
        Update $a_t = a_{t+1}$
        Every $C$ steps set $Q_{[c]} = Q$

    **End For**
**End For**

# 6 Implementation and Testing

Having developed the necessary algorithms, the next step is implementing them in code and testing them using different environments. This section presents the environments which were used in evaluating the various agents, as well as a series of details concerning the algorithms' implantation in code. Finally, this section describes the hyperparameter tuning and code evaluation procedures.

## 6.1 Testing Environments

For this study two sets of environments were used. Firstly, the simple reinforcement learning algorithms were evaluated using two simple games, consisting of standard, undiscounted, episodic tasks, with clearly defined states, actions and goals. Following these, the more complex agents are tested using different environments from the OpenAI Gym toolkit. This was chosen as it provides a large library of standardized environments operating from a shared interface which makes no assumptions about the structure of the utilized reinforcement learning code (OpenAI, 2019b). This allows results to be reproducible and easily compared with other results from literature.

### 6.1.1 Simple Games

As previously mentioned, the first step in this study was creating two simple RL agents, using SARSA and Q-learning, such that the behaviour and dynamics of the two algorithms can be captured. To evaluate these agents, two games were created: Gridworld Maze Runner and Gridworld Cliff Walker. These games have been used extensively throughout the RL literature to evaluate simple algorithms (Sutton and Barto, 2018), as they have a low number of state-action pairs, eliminating the need for a $Q$-function approximator.

In the case of both games, the environment consists of a grid representing all allowed states. Each state within the grid also contains a list with all the allowed actions. The agent can move in one or more of the four directions: up, down, left, right. Each environment contains a pre-defined start position and one or more terminal states, out of which one is the goal state. The agent receives a positive reward if it ends up in a goal state, or a negative reward if it ends up in any of the other terminal states. Arriving at a terminal state also finishes the current episode. A terminal state can be recognised as a state which is accessible but has no valid actions defined for it. A negative grid is also defined, which penalizes with a negative reward each move of the agent. This is done in order to incentivise the agent to solve the task as efficiently as possible, by minimizing the number of moves that it makes. The value of the penalty is a hyperparameter.

In the case of Gridworld Maze Runner (Figure 6.1), the start and end positions are defined together with a wall position, that the agent needs to surround. If the agent arrives in the goal state, it receives a $+1$ reward, while if it arrives at the "trap" terminal state, it is penalized by a $-1$ reward. Each step that the agent takes adds a negative $-0.1$ reward.

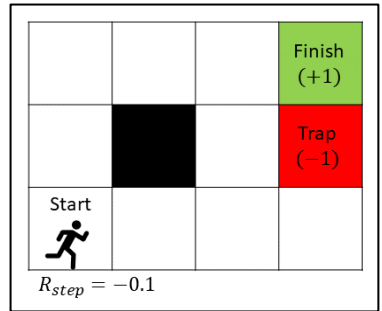

*Figure 6.1: Schematic representation of Gridworld Maze Runner*

Gridworld Cliff Walking follows a similar approach. The state-action space is greatly increased, allowing the agent more liberty in terms of the paths that it can take. This approach is designed to highlight the differences between SARSA and Q-learning. The reward scheme is also redefined, with the goal state generating a reward of $+10$, falling off the cliff incurring a $-100$ penalty and each step adding a $-1$ penalty.

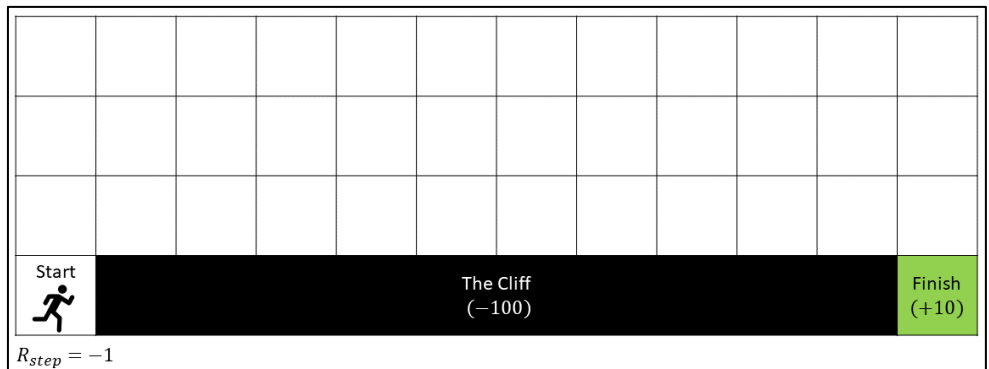

*Figure 6.2: Schematic representation of Gridworld Cliff Walker*

### 6.1.2 Classical Control Problems

To test the more complex algorithms which make use of a DNN for approximation the $Q$-values, a series of more complex environments were used. These are designed to reproduce classical reinforcement learning control problems, whose state-action space is already too big for allowing tabular methods. The information presented bellow is given on the OpenAI webpage (OpenAI, 2019a) and GitHub wiki (OpenAI, 2019c).

The first environment which was used for developing and testing the different agents is "CartPole-v0". In this environment, a pole is attached to an un-actuated joint placed on a cart. The agent controls the cart, which moves left and right across a frictionless surface, by applying a positive or negative unit force, so that the pole-pendulum is kept in a straight vertical position. The second utilised environment is "MountainCar-v0", where a car is placed on a 1D valley between two mountains. The goal of the agent is to build up enough momentum to climb the right-hand side mountain. The final tested environment is "Acrobot-v1", consisting of a robotic arm composed of two articulated pendulums with an actuated joint. The agent needs to swing the lower component above the base to a height equal to at least the length of one of the links (Subramoney, 2019). The links do not collide. Details of each environment is given in Table 6.1

*Table 6.1: Details of utilized classical control environments*

| Game Name | Observation Space | Action Space | Rewards | Termination Conditions |
|---|---|---|---|---|
| CartPole-v0 | 1. Cart Position<br>2. Cart Velocity<br>3. Pole Angle<br>4. Pole Velocity at Tip | 1. Push Left<br>2. Push Right | +1 for every step taken, including the terminal step | 1. Pole is more than $\pm12^o$ from vertical<br>2. Cart moves more than $\pm2.4$ units<br>3. Episode reaches length over 500 steps |
| MountainCar-v0 | 1. The car positions<br>2. The car velocity | 1. Push Left<br>2. Push Right<br>3. Do nothing (null) | -1 for each time step until the goal is reached | 1. Car reaches goal position, at the +0.5 position<br>2. Episode length reaches over 500 steps |
| Acrobot-v1 | 1. Sin and Cos of each of the two rotational joint angles (4 in total)<br>2. The angular velocities | 1. Apply positive torque<br>2. Apply negative torque<br>3. Do nothing (null) | -1 for each time step until the goal is reached | 1. Arm reaches goal position,<br>2. Episode length reaches over 500 steps |

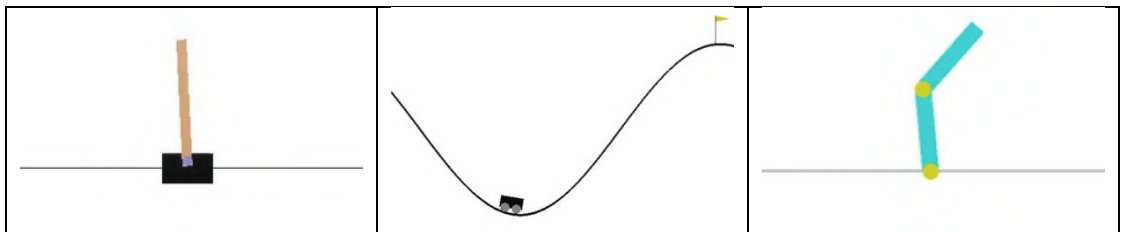

*Figure 6.3: Screenshots of classical control environments: CartPole, MountainCar and Acrobot (OpenAI, 2019a)*

### 6.1.3 Atari 2600 Games

The Atari 2600 games environments are based on the Arcade Learning Environment (Bellemare *et al.*, 2013) embedded in the OpenAI Gym toolkit. For these environments, each action is repeated uniformly for a duration of $k$ frames to account for frame stacking. Two Atari games were used during this investigation: Breakout and Seaquest.

In Breakout, the agent controls a palette which can reflect a ball. The goal of the agent is to gain points by breaking bricks while preventing the ball from falling outside the screen, which would cause the agent to lose a life. The game becomes progressively more complex, with the palette size decreasing and the ball speed increasing at different moments. The game finishes when all lives are depleted (Wikipedia.org, 2019a).

In Seaquest, the agent controls a submarine. The agent's goal is to gain points by saving divers, destroying sharks and enemy submarines. The submarine can hold only 6 divers, and each time it surfaces, one diver is removed. The submarine also has a limited supply of oxygen, which it replenishes when surfacing. Surfacing without any divers causes the agent to lose a life. Lives are also lost when the agent hits a shark, another submarine, a torpedo or runs out of oxygen, and are gained when increasing the score by 10,000. Every time the submarine resurfaces, the game speed and difficulty increase. (Wikipedia.org, 2019b). These particularities make this environment more challenging to solve.

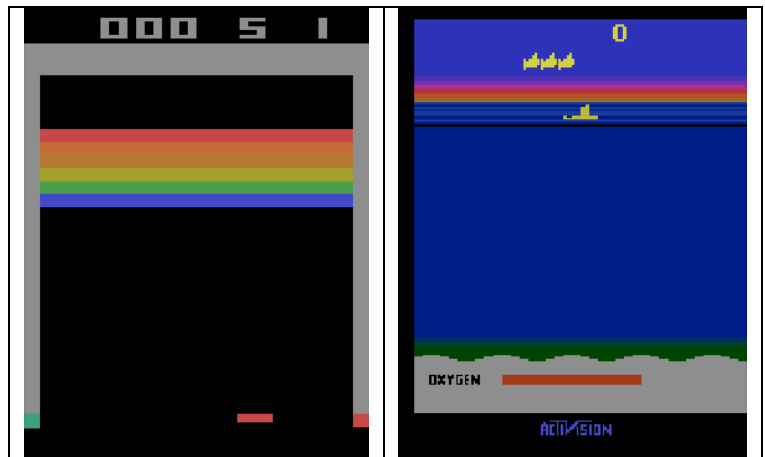

*Figure 6.4: Screenshots of Atari 2600 environments: Breakout and Seaquest*

## 6.2   Programming Language and Frameworks

All the agents utilised in this study were coded using the Python 3.6 programming language, as it is open source and widely-used in industry and scientific research. Several libraries were also used, including NumPy, for multi-dimensional matrix manipulations, SciPy, for image augmentation and manipulation, Matplotlib, for plotting, as well as other, smaller packages, such as csv, copy, os, sys, random and datetime. A dedicated ML library named Tensorflow has also been used, in collaboration with the Nvidia® CUDA GPU interfacing software, made available for research by the University of Sheffield.

## 6.3   Code Implementation

Coding was done in Visual Studio Code, a source-code editor developed by Microsoft. This, together with all relevant libraries, was installed on a virtual machine, created using Oracle VM Virtual Box, which was running Ubuntu 16.04 LTS, an open-source, versatile, Linux-based operating system.

The code was implemented on the ShARC HPC. This was accessed through an SSH. Conda, a Python package manager, was then used to create a virtual environment. Simulation jobs were submitted as batch files, using a bash file, which contained commands for the schedule options, loaded the required modules and activated the relevant virtual environment. This approach allowed multiple batch jobs containing series of scripts to be run in parallel, increasing the efficiency of results generation.

## 6.4   Agent Tuning and Evaluation

The simple RL agents were trained for 10,000 episodes by using the same hyper-parameters (Table 6.2), identified using literature (Sutton and Barto, 2018) and through experimentation. The training performance was assessed by observing the running averages of the obtained rewards and the deltas between the values of consecutive Q-functions. No hyperparameter tuning was performed for these agents, as their purpose was to highlight the difference between the different RL algorithms.

*Table 6.2: Simple RL agents hyperparameters*

| Hyperparameter | Value |
|---|---|
| Discount Factor $\gamma$ | 0.9 |
| Original Learning Rate $\alpha_0$ | 1.0 |
| Time Parameter $t_\varepsilon$ | 1.0 |
| Episodic Increment for $count(s_t, a_t)$ | 0.005 |
| Number of Training Episodes | 10,000 |

Before evaluating the agents for classical control, an optimisation operation was performed to determine a set of optimal hyperparameters. This was performed by first experimenting with different parameter values, based on experience and literature recommendations, followed by a grid search accompanied by a sensitivity study, where a single parameter was varied at a time. The original parameters are described in Table 6.3.

*Table 6.3: List of original hyperparameters of classical control agents and their values*

| | Hyperparameter | Original Value | Description |
|---|---|---|---|
| Tuned Parameters | Learning Rate $\alpha$ | $1e-2$ | A parameter controlling the agent's learning speed. Large values can prevent the agent from converging and lead to oscillations, while small values could increase learning times. |
| | Decay Rate $\beta$ | 0.99 | A factor used for controlling the degree of network overfitting by discounting the history gradient. |
| | Discount Factor $\gamma$ | 0.99 | The discount factor in RL, defining how significant future rewards are when in a given state. Should be close to 1 to help the agent determine the best policy by making good long-term decisions. |
| | Small Scalar $\epsilon$ | $1e-8$ | Small value used to avoid a zero denominator in the optimisation algorithm. |
| | Copy Period | 50 | The number of training steps after which the network parameters are copied to the target network. |
| | Training Penalty | $-200$ | Negative reward applied to the agent if it finished the episode in several steps lower than the maximum defined value. This incentivises the agent to find the best policy rather than randomly explore the environment. |
| | Training Episodes | 500 | Number of training episodes, which should balance training performance and potential data overfitting. |
| Fixed Parameters | Minibatch Size | 32 | Size of minibatch used in training. |
| | Replay Memory | 10,000 | The size of the experience replay buffer. |
| | Replay Start Size | 100 | The minimum number of experiences to be collected before training starts. |
| | Initial $\varepsilon$ | 1.0 | The initial value of the $\varepsilon$ parameter in epsilon-greedy or softmax. |

The values obtained through the grid search were compared across the different agents and games and a global set was selected. Using these parameters, the new agents were trained and evaluated in accordance with Table 3.2 and Appendix B, with their performance being compared.

Due to the limited time availability and the long training periods for the complex Atari agents, no hyperparameter tuning study was performed for them. The parameters for a first set of simulations were set in line with Mnih's findings (Mnih *et al.*, 2015). Then, several parameters were slightly modified in order to account for the previous classical control findings. (Table 6.4). For the original hyperparameters, the rewards are also capped to $\pm 1$, in line with the original paper.

*Table 6.4: List of hyperparameters and their values for Atari agents*

| Hyperparameter | Original Value (Mnih *et al.*, 2015) | Modified Value |
|---|---|---|
| Learning Rate $\alpha$ | $2.5e-4$ | $2.5e-4$ |
| Decay Rate $\beta$ | 0.95 | 0.99 |
| Momentum Gradient | 0.95 | 0 |
| Discount Factor $\gamma$ | 0.99 | 0.99 |
| Small Scalar $\epsilon$ | $1e-2$ | $1e-3$ |
| Copy Period | 10,000 | 10,000 |
| Training Frames | ~50,000,000 | ~50,000,000 |
| Minibatch Size | 32 | 32 |
| Replay Memory | 1,000,000 | 500,000 |
| Replay Start Size | 50,000 | 50,000 |
| Initial $\varepsilon$ | 1.0 | 1.0 |
| Final $\varepsilon$ | 0.1 | 0.1 |
| Final Exploration Frame | 1,000,000 | 500,000 |
| Agent History Length | 4.0 | 4.0 |
| Action Repeat | 4.0 | 4.0 |

# 7 Results and Discussion

Using the previously described algorithms and methods, the various agents created for the purpose of this study were trained and evaluated. This section presents the obtained results for the three sets of investigated agents. The focus of the analysis is on how the different agent architectures impact the performance of the various agents both during and after training.

## 7.1 Simple Reinforcement Learning Codes

The first step in this study was to train the simple RL agents, with the purpose of observing the differences between on-policy and off-policy learning methods. This was done by having the agents solve the Gridworld Maze Runner and Gridworld Cliff Walking tasks.

In the case of the Gridworld Maze Runner game (Figure 7.1), both agents asymptotically converge to the optimal policy (Figure 7.2), choosing the shortest and safest path, preventing them from reaching the trap state.

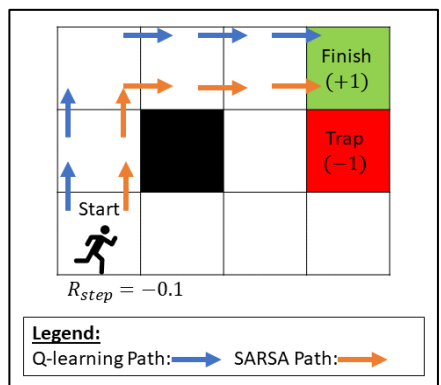

*Figure 7.1: Paths selected by the Q-learning and SARSA agents in Maze Runner*

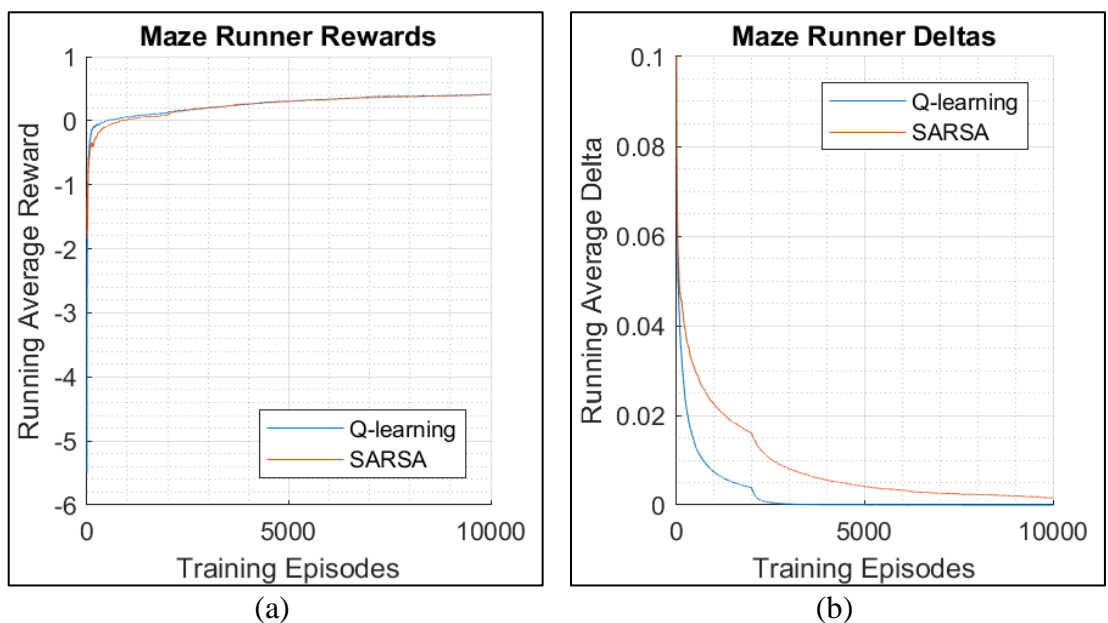

(a)                                                (b)

*Figure 7.2: Running average of the rewards and deltas for agents in Maze Runner*

When testing the agents in a more complex environment, which gives them more freedom in choosing a route, the differences between the on and off-policy learning methods can be better observed. When tasked with solving the Cliff Walker environment (Figure 7.3), for the same hyper parameters, it can be observed that the Q-learning agent learns the optimal, shortest path, while the SARSA agent learns a longer, but safer path. When analysing the rewards (Figure 7.4), it can be observed that although the Q-learning argent determines the optimal path, due to the stochastic nature of epsilon-greedy, it travels on the edge of the cliff, which leads to it sometimes falling off the edge, incurring a penalty. The SARSA agent, takes the longer yet safer path, as it takes the action selection into account (Sutton and Barto, 2018). This costs the SARSA agent more in terms of the negative rewards incurred for the number of taken steps, but allows it to perform better online, as it prevents it from incurring the larger penalties associated with falling off the cliff, as with the Q-learning agent.

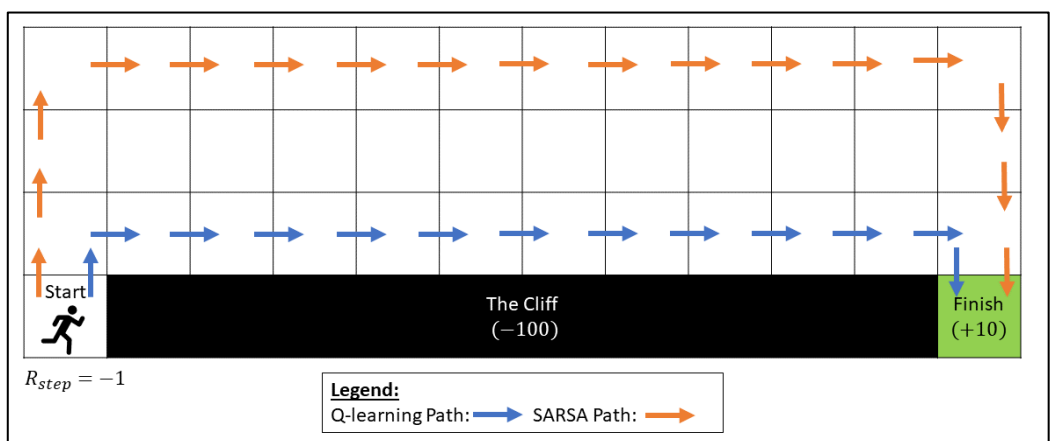

*Figure 7.3: Paths selected by the Q-learning and SARSA agents in Cliff Walker*

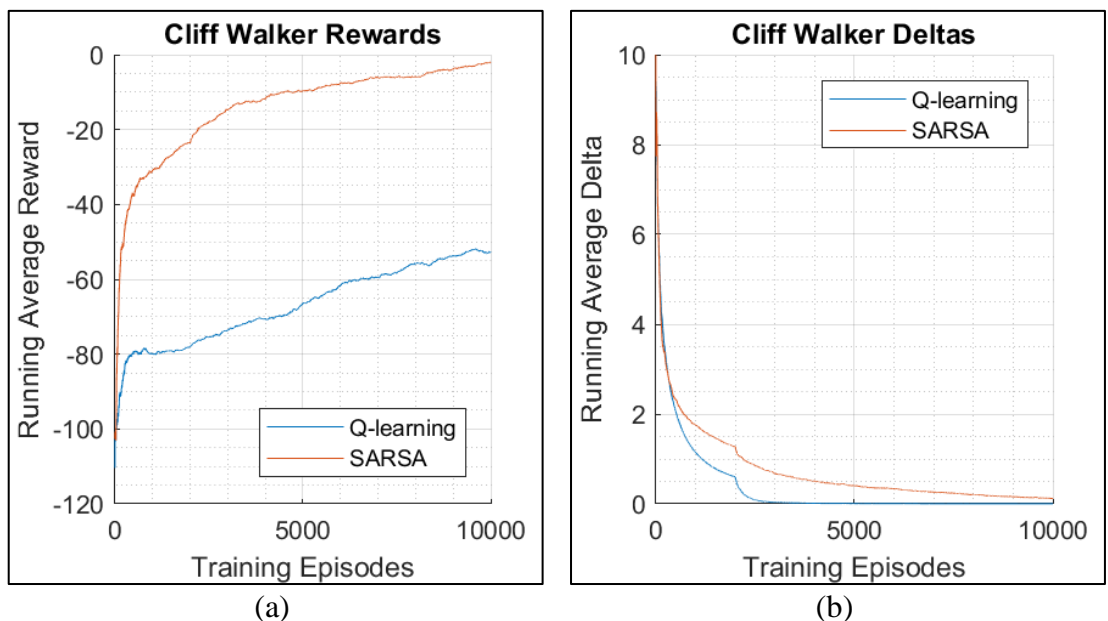

*Figure 7.4: Running average of the rewards and deltas for agents in Cliff Walker*

This first finding suggests that, in certain complex environments, the on-policy SARSA learning method might perform better than the off-policy Q-learning method. This finding prompted the development of deep SARSA agents for the subsequent, more complex tasks.

## 7.2 Agents for Classical Control Problems

### 7.2.1 Hyperparameter Tuning

Prior to evaluating the classical control agents, a set of hyperparameters which maximise the agents' performance was determined using a grid search, in line with the methodology described in Section 6.4. The tuning operation was performed on all the four investigated agents (DQN, DSN, DQN-DFA, DSN-DFA) in all the considered environments. However, due to time constraints, this was only performed on agents trained using epsilon-greedy, thus excluding softmax based agents. Subsequent results show that the parameters identified are compatible with the softmax-based agents, yet, for completion, any further work should include a grid search for these agents. The best hyperparameter values were selected by employing an alternative ranking approach. If a given value was clearly better than the others, obtaining the best performance in over 50% of the tested cases, than that value was selected. Alternatively, the two top performing values were selected, and their performance ranked across all the test cases, with the one achieving better overall performance being selected. This approach allows for a compromise to be established when no clear value achieves the best performance.

Firstly, the learning rate was tuned. The first observation which is made is that for CartPole (Figure 7.5), all the four agents are learning, including the agents using the novel DFA architectures. The learning performance varies widely across the three games (Figures 7.5-7.7), though this is probably due to some of the other fixed parameters, as all the agents seem to perform equally poor in the MountainCar and Acrobot games. Despite this, the top learning rate values appear to be $1e-4$ and $1e-1$, with $1e-4$ performing better throughout.

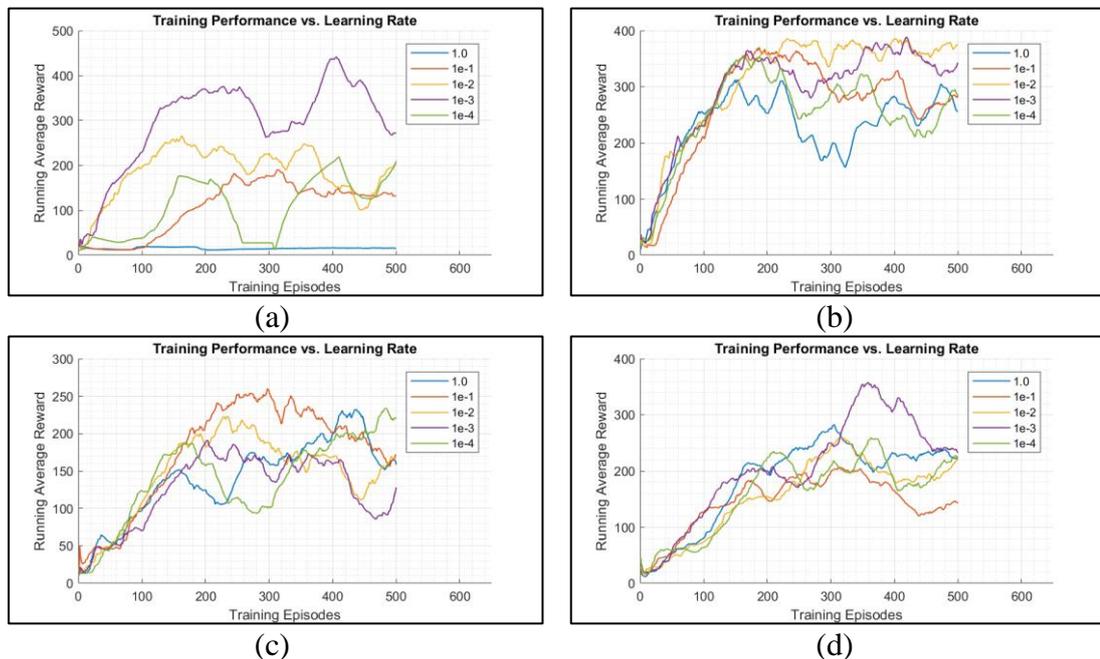

*Figure 7.5: Learning rate tuning for DQN(a), DSN(b), DQN-DFA(c) and DSN-DFA(d) in CartPole*

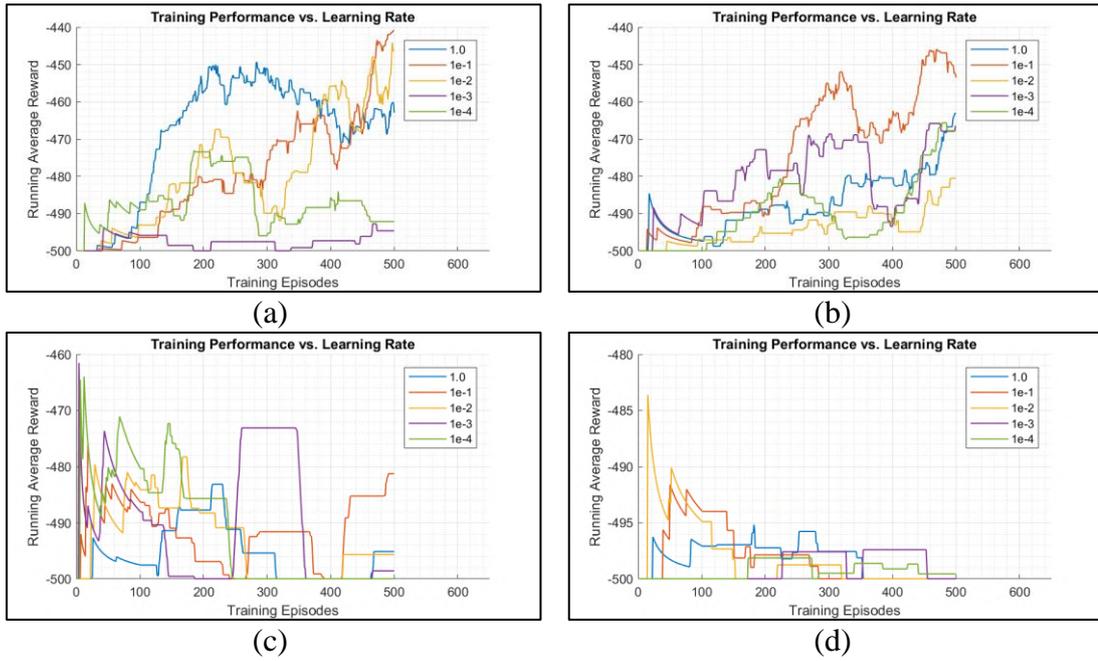

*Figure 7.6: Learning rate tuning for DQN(a), DSN(b), DQN-DFA(c) and DSN-DFA(d) in MountainCar*

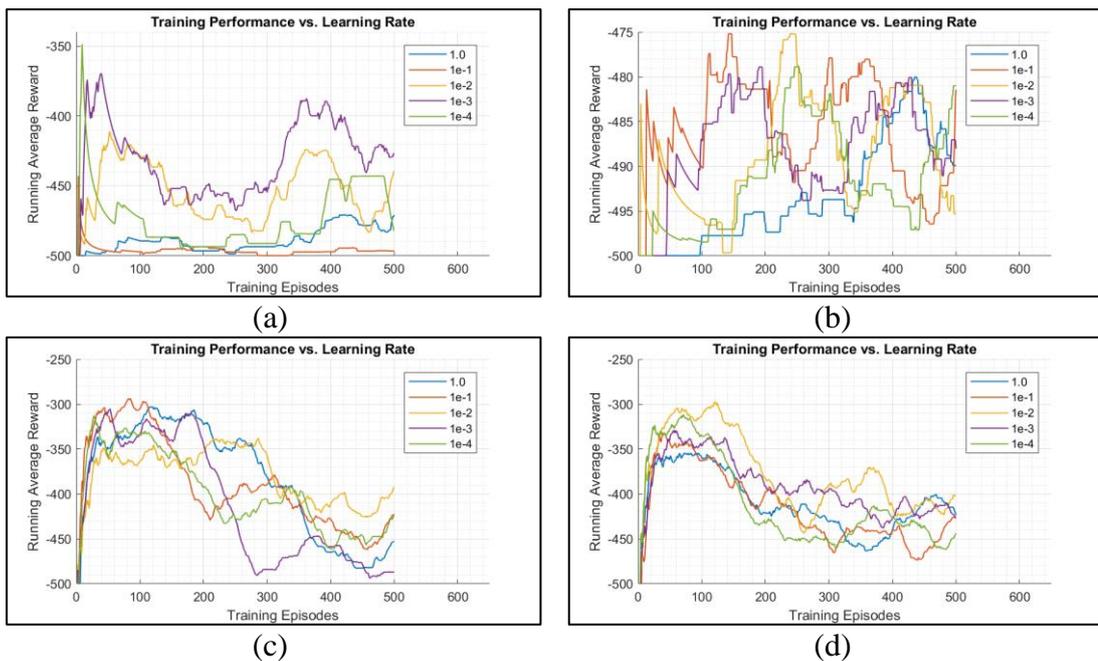

*Figure 7.7: Learning rate tuning for DQN(a), DSN(b), DQN-DFA(c) and DSN-DFA(d) in Acrobot*

The second tuned parameter was the decay rate. With a few exceptions, the 0.99 value performed best in all training cases, being the only one which appeared to train in certain instances (Figure 7.9 (a) and (c)).

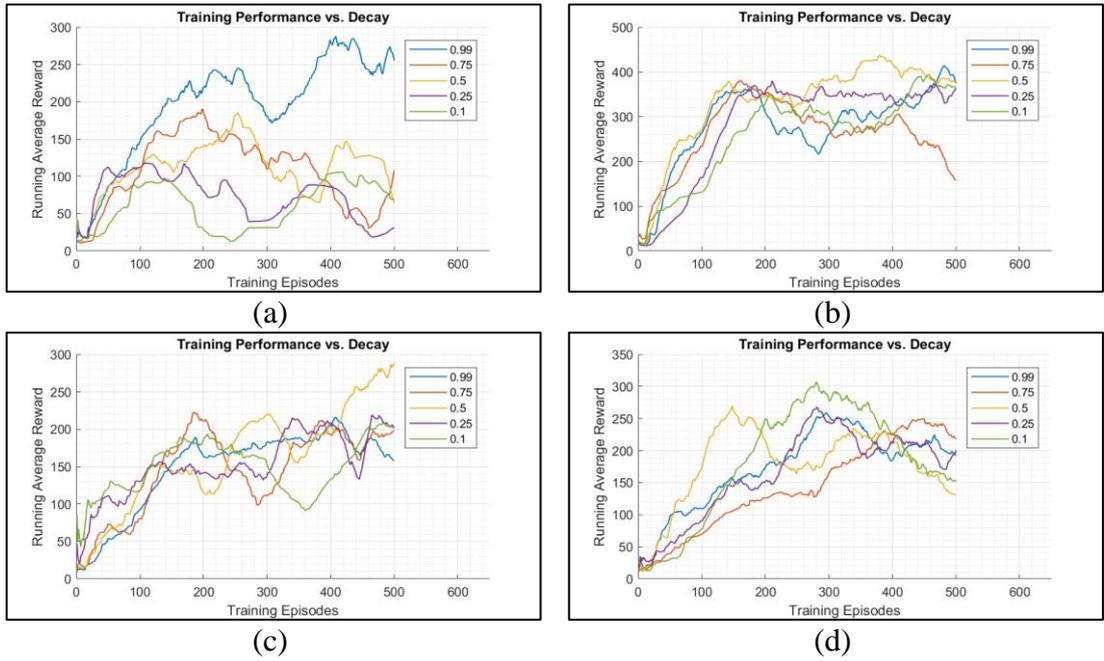

*Figure 7.8: Decay rate tuning for DQN(a), DSN(b), DQN-DFA(c) and DSN-DFA(d) in CartPole*

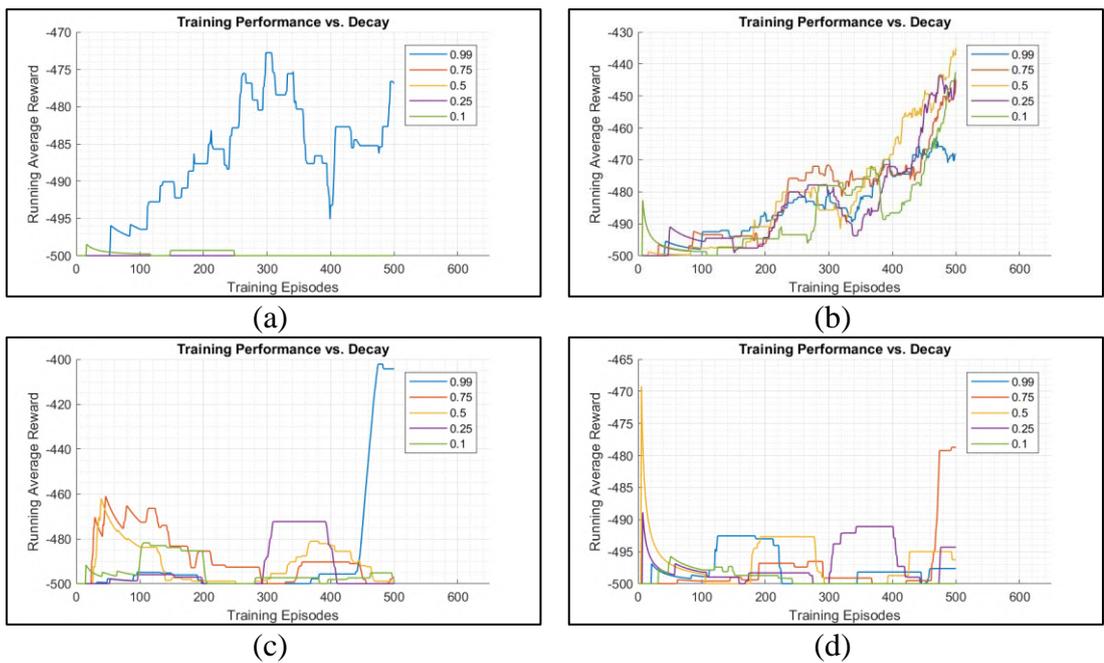

*Figure 7.9: Decay rate tuning for DQN(a), DSN(b), DQN-DFA(c) and DSN-DFA(d) in MountainCar*

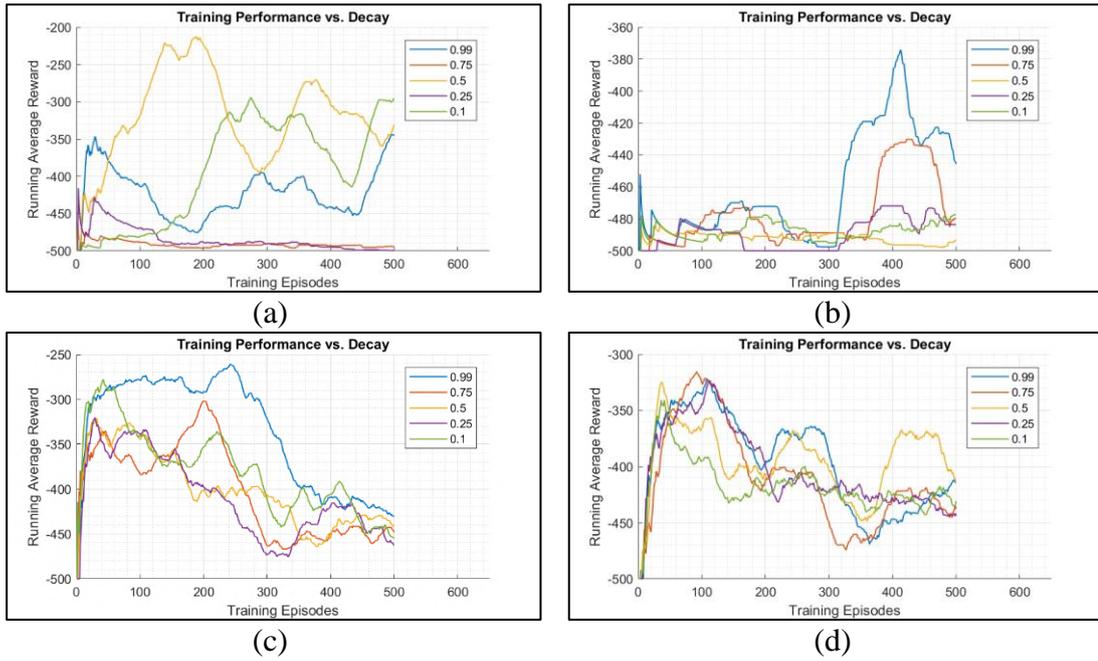

*Figure 7.10: Decay rate tuning for DQN(a), DSN(b), DQN-DFA(c) and DSN-DFA(d) in Acrobot*

Akin to the decay rate, the discount factor tuning showed that the 0.99 value appears to offer the best training performance throughout all the agents and games.

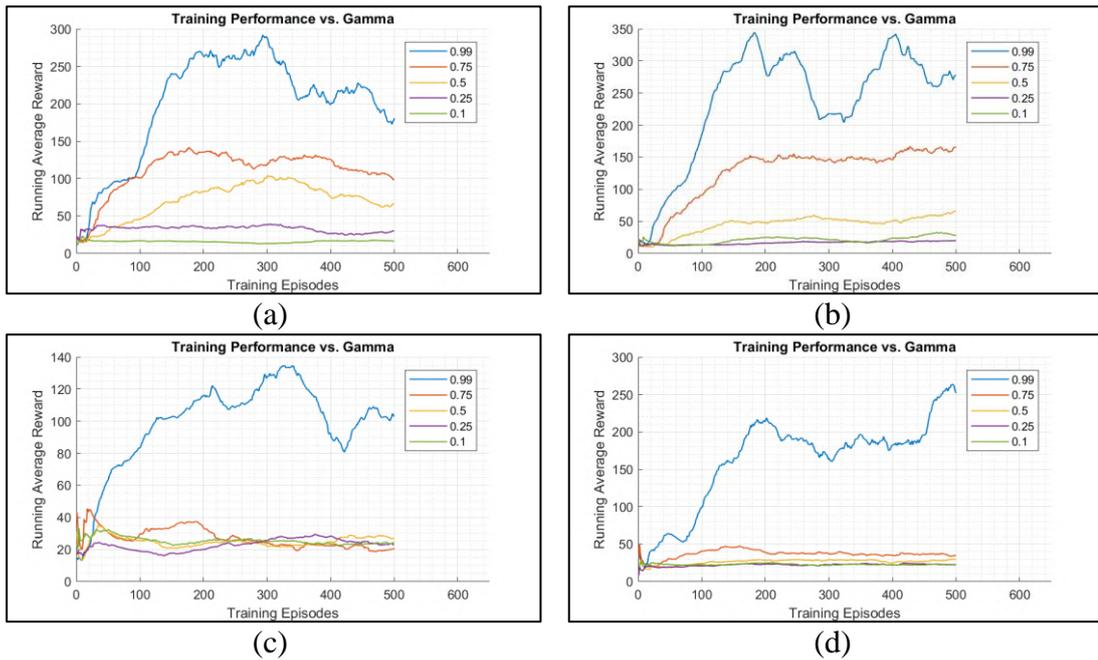

*Figure 7.11: Discount factor tuning for DQN(a), DSN(b), DQN-DFA(c) and DSN-DFA(d) in CartPole*

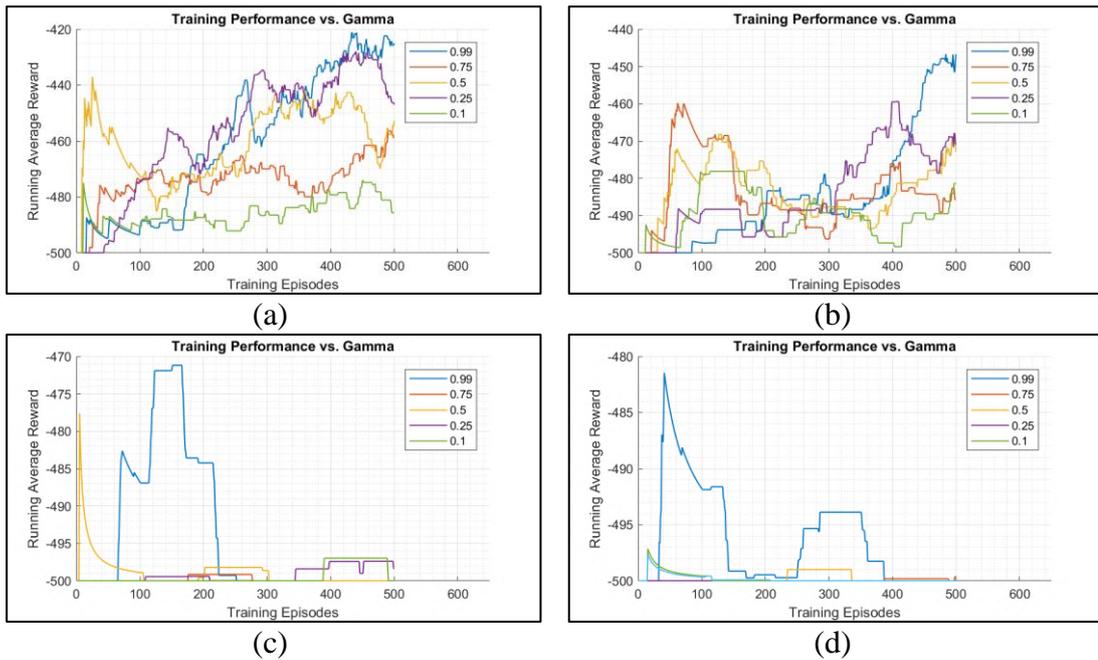

*Figure 7.12: Discount factor tuning for DQN(a), DSN(b), DQN-DFA(c) and DSN-DFA(d) in MountainCar*

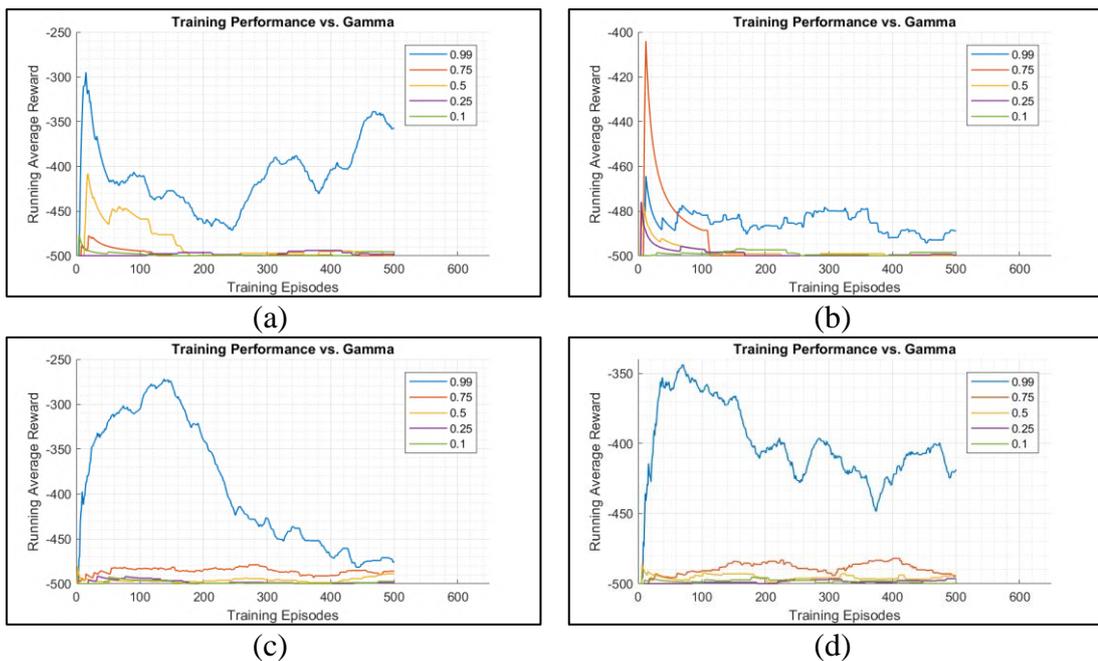

*Figure 7.13: Discount factor tuning for DQN(a), DSN(b), DQN-DFA(c) and DSN-DFA(d) in Acrobot*

The training penalty was the next parameter which was tuned. It can be observed that, for the CartPole game (Figure 7.14), penalizing the agents if they finish the training episodes faster than the maximum number of steps can accelerate their training. Despite this, using this approach has an unintended consequence of inducing an oscillatory behaviour in the agents' training, as it appears to facilitate having the agents getting stuck in local minimums. For MountainCar and Acrobot (Figures 7.15, 7.16), the classical agents trained with no penalty train significantly better than the agents trained with a penalty. For these games, the training penalty value does not seem to impact the training performance of the novel agents, which remains poor. Taking all into consideration, a 0-training penalty was selected.

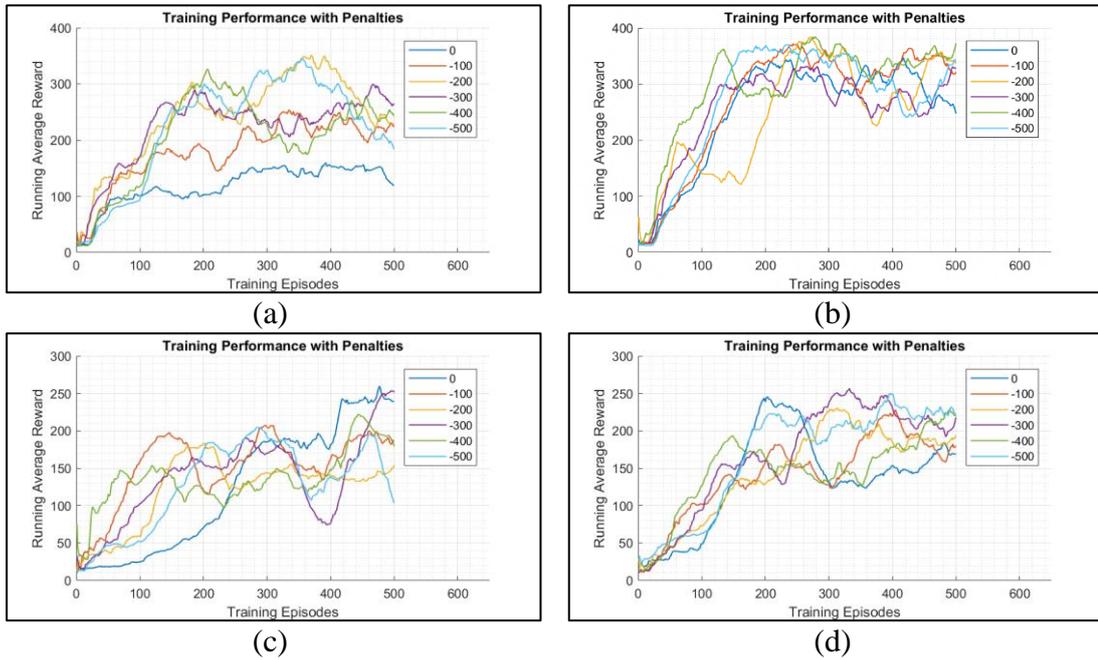

*Figure 7.14: Penalty rate tuning for DQN(a), DSN(b), DQN-DFA(c) and DSN-DFA(d) in CartPole*

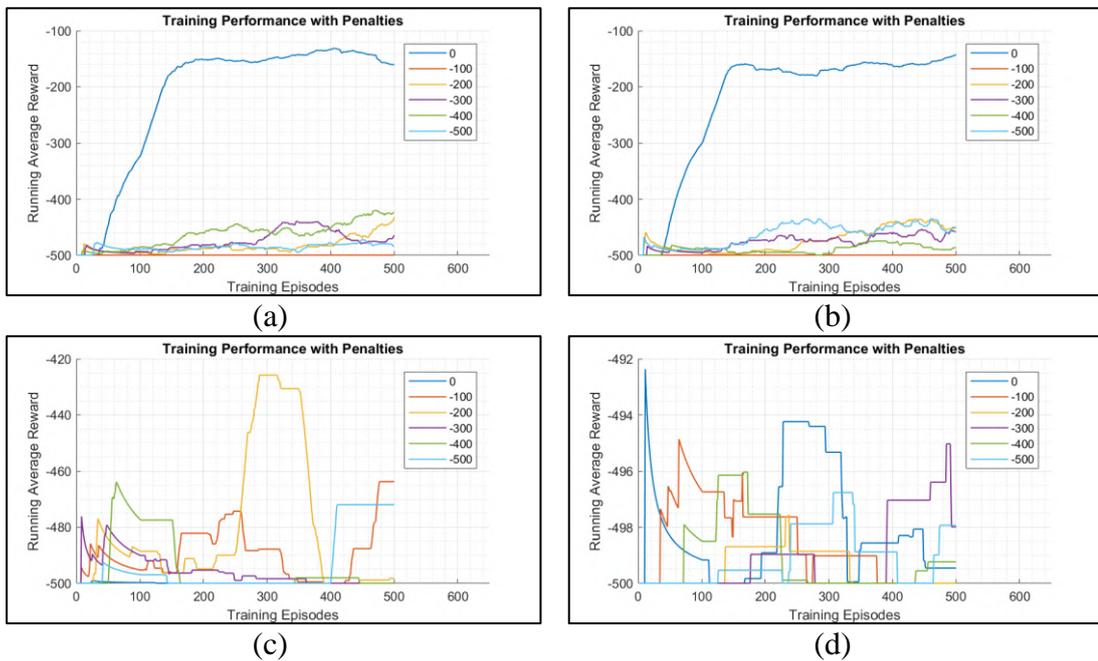

*Figure 7.15: Penalty rate tuning for DQN(a), DSN(b), DQN-DFA(c) and DSN-DFA(d) in MountainCar*

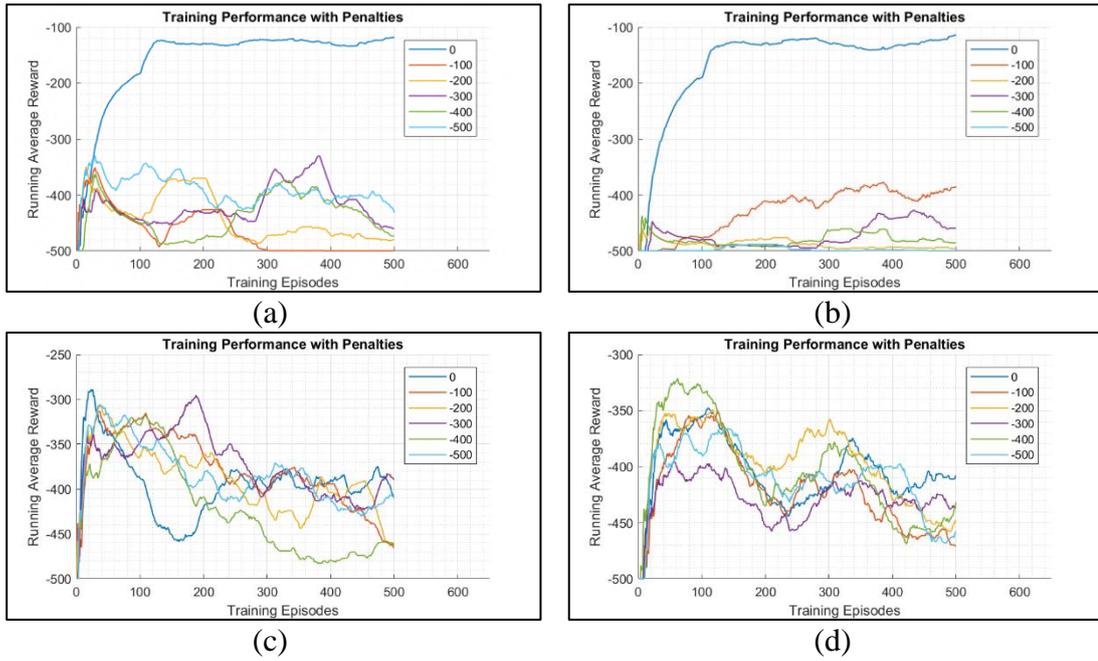

*Figure 7.16: Penalty rate tuning for DQN(a), DSN(b), DQN-DFA(c) and DSN-DFA(d) in Acrobot*

The tuning results were not as obvious for the small scalar epsilon parameter as they were in the previous case. Overall, it appears that two values perform better ($1e-1$ and $1e-3$), out of which the $1e-3$ offered the best compromise across the different agents and environments.

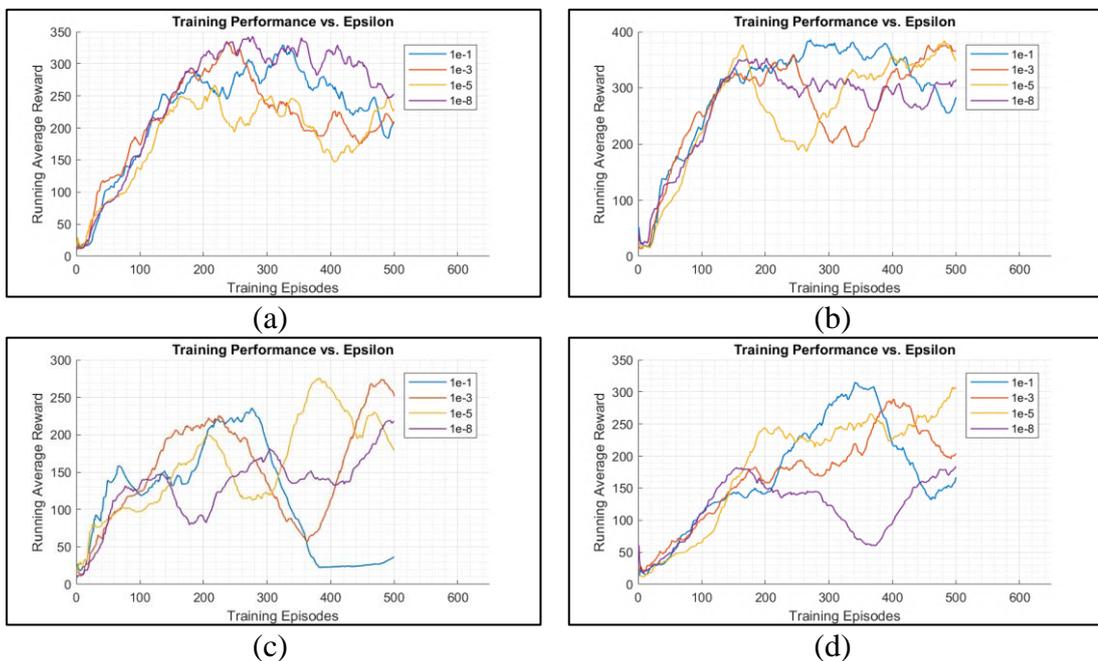

*Figure 7.17: Scalar epsilon tuning for DQN(a), DSN(b), DQN-DFA(c) and DSN-DFA(d) in CartPole*

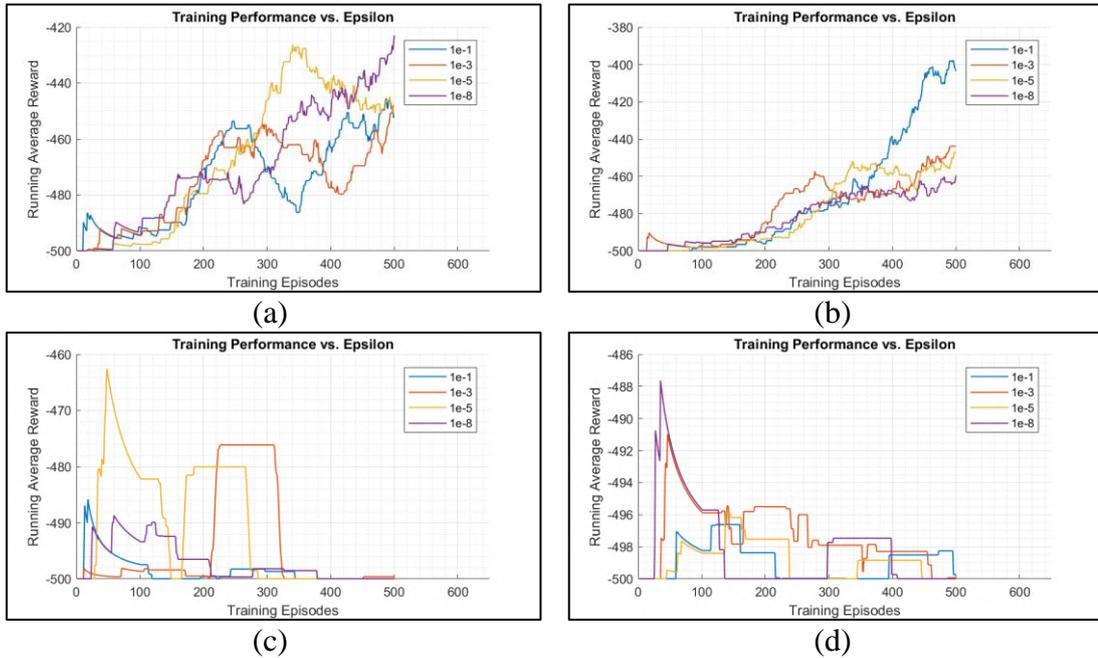

*Figure 7.18: Scalar epsilon tuning for DQN(a), DSN(b), DQN-DFA(c) and DSN-DFA(d) in MountainCar*

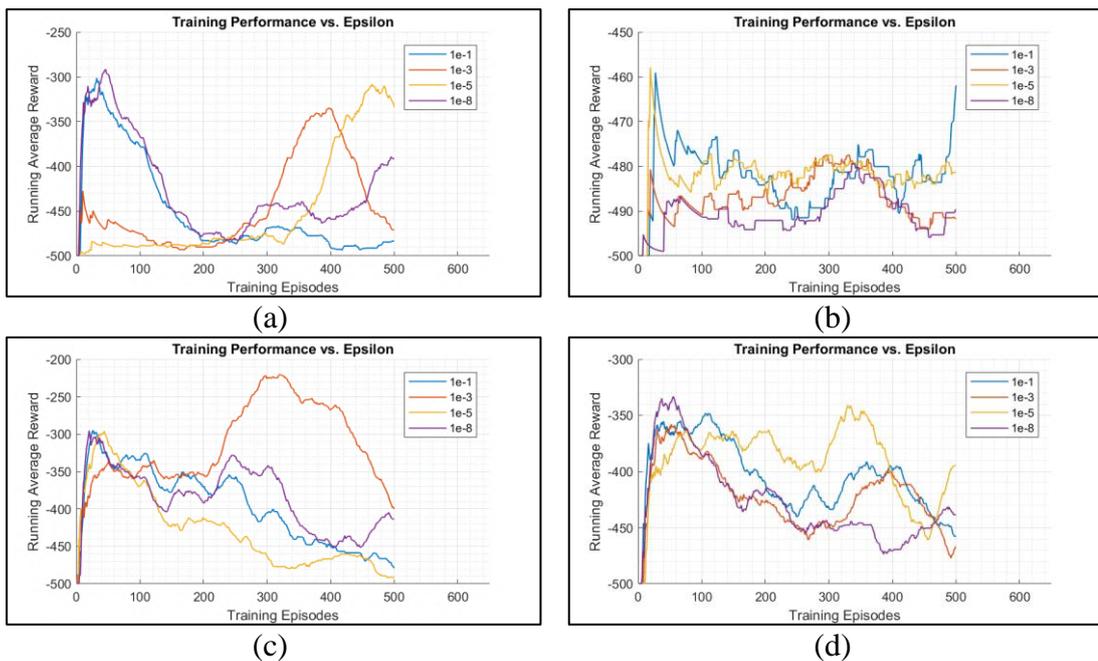

*Figure 7.19: Scalar epsilon tuning for DQN(a), DSN(b), DQN-DFA(c) and DSN-DFA(d) in Acrobot*

Out of all the tuned hyperparameters, the copy period, or number of training steps after which the network is copied, yielded the most diverse set of results, with some agents and environments appearing to respond better to short copy periods, while others appearing to prefer the longer copy periods, up to 1000 steps. By progressively ranking the options, the 200 steps value was found to be the best compromise.

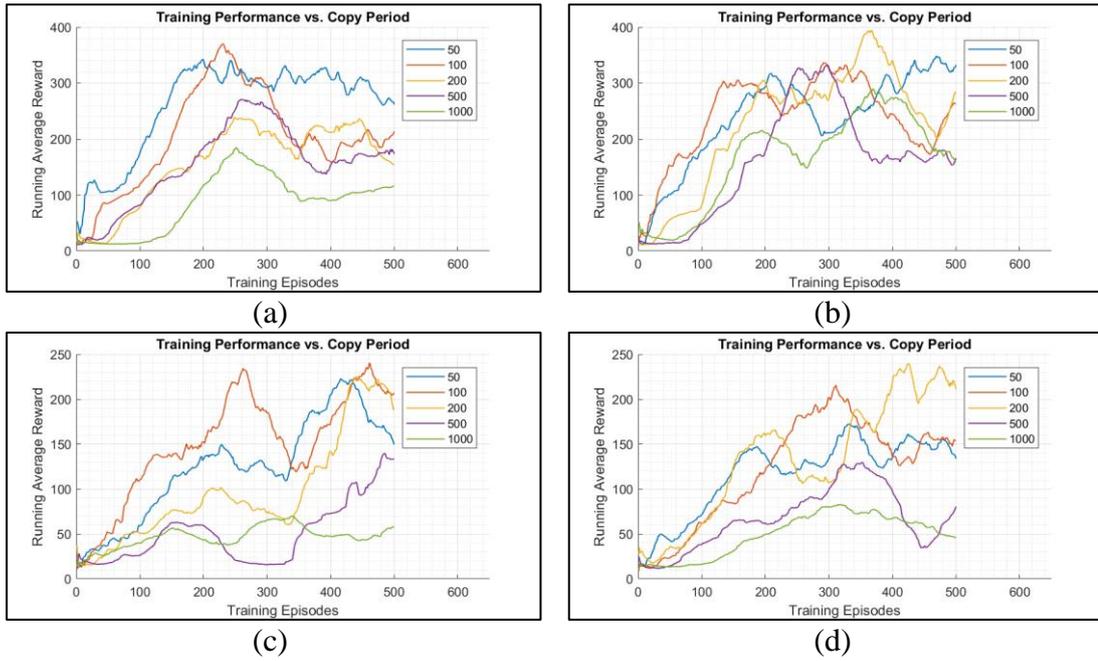

*Figure 7.20: Copy period tuning for DQN(a), DSN(b), DQN-DFA(c) and DSN-DFA(d) in CartPole*

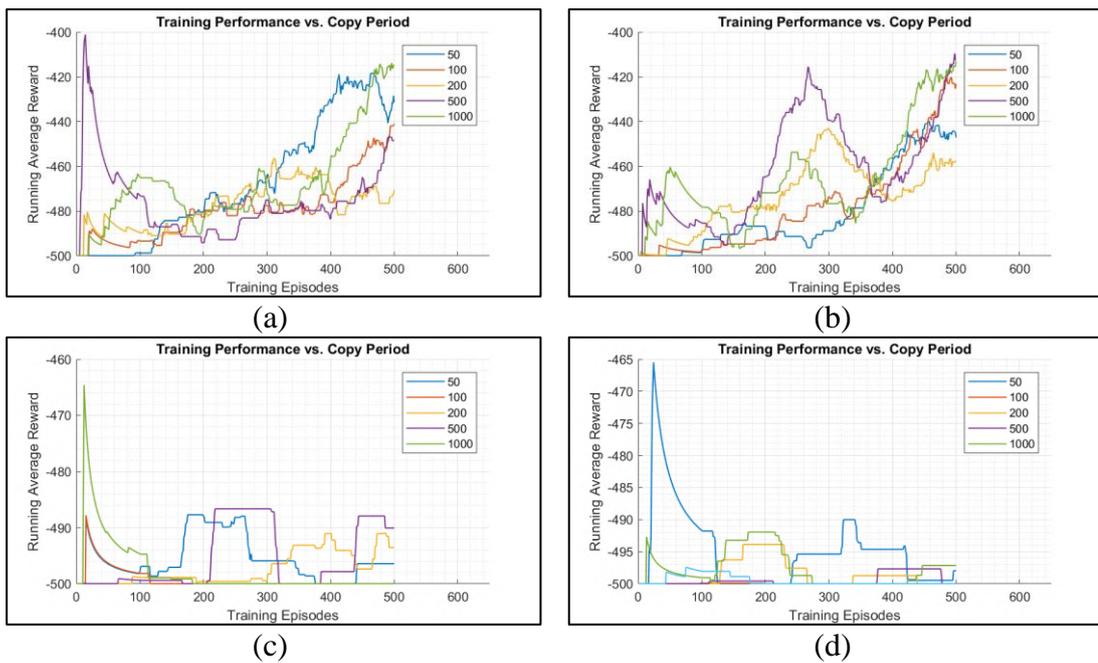

*Figure 7.21: Copy period tuning for DQN(a), DSN(b), DQN-DFA(c) and DSN-DFA(d) in MountainCar*

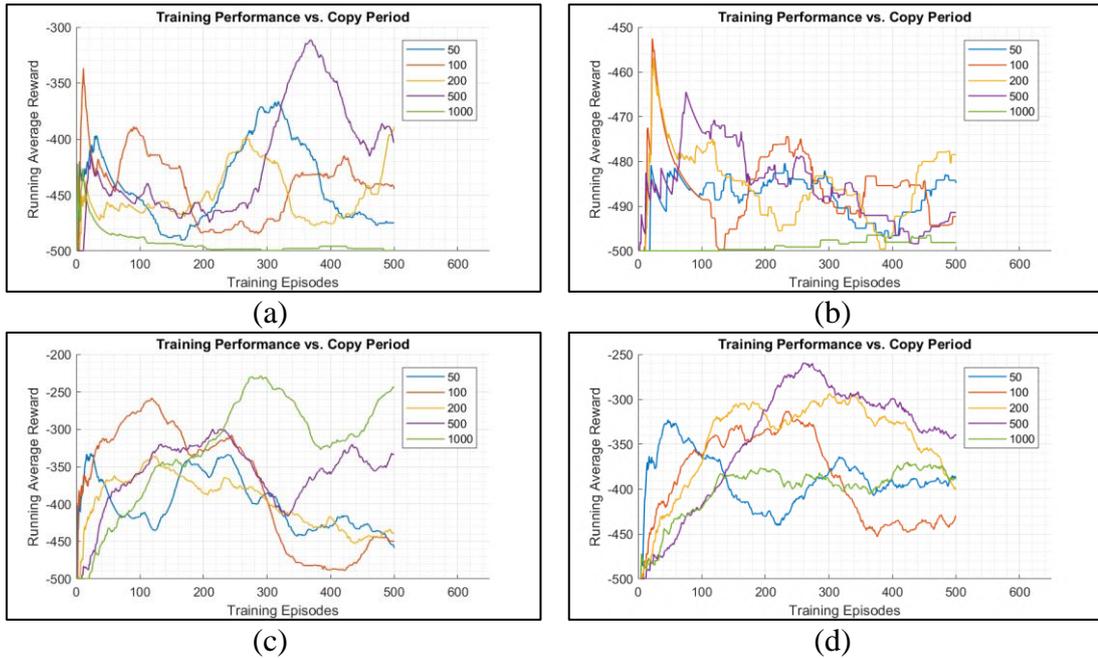

*Figure 7.22: Copy period tuning for DQN(a), DSN(b), DQN-DFA(c) and DSN-DFA(d) in Acrobot*

The last parameter to be tuned was the number of training episodes. This proved to be difficult, as for certain agent-environment pairs, training times exceeded the hardware time constraints which triggered a KILL signal. By analysing the results, it was found that a training period of 1000 episodes is long enough to ensure that the training performance converges to a stable value, while also not exceeding HPC time constraints.

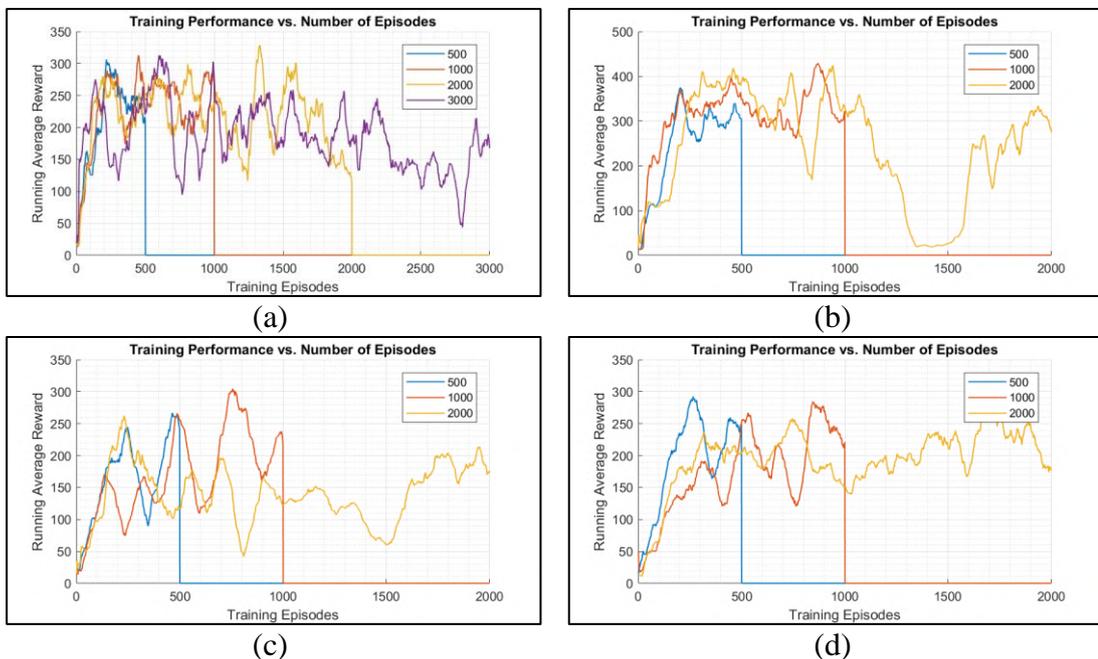

*Figure 7.23: Training episodes tuning for DQN(a), DSN(b), DQN-DFA(c) and DSN-DFA(d) in CartPole*

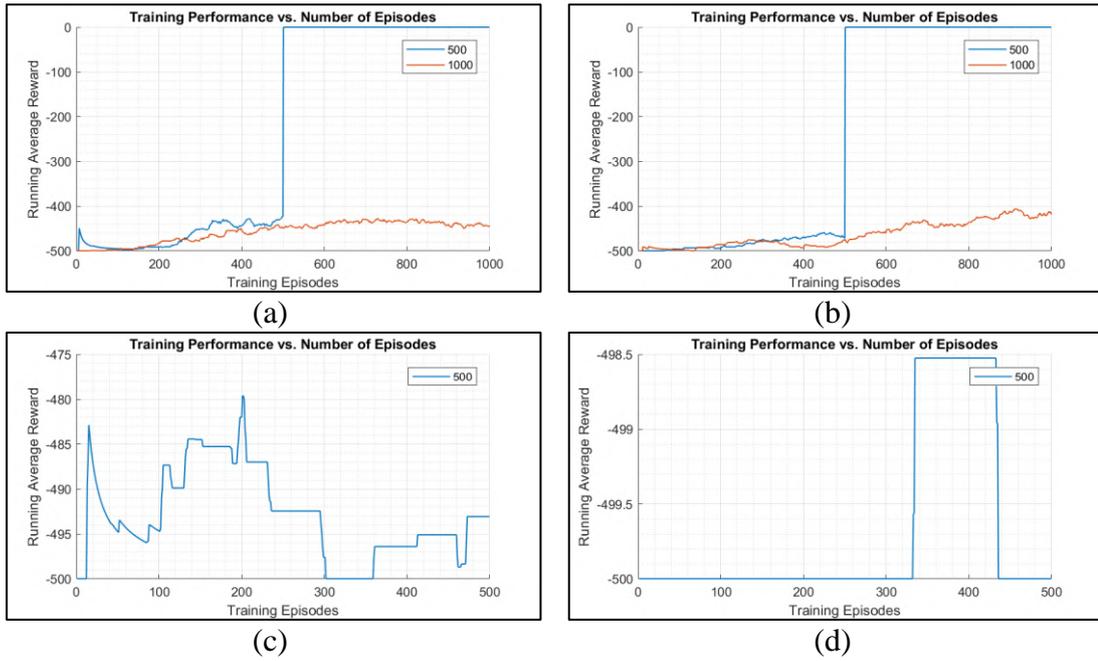

*Figure 7.24: Training episodes tuning for DQN(a), DSN(b), DQN-DFA(c) and DSN-DFA(d) in MountainCar*

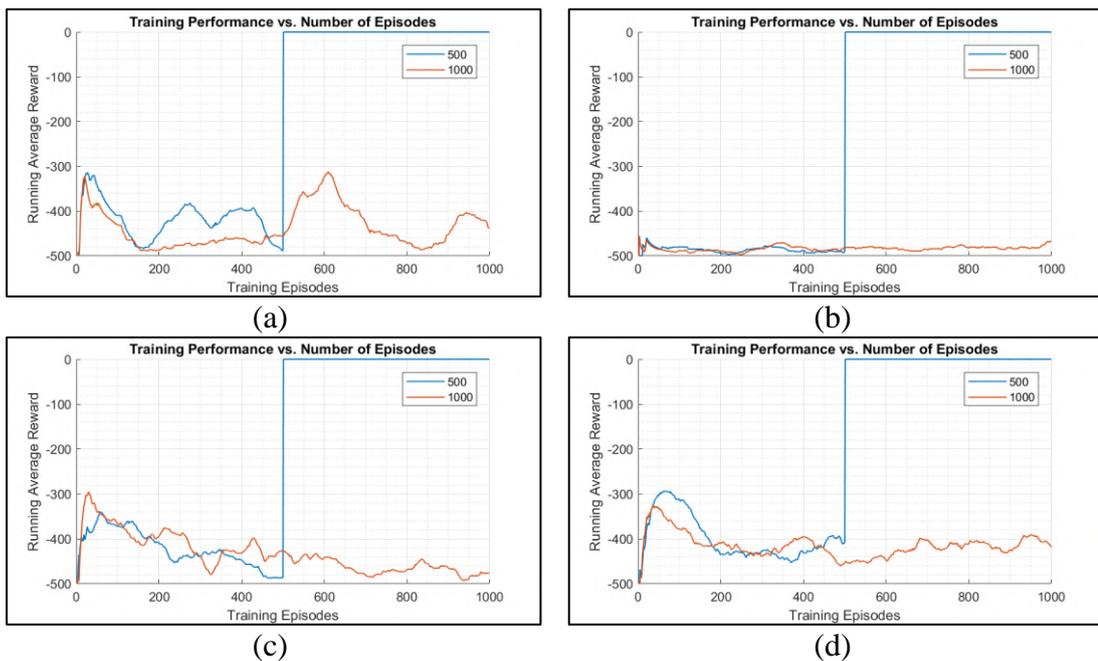

*Figure 7.25: Training episodes tuning for DQN(a), DSN(b), DQN-DFA(c) and DSN-DFA(d) in Acrobot*

The selected hyperparameters are presented in Table 7.1. It can be observed that they are close to the values presented in Table 6.4 and those described by Mnih (Mnih *et al.*, 2015).

*Table 7.1: List of hyperparameters of classical control agents and their values*

| Hyperparameter | Selected Value |
|---|---|
| Learning Rate $\alpha$ | $1e-4$ |
| Decay Rate $\beta$ | 0.99 |
| Discount Factor $\gamma$ | 0.99 |
| Small Scalar $\epsilon$ | $1e-3$ |
| Copy Period | 200 |
| Training Penalty | 0 |
| Training Episodes | 1000 |
| Minibatch Size | 32 |
| Replay Memory | 10,000 |
| Replay Start Size | 100 |
| Initial $\varepsilon$ | 1.0 |

### 7.2.2 Evaluation

Using the previously identified hyperparameters, a set of final agents was created, trained and evaluated. When assessing the training performance, it can be observed that, overall, all the agents train well, with the exception of the DSN and novel DFA agents using a softmax policy in the MountainCar environmnet (Figure 7.26 (d)). This could be due to the small number of input parameters (Table 6.1) which makes training more difficult. In Figure 7.26 (d) it can be observed that even the DQN algorithm only starts showing good training performance after 600 episodes. A longer training period could allow the other algorithms to show better performance. Across the entire agent-environment spectrum, the traditional algorithms appear to be training better than the novel ones, with some traditional agents managing to asymphtotically converge to an optimal policy. However, results such as the one in Figure 7.26 (f) show that novel algorithms, such as DSN-DFA, can actually achieve better performance than the traditional ones. Another observation is the fact that the stochastic nature of the epsilon-greedy policy can lead to large oscillations in the training performance (Figure 7.26 (a)), which causes the agent to get stuck in a local minimum for a number of episodes. As it will be later seen, this could cause problems during the agent's post-training evaluation.

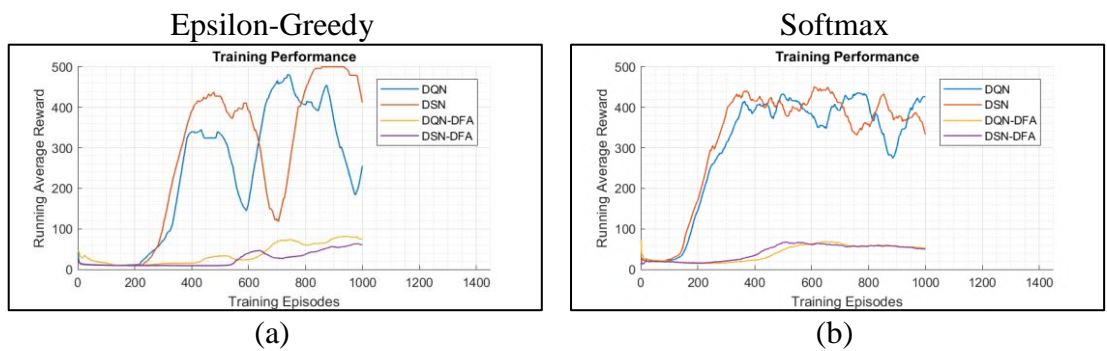

(a)                     (b)

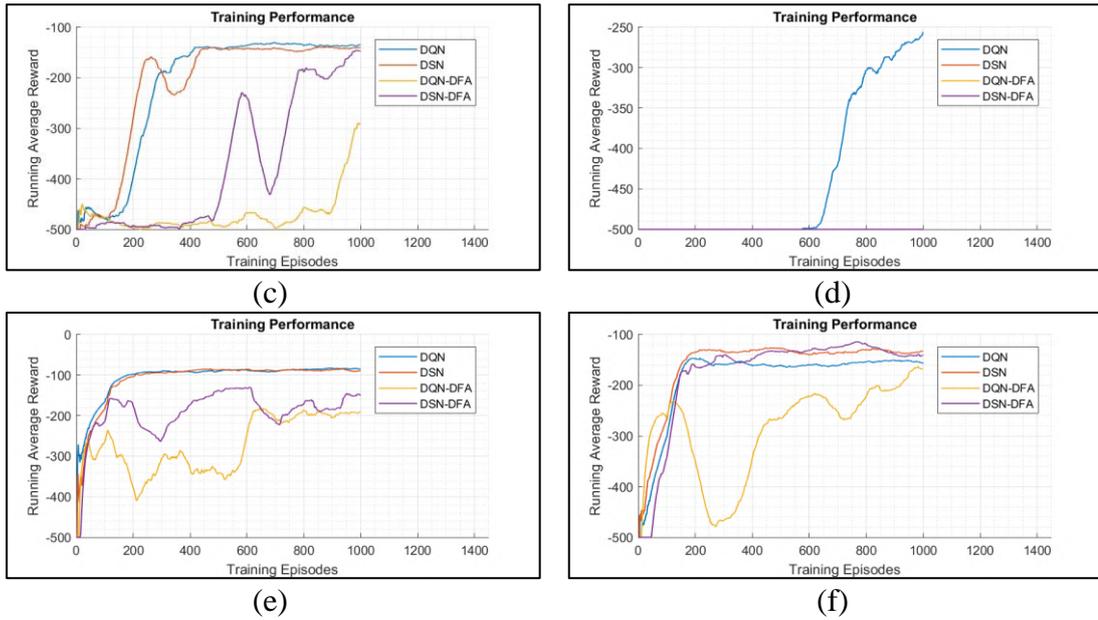

*Figure 7.26: Training performance of classical control agents using epsilon-greedy (a,c,e) and softmax (b,d,f) in CartPole (a,b), MountainCar (c,d) and Acrobot (e,f)*

The policy-induced oscillations can be even better observed when evaluating the agents during training. Although many of the agent-environment pairs display convergence during the training process and the evaluation, some others, such as the DSN epsilon-greedy CartPole agent (Figure 7.27 (a)) show how the agent can get stuck in a local minimum, which impacts the agent's overall evaluation performance. The opposite is also true, as agents can be stuck in a local maximum although their training performance idicates the oppsite. An example of this is the DQN epsilon-greedy agent, which shows a medioctre training performance (Figure 7.26 (a)) but displays a convergent behaviour when evaluated during training, even achieving the maximum score (Figure 7.27 (a)). Based on this finding, a convergence criteria can be derived for future investigations stating that training only stops when the mean evaluation reward during training stays withing a user-defined boundary.

Epsilon-Greedy                    Softmax

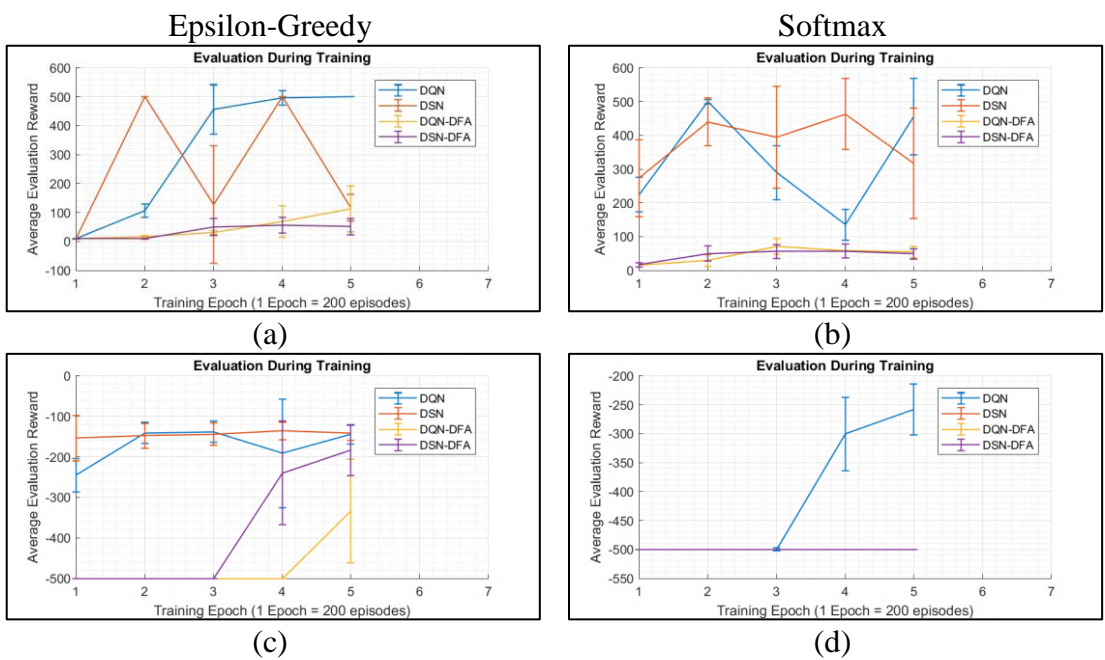

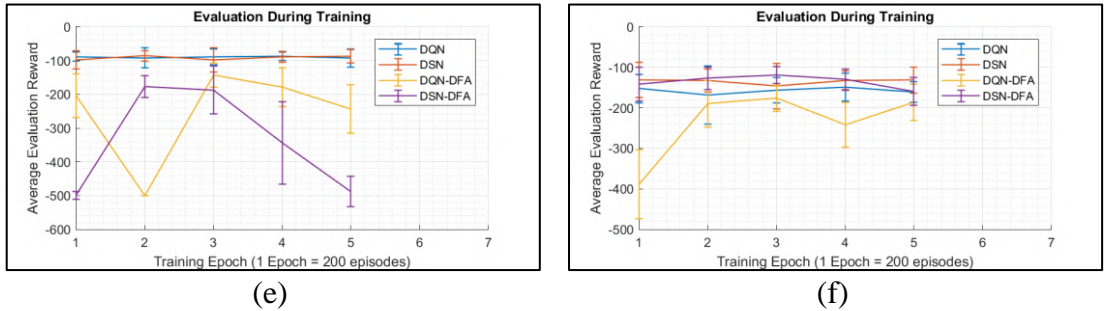

*Figure 7.27: Performance during training of classical control agents using epsilon-greedy (a,c,e) and softmax (b,d,f) in CartPole (a,b), MountainCar (c,d) and Acrobot (e,f)*

After the agents were trained, they were evaluated for 1000 episodes and their ranked performances plotted in Figure 7.28. For comparison, an agent wich performs completely random actions was introduced for each game. The first observation which can be made when assessing these results is that, overall, the agents trained using backpropagation generally perform better than those trained using direct feedback alignment. This is in line with Nøkland's previous observations for DNN's trained using DFA (Nøkland, 2016). Still, it can be observed that, for MountainCar, the DSN-DFA epsilon-greedy algorithm's performance (Figure 7.28 (b)) is close to that of the traditional algorithms. This is reinforced by the the data presented in Figures 7.26 (c) and 7.27 (c), where it can be seen that the DSN-DFA algorithm almost converges to the same values as the BP DQN and DSN agents. The same observation can also be made for the DSN-DFA and DQN-DFA softmax agents in Acrobot (Figure 7.28 (c)), which are also shown to have a good training behaviour in Figures 7.26 (f) and 7.27 (f). Judging from these findings, for a larger number of training episodes, it is possible that these algorithms would converge to the optimal policy and obtained maximum scores.

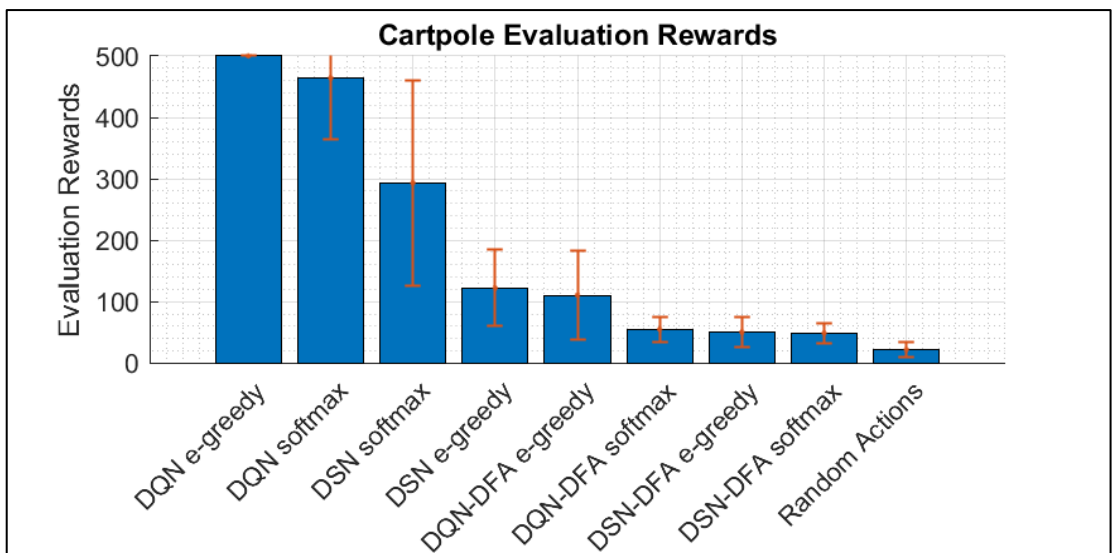

(a)

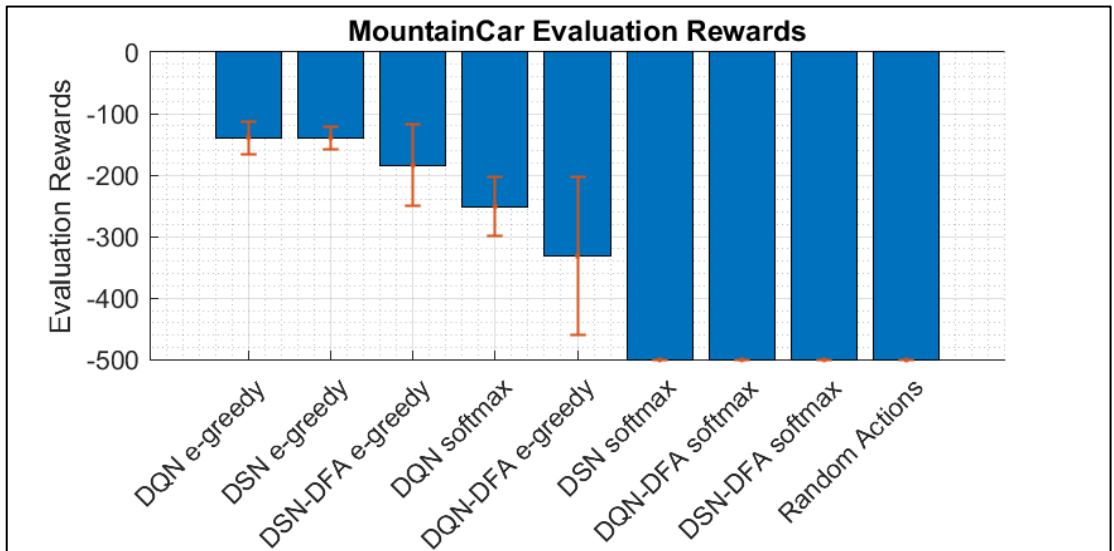

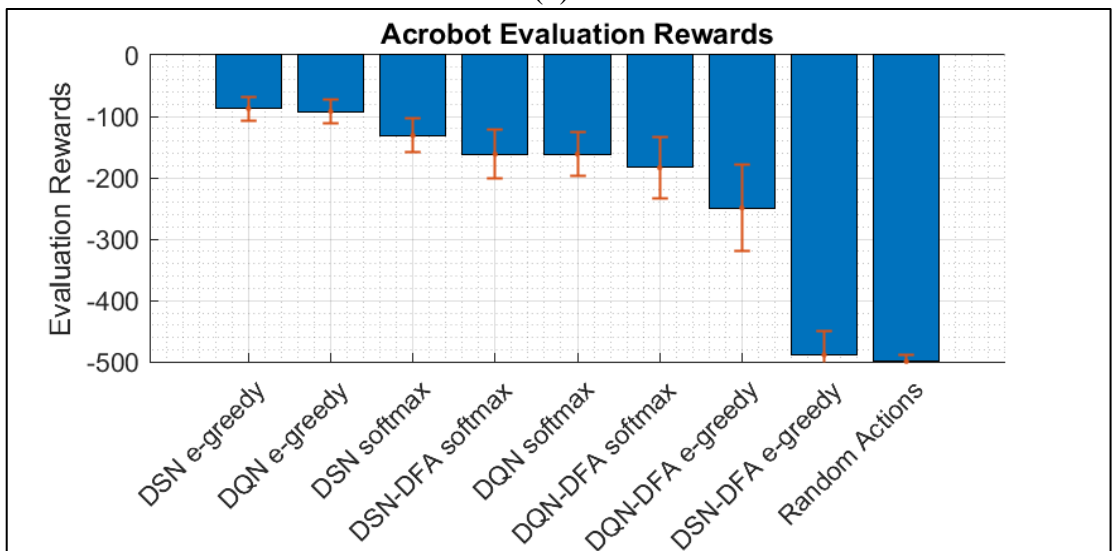

*Figure 7.28: Post-training performance of classical control agents in CartPole (a), MountainCar (b) and Acrobot (c)*

This represents an important finding, as it shows that DRL agents using DFA, despite taking slightly longer to train, can achieve comparable performances to already established algorithms. As DFA does not require symmetrical weights, or reciprocal connections, and can provide local training signals specific to each computational block by using local error signals, it represents a good contender for neuromorphic hardware applications. This finding opens the possibility of DRL algorithms being implemented on neuromorphic hardware, allowing the same agent to be trained for a variety of tasks. This solves the previous issues requiring the tailoring of each piece of hardware for a specific task, and represents and element of novelty.

To ensure the viability of this finding, a series of statistical tests was performed. Firstly, several univariate statistical values (Appendix B.1) were calculated for all the agents (Table 7.2). By analysing the values, it can be observed that none of the evaluation datasets display a normal distribution. These findings were reinforced by graphical statistical tests (Appendix B.2). In the interest of space, these plots are not displayed in this report.

*Table 7.2: Univariate statistical tests for normality for classical control agents*

| Game | Agent | Mean | Median | Standard Deviation | Skewness | Kurtosis |
|---|---|---|---|---|---|---|
| CartPole | DQN e-greedy | 500 | 500 | 0 | - | - |
| | DQN softmax | 464.22 | 500 | 100.77 | -2.77 | 9.44 |
| | DSN softmax | 293.32 | 196.5 | 167.36 | 0.32 | 1.26 |
| | DNS e-greedy | 122.24 | 106 | 62.33 | 4.21 | 22.76 |
| | DQN-DFA e-greedy | 110.72 | 82.5 | 71.7 | 1.78 | 7.16 |
| | DQN-DFA softmax | 54.99 | 50 | 20.01 | 2.35 | 15.53 |
| | DSN-DFA e-greedy | 50.45 | 44 | 24.11 | 3.13 | 22.74 |
| | DSN-DFA softmax | 49.05 | 45 | 16.05 | 1.75 | 7.41 |
| MountainCar | DQN e-greedy | -139.04 | -143 | 26.79 | 0.77 | 2.64 |
| | DSN e-greedy | -139.43 | -142 | 17.74 | 0.68 | 9.14 |
| | DSN-DFA e-greedy | -183.71 | -167 | 66.63 | -2.32 | 10.37 |
| | DQN softmax | -250.86 | -245 | 47.25 | -0.29 | 3.58 |
| | DQN-DFA e-greedy | -331.39 | -247 | 128.39 | -0.51 | 1.33 |
| | DSN softmax | -500 | -500 | 0 | - | - |
| | DQN-DFA softmax | -500 | -500 | 0 | - | - |
| | DSN-DFA softmax | -500 | -500 | 0 | - | - |
| Acrobot | DSN e-greedy | -87.3 | -86 | 19.6 | -3.14 | 26.44 |
| | DQN e-greedy | -91.9 | -90 | 20.01 | -2.75 | 14.82 |
| | DSN softmax | -130.98 | -128 | 27.64 | -4.07 | 46.68 |
| | DSN-DFA softmax | -161.44 | -154 | 40.36 | -2.88 | 17.46 |
| | DQN softmax | -161.57 | -157 | 35.14 | -2.26 | 14.76 |
| | DQN-DFA softmax | -183.46 | -176 | 49.19 | -3.25 | 19.75 |
| | DQN-DFA e-greedy | -248.86 | -232 | 69.76 | -1.78 | 6.52 |
| | DSN-DFA e-greedy | -488.05 | -500 | 39.32 | 3.87 | 18.74 |

Finally, to quantify the differences between the various datasets, Wilcoxon signed rank tests (Appendix B.3) were performed for all agents (Tables 7.3 – 7.5). It can be observed that for certain agents, the logical h-value is 0, indicating that the test fails to reject the null hypothesis. These findings can be explained for all agents, with some agents either not training at all, such as the MountainCar ones (Table 7.4, Figure 7.28 (b)), or with other agents converging to the same policy.

*Table 7.3: Wilcoxon signed rank tests for the Cartpole agents (p-value / logical h-value)*

| Cartpole | DQN e-greedy | | DQN softmax | | DSN softmax | | DNS e-greedy | | DQN-DFA e-greedy | | DQN-DFA softmax | | DSN-DFA e-greedy | | DSN-DFA softmax | |
|---|---|---|---|---|---|---|---|---|---|---|---|---|---|---|---|---|
| DQN e-greedy | | | 9e-24 | 1 | 1e-104 | 1 | 4e-165 | 1 | 4e-165 | 1 | 3e-165 | 1 | 3e-165 | 1 | 3e-165 | 1 |
| DQN softmax | 9e-24 | 1 | | | 1e-82 | 1 | 1e-161 | 1 | 1e-162 | 1 | 4e-165 | 1 | 3e-165 | 1 | 3e-165 | 1 |
| DSN softmax | 1e-104 | 1 | 1e-82 | 1 | | | 2e-132 | 1 | 9e-124 | 1 | 1e-164 | 1 | 1e-164 | 1 | 4e-165 | 1 |
| DNS e-greedy | 4e-165 | 1 | 1e-161 | 1 | 2e-132 | 1 | | | 1e-8 | 1 | 7e-160 | 1 | 1e-159 | 1 | 2e-107 | 1 |
| DQN-DFA e-greedy | 4e-165 | 1 | 1e-162 | 1 | 9e-124 | 1 | 1e-8 | 1 | | | 2e-107 | 1 | 3e-114 | 1 | 6e-132 | 1 |
| DQN-DFA softmax | 3e-165 | 1 | 4e-165 | 1 | 1e-164 | 1 | 7e-160 | 1 | 2e-107 | 1 | | | 1e-10 | 1 | 1e-13 | 1 |
| DSN-DFA e-greedy | 3e-165 | 1 | 3e-165 | 1 | 1e-164 | 1 | 1e-159 | 1 | 3e-114 | 1 | 1e-10 | 1 | | | 0.4 | 0 |
| DSN-DFA softmax | 3e-165 | 1 | 3e-165 | 1 | 4e-165 | 1 | 2e-107 | 1 | 6e-132 | 1 | 1e-13 | 1 | 0.4 | 0 | | |

*Table 7.4: Wilcoxon signed rank tests for the MountainCar agents (p-value / logical h-value)*

| MountainCar | DQN e-greedy | | DSN e-greedy | | DSN-DFA e-greedy | | DQN softmax | | DQN-DFA e-greedy | | DSN softmax | | DQN-DFA softmax | | DSN-DFA softmax | |
|---|---|---|---|---|---|---|---|---|---|---|---|---|---|---|---|---|
| DQN e-greedy | | | 0.1 | 0 | 2e-85 | 1 | 1e-164 | 1 | 3e-165 | 1 | 3e-165 | 1 | 3e-165 | 1 | 3e-165 | 1 |
| DSN e-greedy | 0.1 | 0 | | | 4e-92 | 1 | 6e-165 | 1 | 3e-165 | 1 | 1e-165 | 1 | 1e-165 | 1 | 1e-165 | 1 |
| DSN-DFA e-greedy | 2e-85 | 1 | 4e-92 | 1 | | | 1e-101 | 1 | 3e-133 | 1 | 3e-164 | 1 | 3e-164 | 1 | 3e-164 | 1 |
| DQN softmax | 1e-164 | 1 | 6e-165 | 1 | 1e-101 | 1 | | | 7e-35 | 1 | 3e-165 | 1 | 3e-165 | 1 | 5e-107 | 1 |
| DQN-DFA e-greedy | 3e-165 | 1 | 3e-165 | 1 | 3e-133 | 1 | 7e-35 | 1 | | | 5e-107 | 1 | 5e-107 | 1 | 5e-107 | 1 |
| DSN softmax | 3e-165 | 1 | 1e-165 | 1 | 3e-164 | 1 | 3e-165 | 1 | 5e-107 | 1 | | | 1 | 0 | 1 | 0 |
| DQN-DFA softmax | 3e-165 | 1 | 1e-165 | 1 | 3e-164 | 1 | 3e-165 | 1 | 5e-107 | 1 | 1 | 0 | | | 1 | 0 |
| DSN-DFA softmax | 3e-165 | 1 | 1e-165 | 1 | 3e-164 | 1 | 5e-107 | 1 | 5e-107 | 1 | 1 | 0 | 1 | 0 | | |

*Table 7.5: Wilcoxon signed rank tests for the Acrobot agents (p-value / logical h-value)*

| Acrobot | DSN e-greedy | | DQN e-greedy | | DQN e-greedy | | DSN-DFA softmax | | DQN softmax | | DQN-DFA softmax | | DQN-DFA e-greedy | | DSN-DFA e-greedy | |
|---|---|---|---|---|---|---|---|---|---|---|---|---|---|---|---|---|
| DSN e-greedy | | | 3e-9 | 1 | 5e-150 | 1 | 2e-161 | 1 | 3e-162 | 1 | 2e-163 | 1 | 5e-165 | 1 | 3e-165 | 1 |
| DQN e-greedy | 3e-9 | 1 | | | 4e-137 | 1 | 4e-161 | 1 | 2e-160 | 1 | 4e-162 | 1 | 3e-165 | 1 | 3e-165 | 1 |
| DSN softmax | 5e-150 | 1 | 4e-137 | 1 | | | 4e-88 | 1 | 1e-93 | 1 | 4e-140 | 1 | 6e-162 | 1 | 5e-165 | 1 |
| DSN-DFA softmax | 2e-161 | 1 | 4e-161 | 1 | 4e-88 | 1 | | | 0.3 | 0 | 7e-9 | 1 | 2e-138 | 1 | 3e-165 | 1 |
| DQN softmax | 3e-162 | 1 | 2e-160 | 1 | 1e-93 | 1 | 0.3 | 0 | | | 2e-37 | 1 | 2e-148 | 1 | 3e-165 | 1 |
| DQN-DFA softmax | 2e-163 | 1 | 4e-162 | 1 | 4e-140 | 1 | 7e-9 | 1 | 2e-37 | 1 | | | 1e-108 | 1 | 1e-163 | 1 |
| DQN-DFA e-greedy | 5e-165 | 1 | 3e-165 | 1 | 6e-162 | 1 | 2e-138 | 1 | 2e-148 | 1 | 1e-108 | 1 | | | 3e-161 | 1 |
| DSN-DFA e-greedy | 3e-165 | 1 | 3e-165 | 1 | 5e-165 | 1 | 3e-165 | 1 | 3e-165 | 1 | 1e-163 | 1 | 3e-161 | 1 | | |

## 7.3 Agents for Atari 2600

As previously mentioned, in contrast to the agents trained for classical control problems, due to time limitations and the long training periods, no hyperparameter tuning was performed for the Atari agent. Rather, two sets of agents were trained, using the original parameters identified by Mnih (Mnih *et al.*, 2015) and a set of modified parameters in line with observations made with the classical control problems. In addition to this aspect, the creation and training of these agents encountered several issues, which limited the availability of obtained data. Firstly, tensor dimensionality issues emerged when implementing the DFA algorithm, which in combination with the limited time available, caused the postponement of the implementation of these agents. A solution to this is proposed by Scheller (Scheller, 2017). Also, due to time constraints, it was decided only to implement agents employing epsilon-greedy in policy selection. Finally, driver issues on the NVIDIA® DGX-1™ introduced further delays in the agent training, preventing early identification of potential code errors.

When assessing the training performance (Figures 7.29 – 7.30), it can be observed that most of the agents are training well, except for the original DQN and DSN Seaquest agents. It can also be observed that all the training plots are experiencing periodic spikes. These can be attributed to the time limitation imposed on the HPC. To prevent the KILL signal from being activated and data being lost, agents were trained for 2-3 days, then stopped and then restarted, with the network parameters being saved and then reloaded. However, this approach caused the experience replay memory buffer to be lost and recreated every time, which slowed down training considerably and produced the observable spikes. Moreover, this lead the original Seaquest agents to be stuck in a local minimum, which explains their performance.

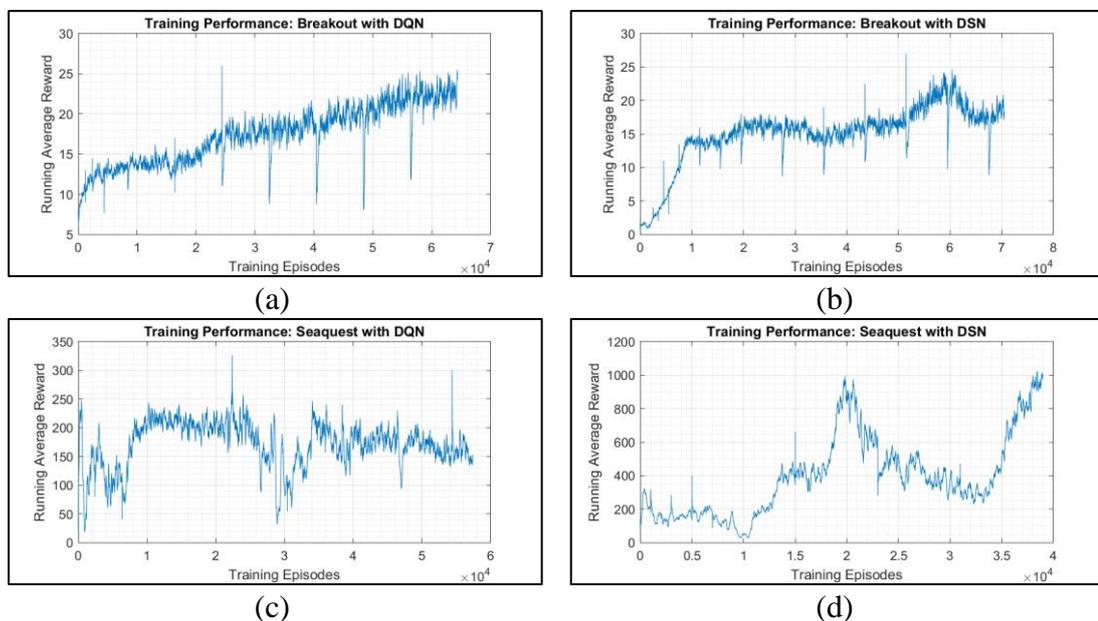

(a)          (b)

(c)          (d)

*Figure 7.29: Performance during training of modified Atari agents using epsilon-greedy in Breakout (a,b) and Seaquest (c,d)*

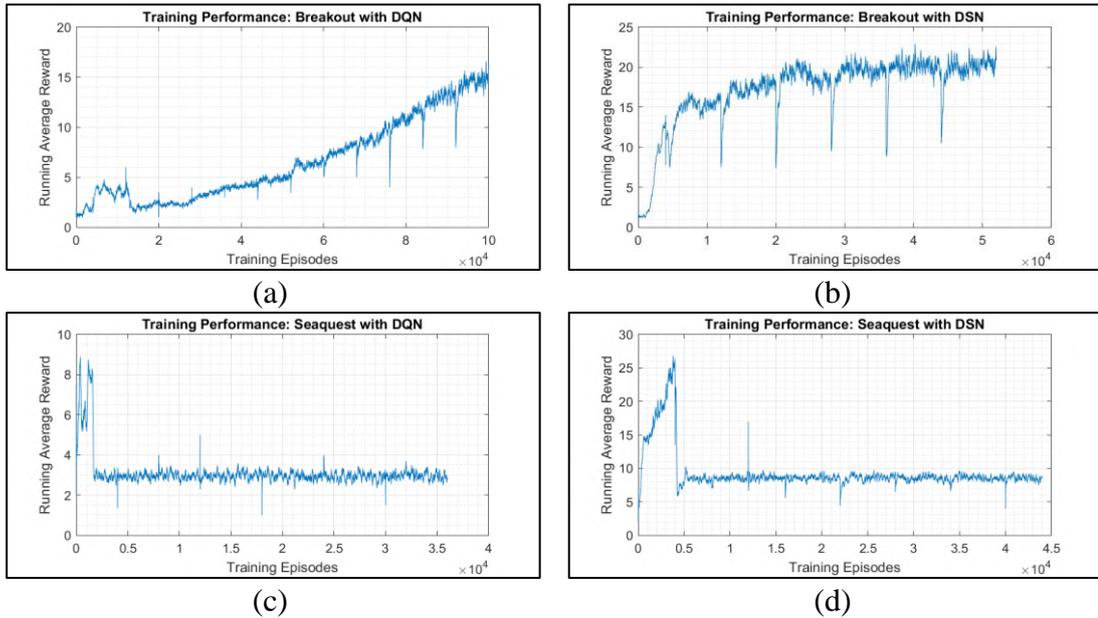

*Figure 7.30: Performance during training of original Atari agents using epsilon-greedy in Breakout (a,b) and Seaquest (c,d)*

Like the classical control agents, after training, the agents were evaluated for 1000 episodes, with their performances ranked in Figure 7.31. For comparison, for each game a random agent was introduced, as well as the score of a human player (the author). The results are then compared to those described in the original DQN paper (Mnih *et al.*, 2015) in Table 7.6.

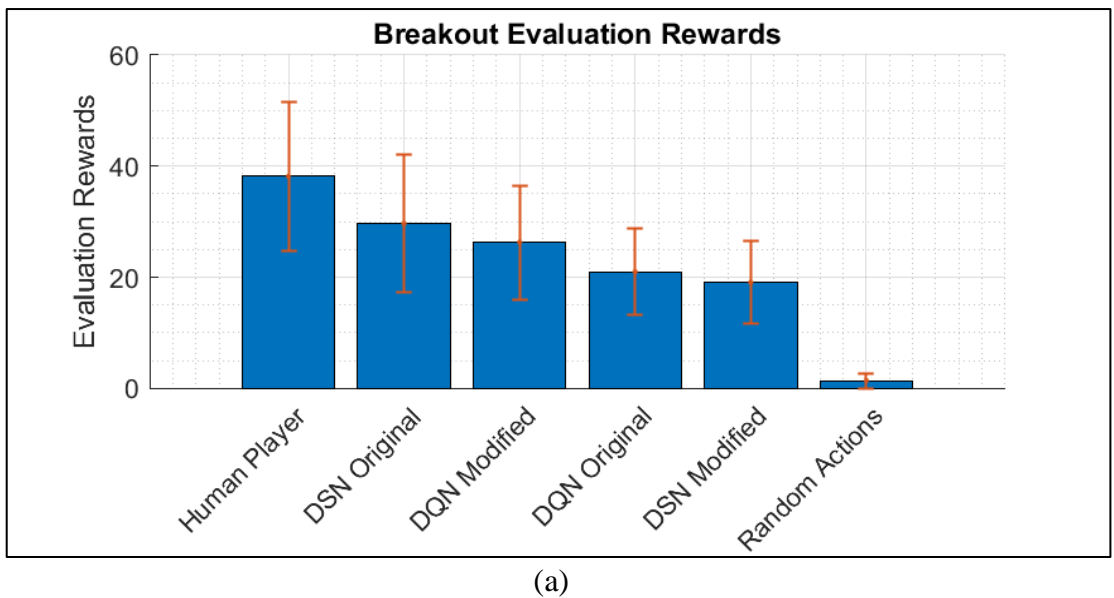

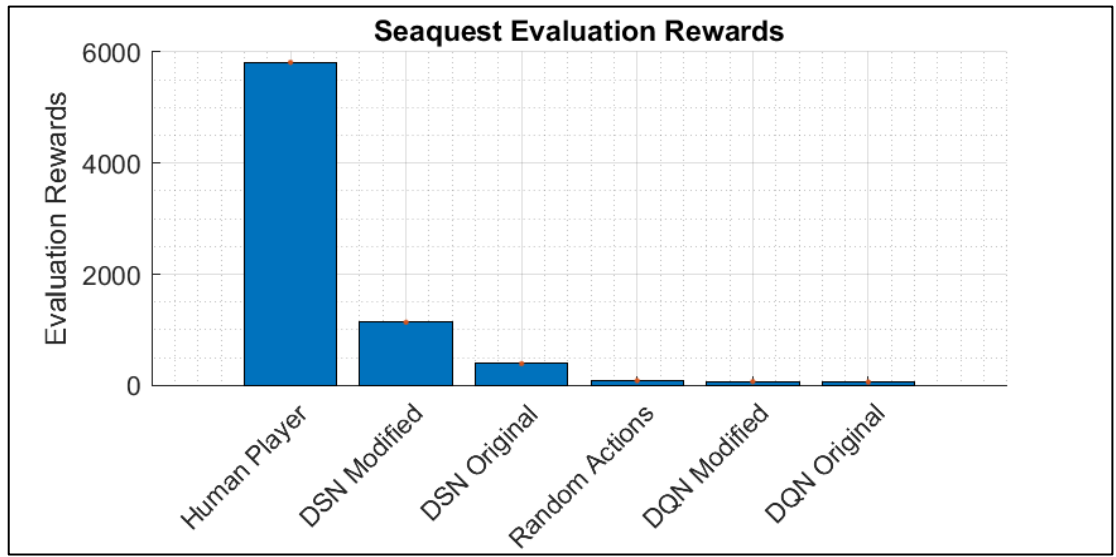

(b)

*Figure 7.31: Post-training performance of Atari agents in Breakout (a) and Seaquest(b)*

A first observation which can be made is that none of the agents achieve above-human level performance, not even the original agents. Moreover, both DQN Seaquest agents (Figure 7.31(b)) appear to perform worse than a random agent. This can be due to the previously identified training issues, as well as differences in the employed hyperparameters and architectures compared to literature. Another observation is that the DSN agents are typically performing better than the DQN agents, which is in line with Zhao's findings for the same games (Zhao *et al.*, 2017). However, none of the agents obtained comparable scores with those presented by Mnih or Zhao in their respective papers (Mnih *et al.*, 2015; Zhao *et al.*, 2017).

*Table 7.6: Comparison of Agent scores with similar DQN agents from literature*

| Game | Agent | Score (± standard deviation) | DQN score from literature (± standard deviation) (Mnih *et al.*, 2015) |
|---|---|---|---|
| Breakout | DSN original | 29.67 (±12.34) | 401.2 (±26.9) |
| | DQN modified | 26.29 (±10.25) | |
| | DQN original | 20.96 (±7.76) | |
| | DSN modified | 19.08 (±7.35) | |
| Seaquest | DSN modified | 1135.2 (±213.56) | 5286 (±1310) |
| | DSN original | 390.3 (±116.54) | |
| | DQN modified | 68.5 (±42.48) | |
| | DQN original | 59.88 (±37.01) | |

Finally, to verify the statistical significance of the above findings, the univariate statistical values (Table 7.7) and the Wilcoxon signed rank tests (Tables 7.8 – 7.9) were calculated for the data. None of the obtained datasets display a normal distribution, and neither dataset fails to reject the null hypothesis.

*Table 7.7: Univariate statistical tests for normality for Atari agents*

| Game | Agent | Mean | Median | Standard Deviation | Skewness | Kurtosis |
|------|-------|------|--------|--------------------|----------|----------|
| Breakout | DSN original | 29.67 | 28.5 | 12.35 | 0.62 | 3.21 |
| | DQN modified | 26.29 | 25 | 10.26 | 0.71 | 3.48 |
| | DQN original | 20.96 | 20 | 7.76 | 0.78 | 3.89 |
| | DSN modified | 19.08 | 18 | 7.35 | 0.96 | 3.91 |
| Seaquest | DSN modified | 1135.2 | 1140 | 213.67 | -0.43 | 2.96 |
| | DSN original | 390.3 | 380 | 116.6 | 0.14 | 2.76 |
| | DQN modified | 68.5 | 60 | 42.48 | 0.84 | 4.00 |
| | DQN original | 59.88 | 60 | 37.02 | 0.74 | 3.36 |

*Table 7.8: Wilcoxon signed rank tests for the Breakout agents (p-value | logical h-value)*

| Breakout | DSN original | | DQN modified | | DQN original | | DSN modified | |
|----------|--------------|---|--------------|---|--------------|---|--------------|---|
| DSN original | | | 3e-10 | 1 | 8e-62 | 1 | 2e-81 | 1 |
| DQN modified | 3e-10 | 1 | | | 2e-33 | 1 | 7e-60 | 1 |
| DQN original | 8e-62 | 1 | 2e-33 | 1 | | | 3e-8 | 1 |
| DSN modified | 2e-81 | 1 | 7e-60 | 1 | 3e-8 | 1 | | |

*Table 7.9: Wilcoxon signed rank tests for the Seaquest agents (p-value | logical h-value)*

| Seaquest | DSN modified | | DSN original | | DQN modified | | DQN original | |
|----------|--------------|---|--------------|---|--------------|---|--------------|---|
| DSN modified | | | 3e-165 | 1 | 3e-165 | 1 | 3e-165 | 1 |
| DSN original | 3e-165 | 1 | | | 4e-165 | 1 | 4e-165 | 1 |
| DQN modified | 3e-165 | 1 | 4e-165 | 1 | | | 1e-6 | 1 |
| DQN original | 3e-165 | 1 | 4e-165 | 1 | 1e-6 | 1 | | |

# 8   Conclusions and Proposed Future Work

This project aimed at investigating the performance of AI agents trained with both conventional DRL techniques, such as DQN and DSN, as well as novel methods which employ a the biologically plausible DFA algorithm for computing the gradients.

A detailed survey of relevant work and the state of the art in the field DRL was conducted, capturing relevant knowledge required in the development of the methods, architectures and algorithms needed for the purpose of this investigation. The literature review also served the purpose of accommodating potential readers with aspects of the investigated technology. The relevant numerical methods and techniques required for developing the algorithms necessary for this study were discussed in detail. Then, building on the presented methods, six algorithms were developed: two for simple games, such as Gridworld Maze Runner and Cliff Walker, two for classical control problems, including CartPole, MountainCar and Acrobot, and finally two for solving Atari 2600 games, such as Breakout and Seaquest.

The agents used for simple games highlighted the differences between on and off policy methods and indicated that agents trained using on-policy SARSA can reach better solutions under certain circumstances in complex environments than the off-policy established Q-learning method.

For the classical control problems, all agents were observed to train well, with some of the agents asymptotically converging to an optimal policy. However, more importantly, this investigation has found that under certain circumstances, novel DFA-based algorithms can achieve comparable performances with conventional algorithms, even potentially converging to an optimal policy. This represents an important finding, as it opens the possibility of deep and deep reinforcement learning algorithms being employed on neuromorphic hardware, solving the previous issues induced by backpropagation, such as the need for tailoring hardware for specific tasks.

Finally, the Atari 2600 agents reinforced the idea that AI agents can learn from raw sensory inputs. The agents also brought more evidence to support previous findings that SARSA-based algorithms can outperform Q-learning based agents. This, and the above findings, allowed the main objectives, and some of the secondary objectives of this investigation to be achieved. The drawback of this study consisted in the inability to produce an Atari agent which surpasses human-level performance. Nevertheless, the findings of this study prove that AI agents can be capable of learning from only raw sensory inputs and that DRL agents could theoretically be implemented on neuromorphic hardware.

Moving further, several issues need to be addressed, as well as several techniques can be further investigated. Firstly, this investigation has found that all the agents are very sensitive to the values of the utilized hyperparameters. Any future work should further refine the hyperparameters already described, as well as tune others such as the maximum number of steps per episode for classical agents, the number of hidden layers and nodes, or that of convolutional layers, the parameters characterising the epsilon decay or the softmax bias factor, the size of the experience buffer and the number of minimum training experiences and the size of the training batch. Techniques such as automated grid searches could aid in this effort. Modifications should also be done to the Atari agents to ensure that the experience replay buffer is preserved during training runs. The networks employed by the agents should also be refined through network optimisation techniques, such as dropout. Also, a continuous evaluation and early stopping mechanism can be implemented, stopping training when agents reach the desired performance. The value of epsilon can also be varied based on the training performance. Future work should also explore the opportunities of implementing DFA on other agent architectures, such as A3C DRL agents, or agents which learn using GAN-type approaches, where two agents play against each other.

# 9 Self-Review

This section represents a personal and critical evaluation of my progress to date. Consideration is given to the following topics: knowledge level, time management, quality of work and research-related skills.

My first contact with AI was while I was employed by Rolls-Royce Plc., through networking with colleagues working on AI-related projects. Appreciating how powerful these algorithms are and the vast number of potential applications, while wishing to do something unique, difficult and challenging, I decided to shift my career focus towards the field of AI.

Due to my aeromechanical background, I had little to no experience with neither AI nor programming. Filling the knowledge gap represented my first task which I believe I managed through a series of seven online courses, giving me an understanding of DL and RL algorithms and how to structure AI projects.

While working on the project, I encountered several difficulties. Firstly, during the literature review, I initially diverged from the topic of my investigation, due to the board nature of the field. However, due to my experience in dealing with ambiguous research topics, I managed to regain focus and perform an in-depth literature review. This allowed me to gain a better understanding of the field and of the involved mathematical methods. Then, during the creating and testing of the agents, I had to overcome my lack of coding experience, which was a slow process, ending up taking more than originally anticipated. However, through hard work, I managed to fill any gaps that I originally had.

Overall, I believe that the work I have done is of good quality, all aspects considered. Difficult decisions had to be made during the project, such as focusing more on the simple control problems when dimensionality issues occurred in the complex Atari agents, or changing the programming framework midway through the project when I realized Theno was not capable of achieving what was required. Despite this I enjoyed this project more than any aerospace related project that I have worked on so far. This has helped me learn how to better manage my time and account for unforeseen events, such as other academic assignments. Besides giving me a good understanding of the DRL field, this project has also helped me refine some of my research skills, such as critical thinking and performing literature reviews, as well as to become a better person overall, by helping me understand by analogy how people are modelled by their past experiences and future desires.

## Appendix A: Deep Learning General Concepts

This appendix provides the interested reader with some general, intuitive knowledge in terms of what a neural network is, how it works and what are some of the very basic equations employed in order to create it. At the end of this appendix, the interested reader will also find a set of resources that will give them a more in depth understanding of the various types of neural networks and their applications. This will allow interested readers to gain an appreciation of the more complex topics discussed in this document. However, it must be mentioned that the methods and techniques covered in the report are more complex that what is covered in this brief introduction.

Before addressing the topic of deep learning, one must understand what a neural network is. Put in a very simple way, a neural network is a function, whose goal is to take one or multiple inputs $x$ and labelled outputs $y$ and create an approximation which maps the inputs to the outputs (A.1). The output of this function can take multiple forms, being either a number, a class, a word or a Boolean operator. The output represents the approximate result that could be obtained when using a set of similar input parameters as those used for training.

$$f_{aprox}(x) = y \qquad (A.1)$$

The most basic type of neural network is known as a feed-forward neural network (Figure A.1). This is since the information flows only in one direction, from the input to the output layer. Feed-forward neural networks represent the fundamental block of most deep learning methods and algorithms. Of course, other, more specialized types of networks exist, such as convolutional neural networks, which are used in this investigation. More information about them can be found in the body of the report.

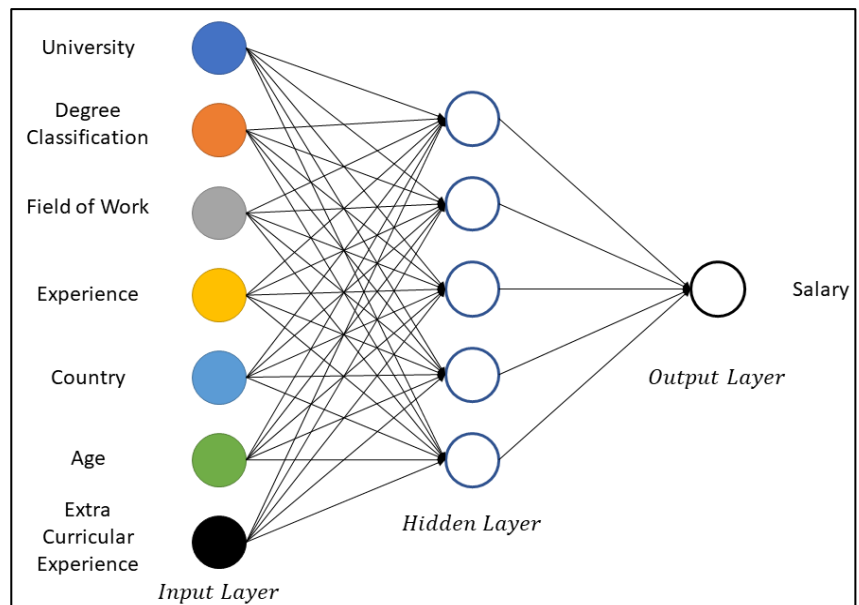

*Figure A.1: Example of a Simple Feed-forward Neural Network*

As an example, the inputs to the neural network in Figure A.1 are a series of metrics characterising a student about to enter the workforce. The network uses these metrics in order to predict the salary that this student will obtain once employed. Still, other types of data can be used as inputs, such as other numerical values, pieces of text, pieces of sound or images.

Neural networks are composed of numerous, smaller functions, associated with a graph which gives an indication of how these functions interact and come together. Therefore, the architecture of the overarching function is called a network. The smaller elements, composing a neural network are represented by stacked cell-like functions, called neurons. Stacks of neurons, in turn, form the hidden layer, which indirectly connects the inputs and the outputs of a network.

Hidden layers can be better understood as sets of vector values. Each neuron in the hidden layer receives numerous inputs from the other unit-neurons in previous layers and computes their own function. Thus, the neurons in a hidden layer act in parallel, presenting a vector to a scalar function (Goodfellow, Bengio and Courville, 2016). In other words, what the neural network does is take the provided inputs, perform a series of calculations using these inputs to obtain a series of new values, or features, and then combine these features to create an output which, hopefully, is as close as possible to the desired output provided in the training data. The behaviour of the neurons is not specified or hard coded by the programmer or the input-output dataset. The algorithm programming the neural network must decide on how to use and modify the parameters of these layers during training in such a way that it manages to obtain the desired outputs. The algorithm learns how to make these decisions by using the difference between what it has predicted and what the expected output is. This difference helps the network to better understand what its error is, and which part of the feed-forward path is responsible for inducing that error. This is achieved by backward propagating the error signal and observing where it is strongest, and then modifying the specific parts of the network in order to increase its accuracy before making the next prediction.

Hidden layers can have multiple neurons, and their number dictates the width of a network. Similarly, a network can have more than one hidden layer, the number of which gives a network's length. Networks which have more than one layer are called "deep", which is where the name of deep neural networks comes from (Figure A.2). It must be noted that the width of each hidden layer, or the neuron count, can be different. Also, the number of outputs of a network can vary, depending on the type of performed task.

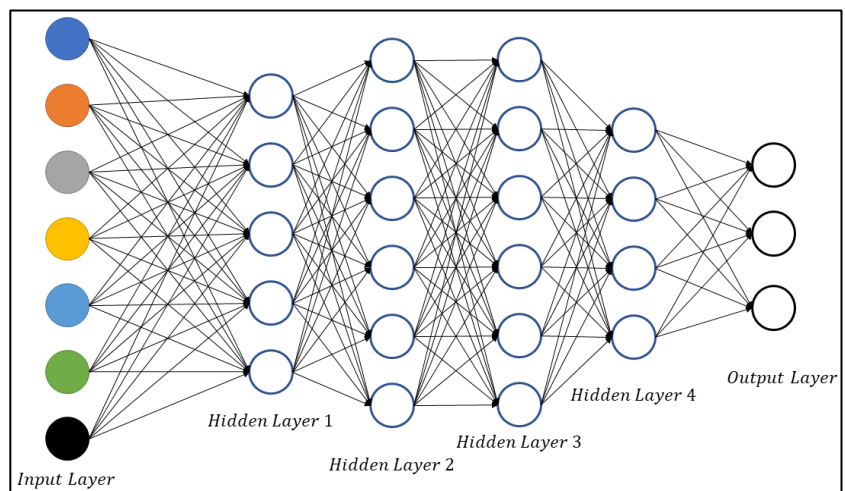

*Figure A.2: Example of a Deep Feed-forward Neural Network.*

Finally, the last layer of a neural network is known as the output layer and, during training, this is where the networks produces the results that it hopes match the ones provided in the training data. The types of unit-neurons used for an output layer are generally similar in terms of their functions to those used for the hidden layers, yet they can vary based on the type of application that the network is used for.

In order to understand how a neural network learns, one needs to understand how it works. When provided with input data, the network feeds the weighted sums of these inputs, or features, through an activation function. This function transforms the values before returning an output, which allows the network to model and extract information from complex features and learn in the process. When propagating the information through a network, each layer $l$ within the network has a set of defined weights, or slopes, $W$ and biases $b$ which are used to calculate the weighted sum of the inputs to that layer $Z$ (A.2 – A.3). The weights are usually initialized to small random values, while the biases are initialized either to 0 or a small positive value. Each neuron in the layer has its own parameters, which collectively form a vector characterising the layer. The value of the weighted sum is then transformed using a non-linear activation function $g$ to obtain the activation of that layer $\Lambda$. There are numerous activation functions being used for deep neural networks, including but not limited to: the sigmoid function, hyperbolic tangent function, rectified and "leaky" rectified linear units and others. Choosing the right activation function is generally dictated by the type of network employed and the purpose of the investigation.

$$Z^{[l]} = W^{[l]}\Lambda^{[l-1]} + b^{[l]} \qquad (A.2)$$

$$\Lambda^{[l]} = g^{[l]}\big(Z^{[l]}\big) \qquad (A.3)$$

Once the information has passed through the entire network, a prediction is made in the output layer. To determine how different this predicted output is from the intended training output, a cost function is usually calculated by using the cross-entropy between the two value sets. The form of this cost function depends on the specific form of the network and the model generated distribution $p_{model}$ (A.4) (Goodfellow, Bengio and Courville, 2016):

$$J(\theta) = -\mathbb{E}_{x,y\sim\hat{p}_{data}} \log P_{model}(y|x) \qquad (A.4)$$

The purpose of the network itself is to find the correct weights and biases so that it minimises the cost function. This ca be done by minimizing the cost as a function of each hidden layer's weights and biases. This is done by using gradient descent, which uses the gradients (or derivatives) of the cost function with respect to these parameters, to update them in place, in the direction of steepest descent, until they converge towards a local minimum. The partial derivatives in this approach are calculated typically using the backpropagation algorithm, which is a method of computing the gradient of the cost function with respect to its ancestor weights within the graph.

$$W^{[l]} = W^{[l]} - \alpha \cdot \frac{\partial J(W,b)}{\partial W^{[l]}} = W^{[l]} - \alpha \cdot dW^{[l]} \qquad (A.5)$$

$$b^{[l]} = b^{[l]} - \alpha \cdot \frac{\partial J(W,b)}{\partial b^{[l]}} = b^{[l]} - \alpha \cdot db^{[l]} \qquad (A.6)$$

The neural network is repeatedly trained by taking steps of gradient descent in order to reduce the cost function.

As mentioned at the beginning of this appendix, this represents just a basic and broad overview of neural network to help give an intuition about the concepts presented in this report. The interested reader can find in the table below a series of links to different resources which might provide them with more information about neural networks and some of the technical concepts and details, as well as algorithms, used throughout this project. These resources can be freely accessed by clicking the provided links, or by accessing the relevant references.

*Table A.1: General Resources for Helping Interested Readers*

| Title | Description | Link | Reference |
|---|---|---|---|
| Machine Learning & Artificial Intelligence (video) | A general and intuitive presentation video of the fields of Machine Learning and AI, destined for a wider audience. | Click Here | (Philbin and PBS Digital Studios, 2017b) |
| Neural Networks (video) | An intuitive clip about the various types of neural networks and their particularities. | Click Here | (Hill and PBS Digital Studios, 2018) |
| Computer Vision (video) | Continuing the Machine Learning and AI introduction in the previous clip, this video gives an intuitive and easy to understand presentation of computer vision and convolutional neural networks. | Click Here | (Philbin and PBS Digital Studios, 2017a) |
| Multi-Layer Neural Networks (article) | A more thorough, yet easy to understand mathematical formulation of multi-layer deep neural networks, building on the information previously presented in this appendix. | Click Here | (Ng *et al.*, 2019) |
| Deep Learning (textbook) | A fundamental textbook designed to help both students and professionals working in the field of machine and deep learning. | Click Here | (Goodfellow, Bengio and Courville, 2016) |
| Reinforcement Learning: An Introduction (textbook) | A widely used textbook, fundamental for the study of the reinforcement learning paradigm of machine learning. | Click Here | (Sutton and Barto, 2018) |
| Machine Learning (MOOC) | An online course providing a broad introduction to the field of machine learning. | Click Here | (Ng, 2019b) |
| Deep Learning Specialization (MOOC) | A five-course series, giving students insight into various types of deep neural networks, their implementation and application. | Click Here | (Ng, Katanforoosh and Mourri, 2019) |

# Appendix B: Statistical Tests

This appendix provides the interested reader with some knowledge in terms of the different statistical tools and tests used during this investigation. Firstly, some general univariate statistical tools are presented. These are used to give some information about the general tendencies and variability of the analysed data. In this investigation, they were used to assess the normality of the evaluation rewards data sets. Then, this appendix will present the quantile-quantile plot, which give a visual indication if the datasets are normally distributed, together with the histogram. Finally, a quick discussion of the Wilcox Signed Rank test is presented, which is used to assess the statistical differences between two datasets.

## B.1   Univariate Statistics

Given a set of $n$ observations $\{x_1, x_2, x_3, x_4 \dots x_n\}$, the average, or mean value $\bar{x}$ in the dataset can be calculated using (B.1).

$$\bar{x} = \frac{1}{n}\sum_{i=1}^{n} x_i \qquad (B.1)$$

For the same dataset, the median can be defined as being the value which separates the data into two equal halves, one containing a data population which is smaller or equal to the median value and another containing data which is larger or equal to the median value. The median $\tilde{x}$ is calculated by first ordering the dataset based on the values of the elements, and then using (B.2). It should be noted that, in the case of a normal distribution, the mean and median should have equal or very similar values.

$$\tilde{x} = \begin{cases} x_{\frac{n+1}{2}}, & if \ n \ is \ odd \\ \frac{x_{\frac{n}{2}} + x_{\frac{n+2}{2}}}{2}, & if \ n \ is \ even \end{cases}$$

$$(B.2)$$

There are cases when the data is not distributed symmetrically about the mean. This allows the introduction of the next metric of interest: the skewness. This represents a measure of how asymmetrically the data is distributed about the mean and is calculated using (B.3), where $s$ represents the sample standard deviation. The skewness for a normal distribution should be 0.

$$Skewness = \frac{1}{s^3}\sum_{i=1}^{n}(x_i - \bar{x})^3 \qquad (B.3)$$

$$s = \sqrt{\frac{1}{n-1}\sum_{i=1}^{n}(x_i - \bar{x})^2} \qquad (B.4)$$

The final metric of interest for this investigation is given by the kurtosis factor, which is an indication of the number of outlier data points. In other words, the kurtosis factors represent an indication of how heavily or lightly tailed the dataset distribution is. The kurtosis factor is calculated using (B.5) and should be equal to 3 for normally distributed datasets.

$$Kurtosis = \frac{1}{n \cdot s^4} \sum_{i=1}^{n} (x_i - \bar{x})^4 \qquad (B.5)$$

## B.2    Graphical Techniques

Histograms and quantile-quantile plots represent graphical techniques utilized for statistical analysis. In this investigation, they are used to indicate the normality of various datasets. A histogram represents an accurate probability distribution of a dataset, indicating the number of elements within each of a series of disjoint categories, or bins (Figure B.1).

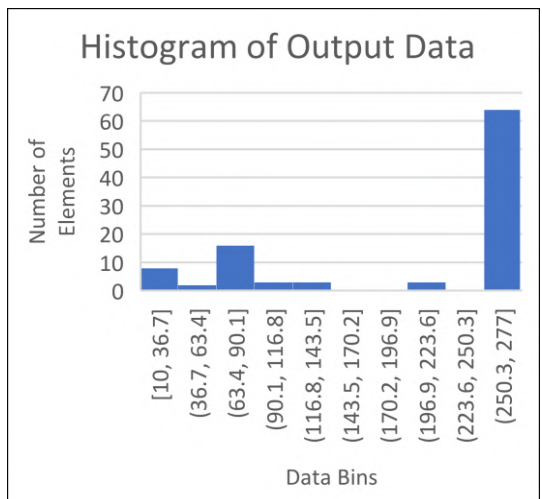

*Figure B.1: Example Histogram*

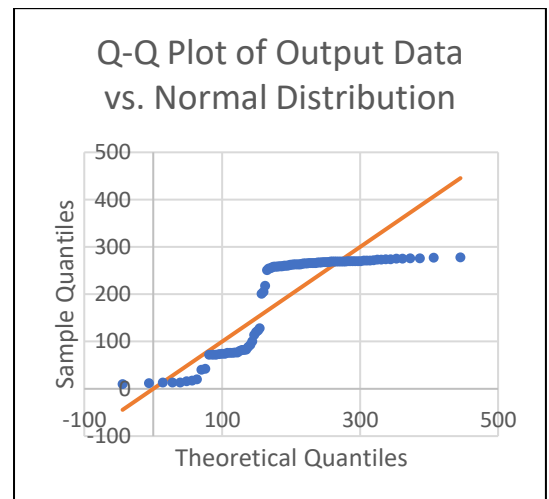

*Figure B.2: Example Q-Q Plot*

A Q-Q plots indicates if a certain dataset matches a specified distribution. Quantiles represent a division of the data into equally sized groups. The Q-Q plot displays the quantiles of the considered dataset against those of the specified distribution. If the dataset matches the intended probability distribution, then the elements will align along a 45-degree line, with little to no variation (Figure B.2).

## B.3    Wilcoxon Signed Rank Test

The Wilcoxon signed rank test is a non-parametric test which quantifies the differences between two data sets, by testing if they come from the same original phenomenon. In this study, the test is used to investigate the score changes between similar AI agents when part of their algorithms is modified. The Wilcoxon test works by testing the null hypothesis that the median of a distribution is equal to some given value. For reference, the null hypothesis states that no relationship is assumed between the two datasets.

Before the Wilcoxson test can be used, a series of checks need to be performed. Firstly, the input data needs to not follow a normal distribution. Then, according to Laerd Statistics, the data needs to pass 3 assumptions (Laerd Statistics, 2019):

1. The dependant variable should be measured at the ordinal or continuous level.
2. The independent variable should consist of two categorical related groups or matched pairs.
3. The distribution of the differences between the two related groups needs to be symmetrical in shape.

The first assumption is met, as this investigation is looking at the total rewards gathered by an AI agent as a measure of their performance. This means that the dependant variable is continuous. Then, the independent variables are kept contestant for the agents, which are placed in similar environments. Measurements are done on the same agent, using the same dependant variable, with only one metric being slightly modified. Finally, as the applied algorithmic changes are uniform, any differences are symmetrically distributed.

For the purpose of this study, the pre-built Wilcox signed rank test from MATLAB R2017b was used. This outputs the p-value and logical h-value. The sign rank function in MATLAB tests the null hypothesis that the input datasets comes from a distribution with zero median. If the p-value is $\leq 0.05$, then the null hypothesis is rejected. Alternatively, a large p value $> 0.05$ indicates that the null hypothesis cannot be rejected, meaning that the two datasets come from the same distribution. The h-value returns a logical 1 if the test rejects the null hypothesis, and a logical 0 if the test fails to reject the null hypothesis of zero median at a default 5% significance level. (MathWorks, 2019).

# Appendix C: Hardware Considerations

This appendix provides the interested reader with some knowledge in terms of the hardware used for the purpose of this investigation. The information concerning the ShARC HPC cluster was taken from The University of Sheffield's High Performance Computing cluster's documentation (The University of Sheffield, 2017).

As mentioned in Section 3.4, simple algorithms and preliminary tests for the more complex algorithms were conducted on the laptop used for creating the code. This computer runs an Intel® Core™ i7-6700 HQ CPU (2.6 GHz, 4 cores), and has 16 GB of installed DDR3 RAM.

The evaluation and testing of the agents based on a conventional DNN was carried out on server-grade CPUs available on ShARC. These were both publicly available and research group specific nodes. The publicly available nodes utilized were Intel® Xeon™ E5-2630 v3 CPUs (2.4 GHz, 8 cores) having access to 64 GB of installed DDR4 RAM, while the research group specific nodes were Intel® Xeon™ E5-2683 v4 CPUs (2.10 GHz, 16 cores) with 256 GB of DDR4 RAM.

Finally, the evaluation and testing of the Atari 2600 agents, which use a deep CNN, was carried out on both publicly available and research group specific GPUs. The publicly available utilized GPUs were NVIDIA® Tesla™ K80, capable of performing up to 8.74 Teraflops of single precision calculations, while the research group specific GPUs were NVIDIA® Tesla™ P100 GPUs, part of the NVIDIA® DGX-1™ deep learning supercomputer, which can perform up to 170 Teraflops of computation.

For reference, a teraflop is a unit of computing speed, equal to $10^{12}$ floating point operations per second.

# Appendix D: Gantt Chart

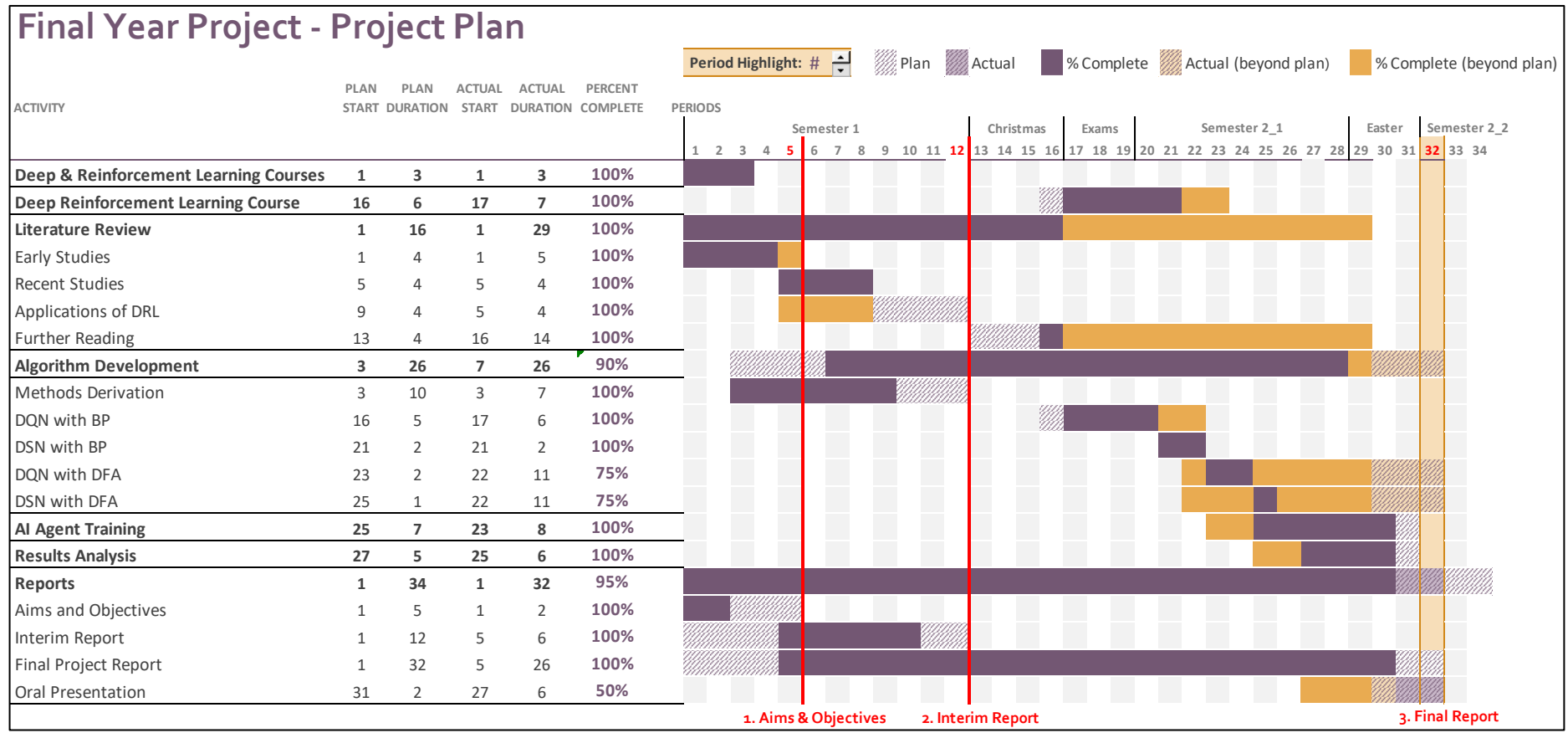